\documentclass{article}
\usepackage{graphicx}
\usepackage{multirow}
\usepackage{amsmath,amssymb} 

\usepackage{color}
\usepackage{url}
\usepackage{lscape}
\usepackage[numbers]{natbib}

\usepackage{algorithm}
\usepackage{algorithmic}
\usepackage[accepted]{icml2014}
\definecolor{lightgray}{gray}{.80}

\icmltitlerunning{Robust Eye Centers Localization}

\begin{document}

\twocolumn[ \icmltitle{Robust Eye Centers Localization \\ with Zero--Crossing Encoded Image
Projections}

\icmlauthor{Laura Florea}{laura.florea@upb.ro} \icmladdress{Image Processing and Analysis Laboratory\\
                University "Politehnica" of Bucharest, Romania,
                Address Splaiul Independen\c{t}ei 313}

\icmlauthor{Corneliu Florea}{corneliu.florea@upb.ro} \icmladdress{Image Processing and Analysis Laboratory\\
                University "Politehnica" of Bucharest, Romania,
                Address Splaiul Independen\c{t}ei 313}
\icmlauthor{Constantin Vertan}{constantin.vertan@upb.ro} \icmladdress{Image Processing and Analysis Laboratory\\
                University "Politehnica" of Bucharest, Romania,
                Address Splaiul Independen\c{t}ei 313}

\icmlkeywords{ eye localization , Image projections , zero-crossing encoding , real-time}

\vskip 0.3in

]

\begin{abstract}
This paper proposes a new framework for the eye centers localization by the joint use of encoding
of normalized image projections and a Multi Layer Perceptron (MLP) classifier. The encoding is
novel and it consists in identifying the zero-crossings and extracting the relevant parameters from
the resulting modes. The compressed normalized projections produce feature descriptors that are
inputs to a properly-trained MLP, for discriminating among various categories of image regions. The
proposed framework forms a fast and reliable system for the eye centers localization, especially in
the context of face expression analysis in unconstrained environments. We successfully test the
proposed method on a wide variety of databases including BioID, Cohn-Kanade, Extended Yale B and
Labelled Faces in the Wild (LFW) databases.
\end{abstract}


\section{Introduction}

\label{sec:intro}

As noted in the review on the eye localization topic, ''eye detection and tracking remains
challenging due to the individuality of eyes, occlusion, variability in scale, location, and light
conditions''\cite{Hansen:10}. Eye data and details of eye movements have numerous applications in
face detection, biometric identification, and particularly in human-computer interaction tasks.
Among the various applications of the eye localization topic, we are particularly interested in
face expression analysis. Thus, while any method is supposed to perform accurately enough on the
real-life cases and be fast-enough for real-time applications, we show an additional interest in
the cases where eye centers are challenged by face expression. We will prove that the proposed
method, which uses a MLP to discriminate among encoded normalized image projections from patches
centered on the eye and, respectively, from patches shifted from the eye, is both accurate and
fast.

\subsection{Related work}
The problem of eye localization was well investigated in literature, within a long history
\cite{Hansen:10}. Methods for eye center (or iris or pupil) localization in passive, remote imaging
may approach the problem either as a particular case of pattern recognition application,
\cite{Hamouz:05}, \cite{Asteriadis:09} or by using the physical particularities of the eye, like
the high contrast to the neighboring skin \cite{Wu:03} or the circular shape of the iris
\cite{Valenti:08}. The proposed method combine a pattern recognition approach with features that
make use of the eye's high contrast.

One of the first eye localization attempts is in the work from \cite{Kanade:73}, who used image
projections for this purpose. Taking into account that our method also uses image projections for
localization, in the next paragraphs we shall present state of the art methods by going from the
conceptually closest to the wider categories. Namely we shall start by presenting  solutions based
on projections, to follow with general eye localization methods and face fiducial points
localization algorithms.

As a general observation, we note that while older solutions \cite{Jesorsky:01}, \cite{Wu:03},
tried to estimate also the face position, since the appearance of the Viola-Jones face detection
solution \cite{Viola:04}, eye center  search is limited to a subarea within the upper face square.

\paragraph{Projections based methods}
The same image projections as in the work of Kanade are used to extract information for eye
localization in a plethora of  methods \cite{Feng:98}, \cite{Zhou:04}, \cite{Turkan:08}.
\cite{Feng:98} start with a snake based head localization followed by anthropometric reduction
(relying on the measurements from \cite{Verjak:94}) to the so-called eye-images and introduce the
variance projections for localization. The key points of the eye model are the projections
particular values, while the conditions are manually crafted.

\cite{Zhou:04} describe convex combinations between integral image projections and variance
projections that are named generalized projection functions. These are filtered and analyzed for
determining the center of the eye. The analysis is also manually crafted and requires
identification of minima and maxima on the computed projection functions. Yet in specific
conditions, such as intense expression or side illumination, the eye center does not correspond to
a minima or a maxima in the projection functions. \cite{Liu:10} use similar conditions with the
ones used in \cite{Zhou:04} but applied solely on the integral projections to detect if an eye is
open or closed.

\citeyear{Turkan:08} introduce the edge projections and use them to roughly determine the eye
position. Given the eye region, a feature is computed by concatenation of the horizontal and
vertical edge image projections. Subsequently, a SVM--based identification of the region with the
highest probability is used for marking the eye. The method from \cite{Turkan:08} is, to our best
knowledge, the only one, except ours which uses image projections coupled with machine learning.
Yet we differ by using supplementary data coupled with the introduction of efficient computation
techniques and elaborated pre and post-processing steps to keep the accuracy high and the running
time low.

\paragraph{General eye localization methods}
There are many other approaches to the problem of eye localization. \citeyear{Jesorsky:01} propose
a face matching method based on the Hausdorff distance followed by a MLP eye finder.
\citeyear{Wu:03} even reversed the order of the typical procedure: they use eye contrast specific
measures to validate possible face candidates.

\citeyear{Cristinacce:04} rely on the Pairwise Reinforcement of Feature  Responses algorithm for
feature localization. \citeyear{Campadelli:06} use SVM on optimally selected Haar wavelet
coefficients.

\citeyear{Hamouz:05} refine with SVM the Gabor filtered faces, for locating 10 points of interest;
yet the overall approach is different from the face feature fiducial points approach that is
discussed in the next paragraph. \citeyear{Niu:06} use an iteratively bootstrapped boosted cascade
of classifiers based on Haar wavelets. \citeyear{Kim:07} use multi scale Gabor jets to construct an
Eye Model Bunch. \citeyear{Asteriadis:09} use the distance to the closest edge to describe the eye
area. \citeyear{Valenti:08}, \cite{Valenti:12} use isophote's properties to gain invariance and
follow with subsequent filtering with Mean Shift (MS) or nearest neighbor on SIFT feature
representation for higher accuracy. \citeyear{Asadifard:10} relies on thresholding the cumulative
histogram for segmenting the eyes. \citeyear{Ding:10} train a set of classifiers to detect multiple
face landmarks, including explicitly the pupil center, by using a sliding window approach and test
in all possible locations and inter-connect them to estimate the shape overall. \citeyear{Tim:11}
rely their eye localizer on gradient techniques and search for circular shapes.
\citeyear{Gonzalez:13} use an exhaustive set of similarity measures over basic features such as
histograms, projections or contours to extract the eye center location having in mind the specific
scenario of driver assistance.

\paragraph{Face fiducial points localization}
More recently, motivated by the introduction by \citeyear{Cootes:01} of the active appearance
models (AAM), that simultaneously determine  a multitude of face feature points, a new class of
solutions, namely the localization of face fiducial points appeared.  In this category we include
the algorithm of \citeyear{Vukadinovic:05} who use a GentleBoost algorithm for combining Gabor
filters extracted features; \citeyear{Milborrow:08}, who extend the original active shape models
with more landmark points and stacks two such models; \citeyear{Valstar:10}, who model shapes using
the Markov Random Field and classify them using SVM in the so-called Borman algorithm;
\citeyear{Belhumeur:11} who use Bayesian inference on SIFT extracted features and most recently
\citeyear{Mostafa:12} who use a combination of regularized boosted classifiers and mixture of
complex Bingham distributions over texture and shape related features.

\subsection{Paper Structure}
In this paper we propose a system for eye centers localization that starts with face detection and
illumination type detection, followed by a novel feature extractor, a MLP classifier for
discriminating among possible candidates and a post-processing step that determines the eye  centers.
We contribute by:
\begin{itemize}
    \item Describing a procedure for \emph{fast} image projections computation. This step is critical in
    having the solution run in real time.
    \item Introducing a \emph{new encoding} technique to image analysis domain.
    \item The combination of \emph{normalized} image projections with zero-crossing based encoding
         results in image description features named \emph{Zero-crossing based Encoded image Projections}
        (ZEP). They are fast, simple, robust and easy to compute and therefore have applicability in a
        wider variety of problems.
    \item The integration of the features in a \emph{framework} for the problem of eye localization. We
        will show that description of the eye area using ZEP leads to significantly better results
        than state of the art methods in real-life cases represented by the extensive and very difficult
        Labeled Faces in the Wild database. Furthermore, the complete system is the fastest known
        in literature among the ones reporting high performance.
\end{itemize}

The remainder of this paper is organized as follows: Section \ref{Sect:ImProj} reviews the concepts
related to Integral Projections and describes a fast computation method for them; Section
\ref{Sect:Encoding} summarizes the encoding procedure and the combination with image projections to
form the ZEP features. The paper ends with implementation details, with a discussion on the
achieved results in the field of eye localization and proposals for further developments.


\section{Image Projections}
\label{Sect:ImProj}

\subsection{Integral Image Projections}

The integral projections, also named integral projection functions (IPF) or amplitude projections,
are tools that have been previously used in face analysis. They appeared as ``amplitude
projections'' \cite{Becker:64} or as ``integral projections'' \cite{Kanade:73} for face
recognition. For a gray-level image sub--window $I(i,j)$ with $i = i_m\dots i_M$ and $j = j_n\dots
j_N$, the projection on the horizontal  axis is the average gray--level along the columns
(\ref{Eq:IPF_H}), while the vertical axis projection is the average gray--level along the rows
(\ref{Eq:IPF_V}):

\begin{equation}
    \label{Eq:IPF_H}
    P_H(j)= \frac{1}{i_M-i_m+1}\sum_{i=i_m}^{i_M} I(i,j), \forall j=j_n , \dots, j_N
\end{equation}

\begin{equation}
    \label{Eq:IPF_V}
    P_V(i)= \frac{1}{j_N-j_n+1} \sum_{j=j_n}^{j_N} I(i,j), \forall i=i_m, \dots, i_M
\end{equation}

The integral projections reduce the dimensionality of data from 2D to 1D, describing it up to a
certain level of details. Also, the projections can be computed on any orthogonal pair of axes, not
necessarily rows and columns. This will be further discussed in subsection \ref{Subsect:DimReduct}.


\subsection{Edge Projections}
Over time, several extensions of the integral projections have been introduced such as variance
projection functions \cite{Feng:98} or edge projection functions (EPF) \cite{Turkan:08}.

Instead of determining edges with wavelet transform as in the case of \cite{Turkan:08}, we use a
different approach for computing the edge projections. First, the classical horizontal and vertical
Sobel contour operators (for details see \cite{Gonzalez:01} sect. 3.7) are applied, resulting in
$S_H$ and $S_V$ which are combined in the $S(i,j)$ image used to extract edges:

\begin{equation}
    \label{Eq:Sobel} S(i,j)= S_H^2(i,j) + S_V^2(i,j)
\end{equation}

The edge projections are computed on the corresponding image rectangle $I(i,j)$:

\begin{equation}
    \label{Eq:EPF_H}
    E_H(j)= \frac{1}{i_M-i_m+1}\sum_{i=i_m}^{i_M} S(i,j), \forall j=j_n,\dots, j_N
\end{equation}

\begin{equation}
    \label{Eq:EPF_V}
    E_V (i)= \frac{1}{j_N-j_n+1} \sum_{j=j_n}^{j_N}  S(i,j), \forall i=i_m,\dots,i_M
\end{equation}

Equations (\ref{Eq:EPF_H}) and (\ref{Eq:EPF_V}) are simply equations (\ref{Eq:IPF_H}) and
(\ref{Eq:IPF_V}) applied on the Sobel edge image $S(i,j)$.

As Sobel operator is invariant to additive changes, if compared to other types of projections, the
edge projections are significantly more stable with respect to illumination changes.

\subsection{Fast Computation of Projections}

While sums over rectangular image sub--windows may be easily computed using the concept of summed
area tables \cite{Crow:84} or integral image \cite{Viola:04}, a fast computation of the integral
image projections may be achieved using the \emph{prefix sums} \cite{Blelloch:90} on rows and
respectively on columns. A prefix-sum is a cumulative array, where each element is the sum of all
elements to the left of it, inclusive, in the original array.  They are the 1D equivalent of the
integral image, but they definitely precede it as recurrence $x_i = a_i + x_{i - 1}$ is known for
many years.

For the fast computation of image projections, two tables are required: one will hold prefix sums
on rows (a table which, for keeping the analogy with integral image, will be named horizontal 1D
integral image) and respectively one vertical 1D integral image that will contain the prefix sums
on columns. It should be noted that computation on each row/column is perform separately. Thus, if
the image  has $M \times N$ pixels, the 1D horizontal integral image, on the column $j$,
$\mathcal{I}_H^j$, is:

\begin{equation}
    \label{Eq:H_II}
    \mathcal{I}_H^j(i)= \sum_{k=1}^{i}  I(k,j)\enspace, \forall i=1,\dots M
\end{equation}

\begin{figure}[tb]
    \center
    \includegraphics[width=0.45 \textwidth]{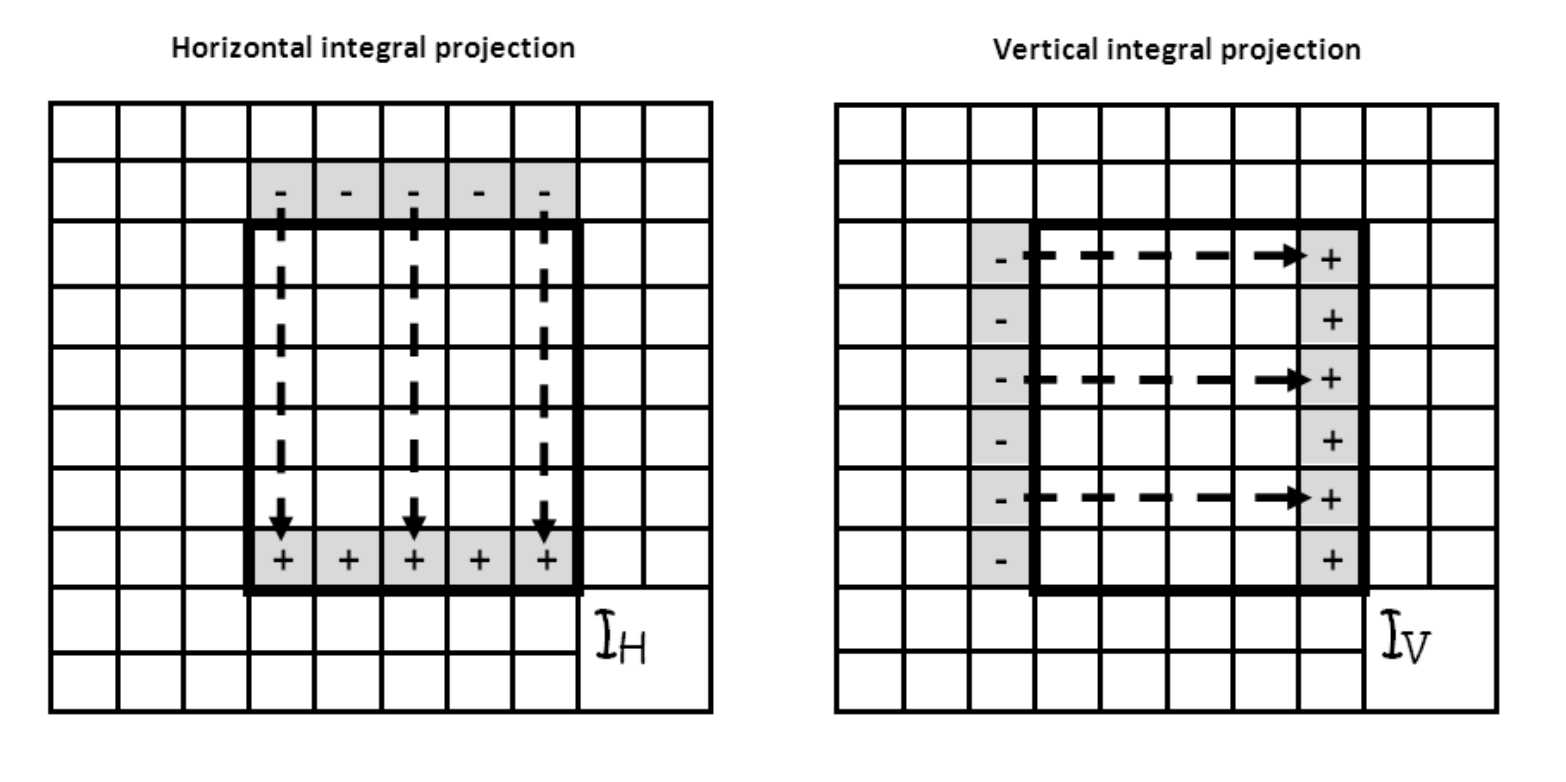}
    \caption{Given the two 1D oriented integral images, $\mathcal{I}_H^j$ and $\mathcal{I}_V^i$,
    each element of the integral projections, $P_H$ and $P_V$ that describes the marked sub-window
    is found by a simple subtraction. } \label{Fig:Fastprojections}
\end{figure}

Thus, the horizontal integral projection corresponding to the rectangle $i = [i_m ;  i_M] \times
[j_n; j_N]$ is:

\begin{equation}
    \label{Eq:PH_II}
    P_H(j)= \frac{1}{i_M-i_m+1} \left( \mathcal{I}_H^j(i_M) -\mathcal{I}^j_H(i_m-1)\right)
\end{equation}
The procedure is visually exemplified in figure \ref{Fig:Fastprojections}.

Using the oriented integral images, the determination of  the integral projections functions on all
sub-windows of size $K\times L$ in an image of $M \times N$ pixels requires one pass through the
image and $2\times M \times N$ additions, $2 \times (M -K) \times (N-L)$ subtractions and two
circular buffers of $(K+1) \times (N+1)$ locations, while the classical determination requires
$2\times K \times L \times (M -K) \times (N-L)$ additions.  Hence, the time to extract the
projections associated with a sub-window, where many sub-windows are considered in an image, is
greatly reduced.

The edge projections require the computation of the oriented integral images over the Sobel edge
image, $S(i,j)$. This image needs to be computed on the areas of interest.

In conclusion, the fast computation of projections opens the direction of real-time feature
localization  on high resolution images.

\section{Encoding and ZEP Feature}
\label{Sect:Encoding}

\begin{figure}[tb]
    \center
    \includegraphics[width=0.45 \textwidth]{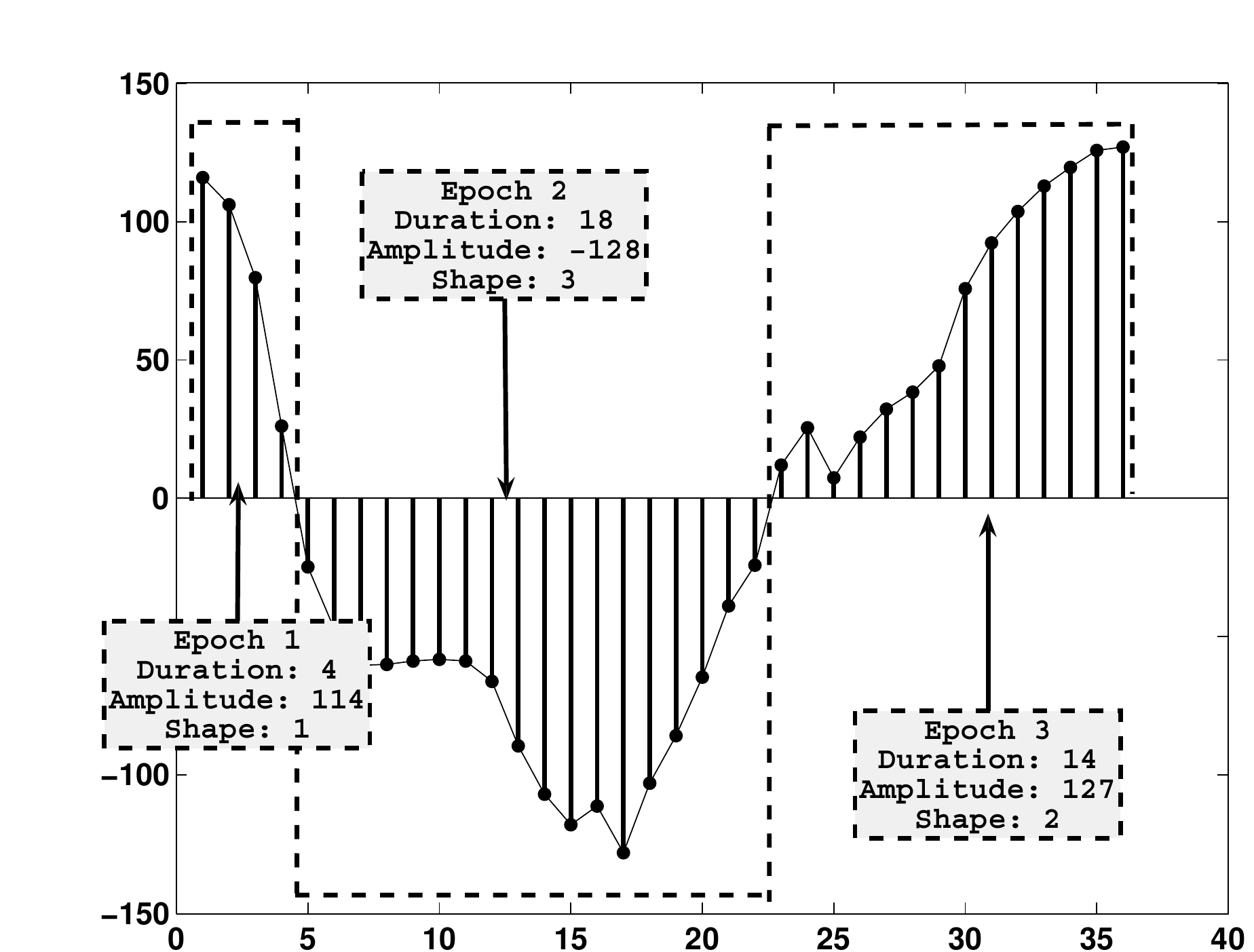}
    \caption{Example of 1D signal (vertical projection of an eye crop) and the associated
     encoding. There are three epochs, each encoded with three parameters. The associated code is:
    $[4,114,1;\enspace \enspace 18,-128,3; \enspace \enspace 14,127,2]$ } \label{Fig:ExTespar}
\end{figure}

To reduce the complexity (and computation time), the projections are compressed using a
zero-crossing based encoding technique. After ensuring that the projections values are  in a
symmetrical range with respect to zero, we will describe, independently, each interval between two
consecutive zero-crossings. Such an interval is called an \emph{epoch} and for its description
three parameters are considered (as presented in figure \ref{Fig:ExTespar}):
\begin{itemize}
    \item \emph{Duration} - the number of samples in the epoch;
    \item \emph{Amplitude} - the maximal signed deviation of the signal with respect to $0$;
    \item \emph{Shape} - the number of local extremes in the epoch.
\end{itemize}

The proposed encoding is similar with the TESPAR (Time-Encoded Signal Processing and Recognition)
technique \cite{King:99} that is used in the representation and recognition of 1D, band--limited,
speech signals. Depending on the problem specifics, additional parameters of the epochs may be
considered (e.g. the difference between the highest and the lowest mode from the given epoch).
Further extensions are at hand if an epoch is considered the approximation of a probability density
function and the extracted parameters are the statistical moments of the said distribution. In such
a case the \emph{shape} parameter corresponds to the number of modes of the distribution.

The reason for choosing this specific encoding is two-fold. First the determination of the
zero-crossings and the computation of the parameters is doable in a single pass through the target
1D signal, and, secondly, the epochs have specific meaning when describing the eye region, as
discussed in the next subsection.

Given an image sub-window, the ZEP feature is determined by the concatenation of four encoded
projections as described in the following:
\begin{enumerate}
    \item Compute both the integral and the edge projection functions ($P_H$, $P_V$, $E_H$, $E_V$);
    \item Independently \emph{normalize} each projection within a symmetrical interval. For instance,
    in our application we normalized each of the projections to the $[-128; 127]$ interval. This
    will normalize the amplitude of the projection;
    \item Encode each projection as described; allocate for each projection a maximum number of
        epochs;
    \item Normalize all other (i.e. duration and shape) encoding parameters;
    \item Form the final Zero-crossing based Encoded image Projections (ZEP) feature by concatenation
    of the encoded projections. Given an image rectangle, the ZEP feature consists of the epochs
    from all the 4 projections: ($P_H$, $P_V$, $E_H$, $E_V$).
\end{enumerate}

Image projections are simplified representations of the original image, each of them carrying
specific information; the encoding simplifies even more the image representation. The normalization
of the image projections, and thus of the epochs amplitudes, ensures independence of the ZEP
feature with respect to uniform variation of the illumination. The normalization with respect to
the number of elements in the image sub-window leads to partial scale invariance: horizontal
projections are invariant to stretching on the vertical direction and vice versa. The scale
invariance property of the ZEP feature is achieved by completely normalizing the encoded durations
to a specific range (e.g. the encoded horizontal projection becomes invariant to horizontal
stretching after duration normalization). We stress that when compared with previous methods based
on projections, which lack the normalization steps, the hereby proposed algorithm increases the
overall stability to various influences.

\begin{figure}
\center
\begin{tabular}{c}
    \begin{tabular}{cc}
        \includegraphics[width=0.15 \textwidth]{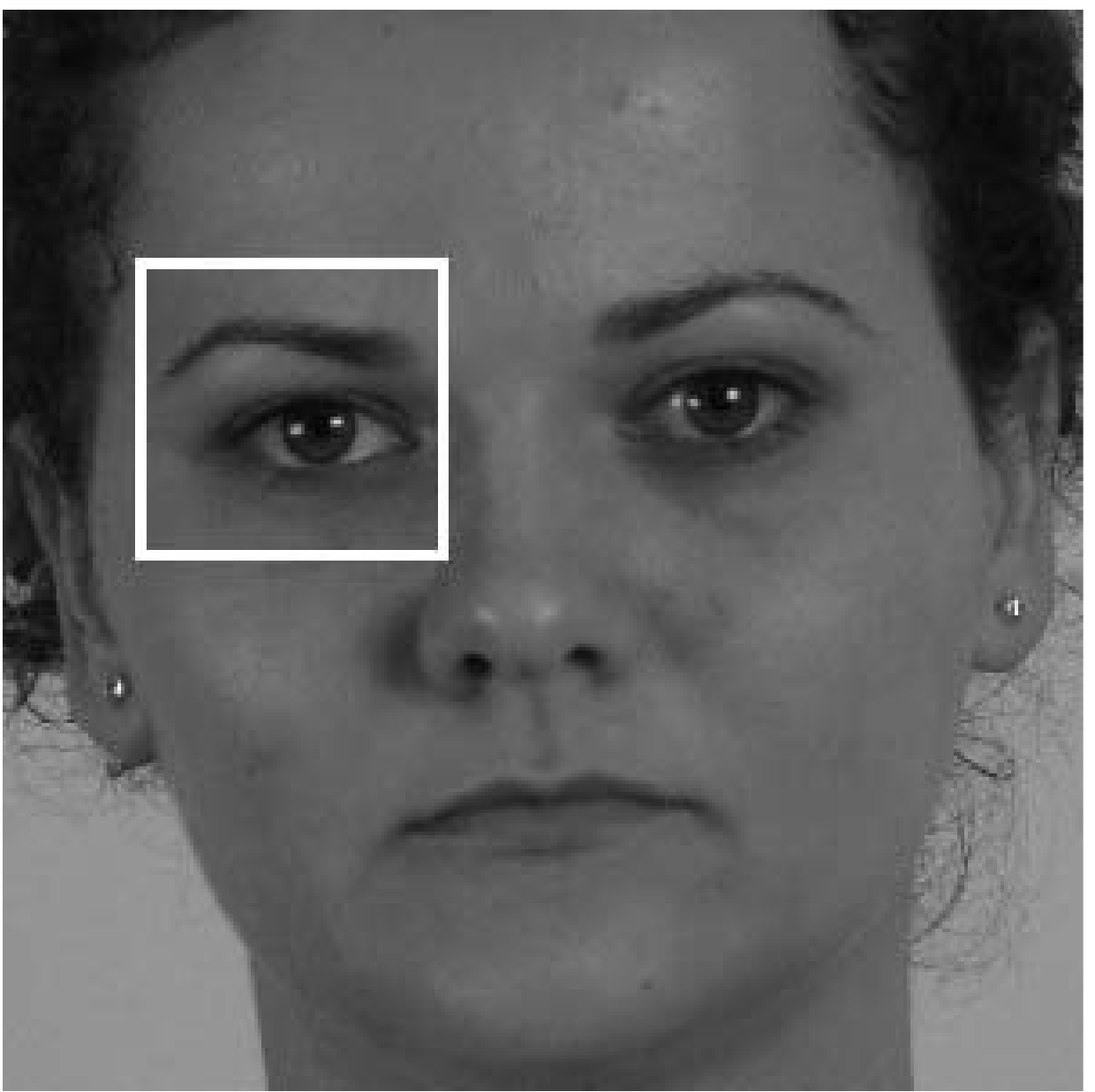} &
        \includegraphics[width=0.15 \textwidth]{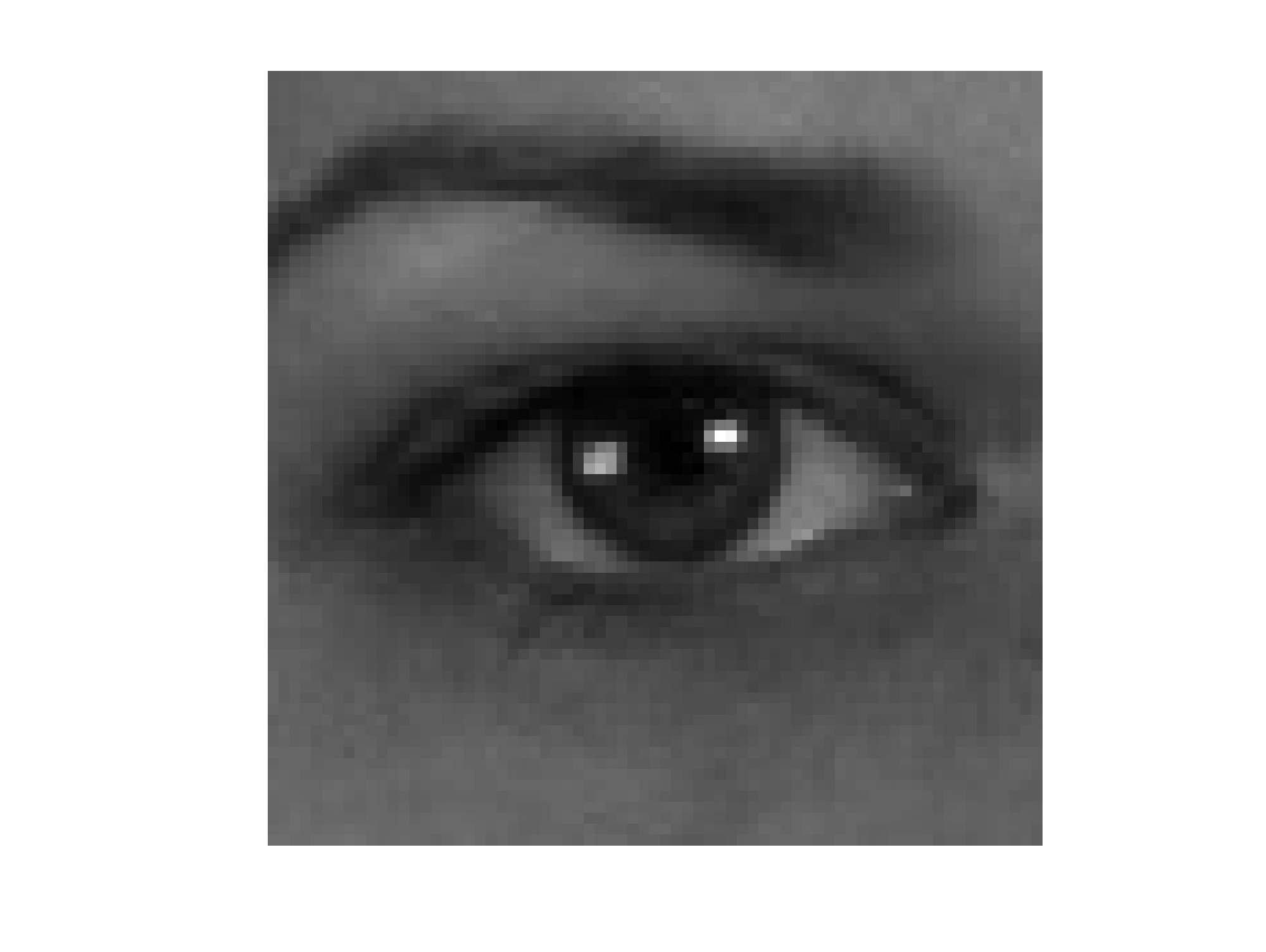} \\
        (a) & (b)
     \end{tabular} \\
     \begin{tabular} {cc}
           \includegraphics[width=0.44 \textwidth, height=0.30 \textwidth]{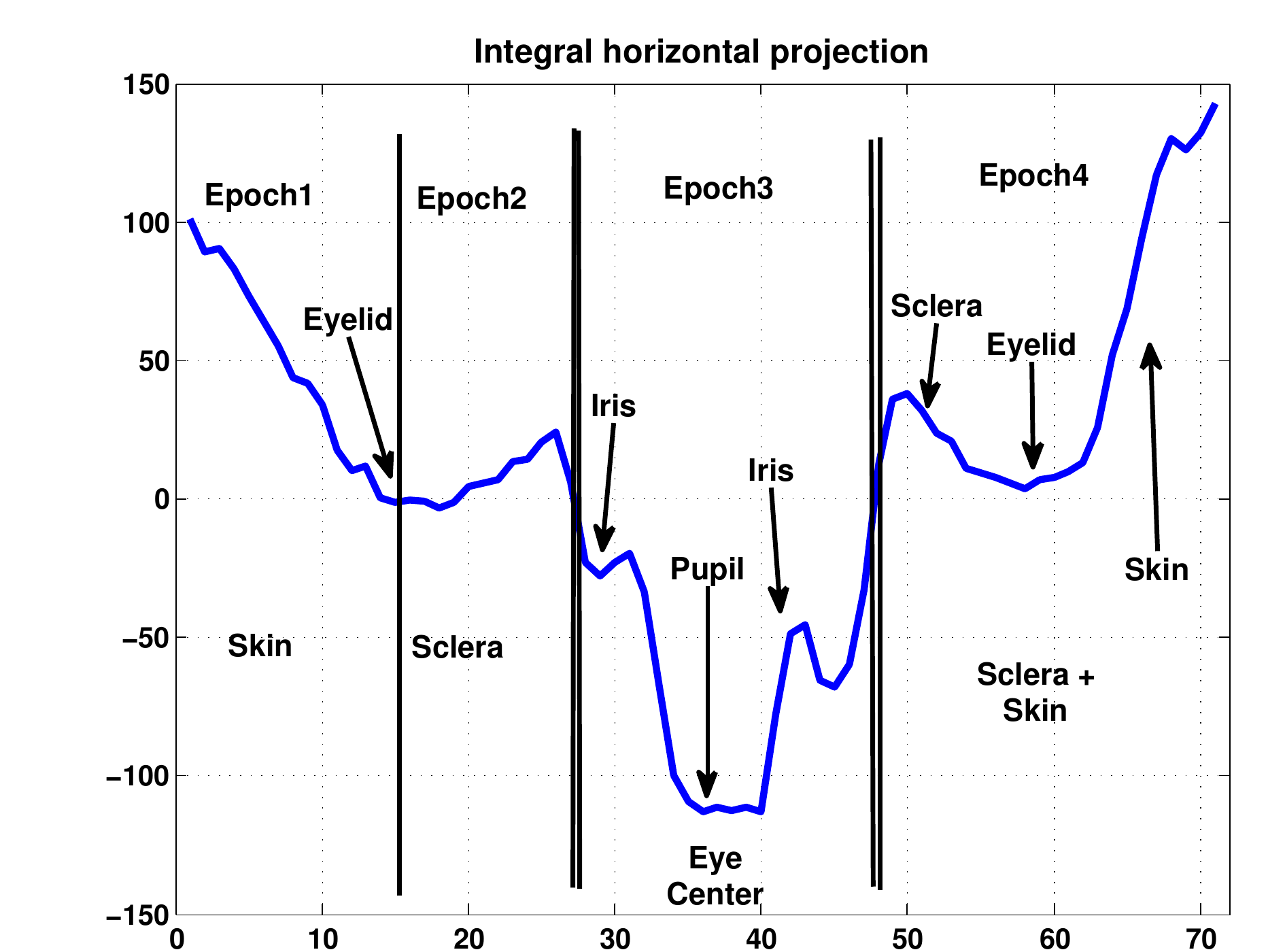} & (c) \\
           \includegraphics[width=0.44 \textwidth, height=0.30 \textwidth]{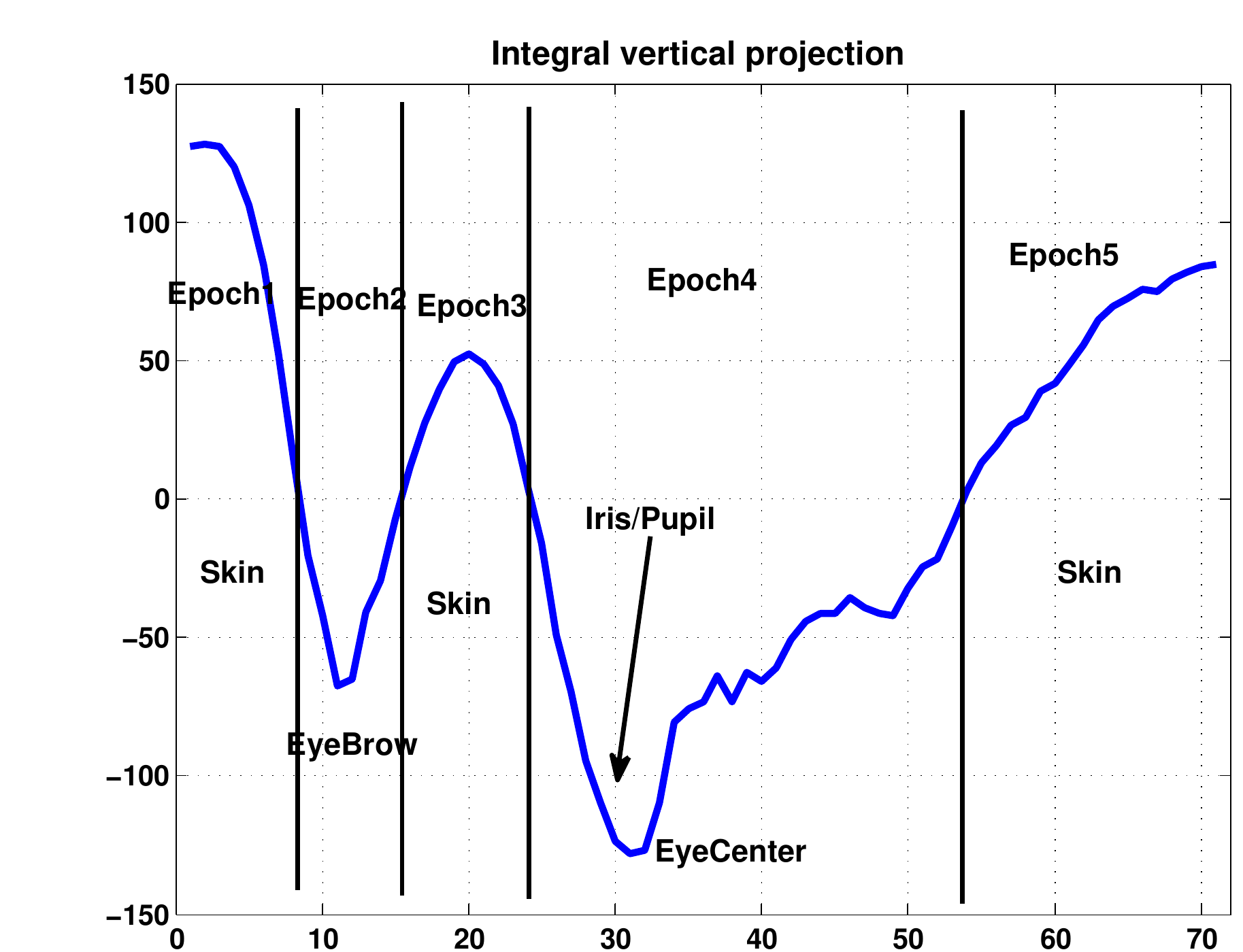} & (d) \\
    \end{tabular}
\end{tabular}
\caption{Image projections from a typical eye patch: (a) face crop, (b) eye crop,  (c) integral
    horizontal image projection on the eye crop, (d) integral vertical image projection on the eye crop. On the horizontal projection
    the double line marks the zero crossing that is found on all eye examples, while the rest of
    zero-crossings may be absent in some particular cases. }
\label{Fig:EyeRegZEP}
\end{figure}

\subsection{ZEP on Eye Localization}
\label{Subsect:ZepOnEye}

As noted, image projections have been used in multiple ways for the problem of eye localization. In
an exploratory work, \citeyear{Kanade:73} determined the potential of image projections for face
description. More recently, \citeyear{Feng:98}, \citeyear{Zhou:04} and \citeyear{Turkan:08}
presented the use of the integral projections and/or their extensions for the specific task of eye
localization. Especially in \cite{Zhou:04} it was noted that image projections, in the eye region
have a specific sequence of relative minima and maxima assigned with to skin (relative minimum),
sclera (relative maximum), iris (relative minimum),  etc.

Considering a rectangle from the eye region including the eyebrow (as showed in figure
\ref{Fig:EyeRegZEP} (a) ), the associated integral projections have specific epochs, as showed in
figure \ref{Fig:EyeRegZEP} (c) and (d). The particular succession of positive and negative modes is
precisely encoded by the proposed technique. On the horizontal integral projection there will be a
large (one-mode) epoch that is assigned to skin, followed by an epoch for sclera, a triple mode,
negative, epoch corresponding to the eye center and another positive epoch for the sclera and skin.
On the vertical integral projection, one expects a positive epoch above the eyebrow, followed by a
negative epoch on the eyebrow, a positive epoch between the eyebrow and eye, a negative epoch (with
three modes) on the eye and  a positive epoch below the eye.

The ZEP feature, due to invariance properties already discussed, achieves consistent performance
under various stresses and is able to discriminate among eyes (patches centered on pupil) and
non-eyes (patches centered on locations at a distance from the pupil center). As explained in
section \ref{Sect:Class}.2, on the validation set, using Fisher linear discriminant over 90\%
correct eye detection rate is achieved by selecting patches that are centered on the pupil with
respect to the ones that are shifted.

\begin{figure}[tb]
\center
     \includegraphics[width=0.45 \textwidth]{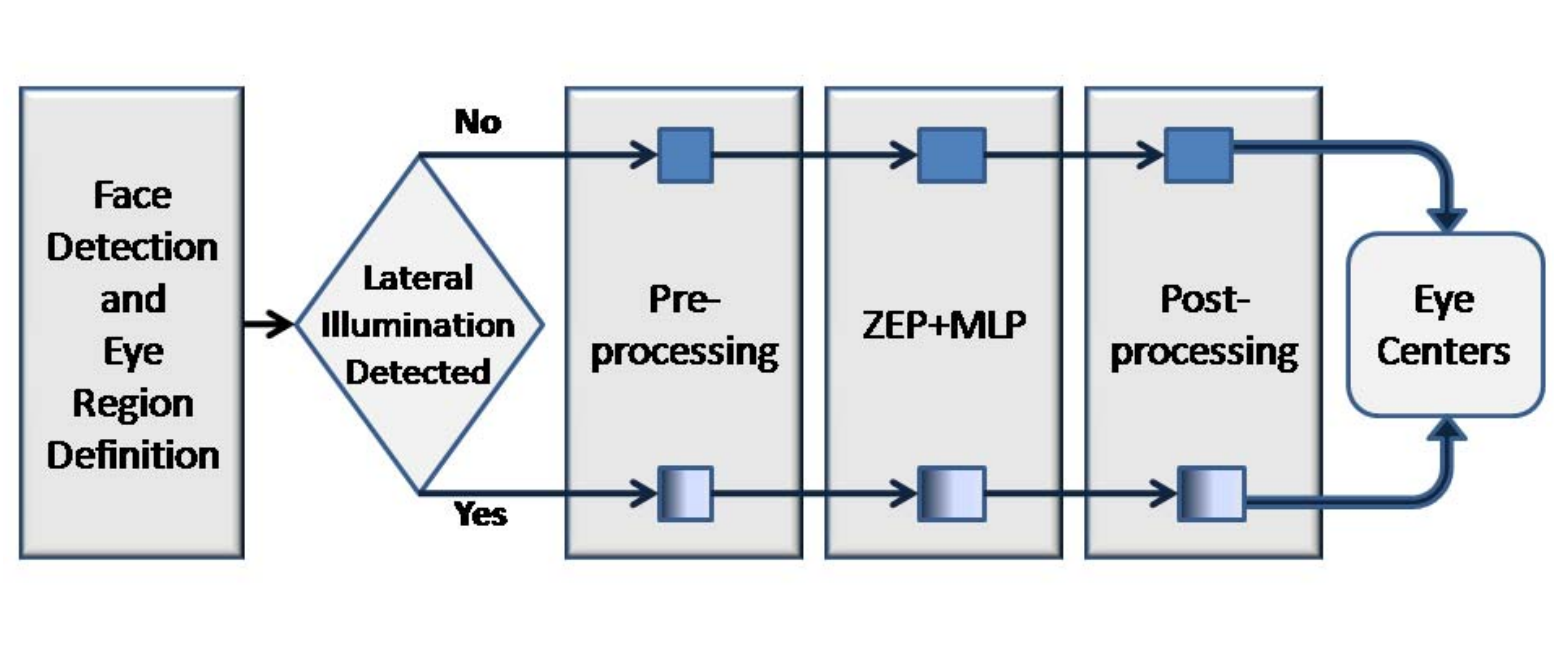}
\caption{The work flow of the proposed algorithm. } \label{Fig:Alg_schema}
\end{figure}

\section{Implementation}
\label{Sect:Class}

The block schematic of our eye center localization algorithm is summarized in figure
\ref{Fig:Alg_schema}. In the first step, a face detector (the cascade of Haar features
\cite{Viola:04} delivered with OpenCV) automatically determines the face square. Next the regions
of interest are set in the upper third of the detected face: from 26\% to 50\% of the face square
on rows, respectively from 25\% to 37\% on columns for the left eye and from 63\% to 75\% on
columns for the right eye.

Noting the susceptibility of the image projections to alter their shape due to lateral
illumination, we introduced a simple method for detecting such a case and we adapt the algorithm to
the type of illumination found. After a very simple preprocessing, the ZEP features for each
possible location are computed and feed  to a classifier to identify the possible eye locations.
The possible eyes are then post-processed and the best positions are located as discussed in
subsection \ref{Subsect:PrePostprocessing}.

Regarding the face detection, the recent solutions use multiple cascades for not only
identification of the face rectangle, but also for determination of the in-plane (roll) and yaw
(frontal/profile) angles of head. Such procedure follows Viola and Jones extension of the initial
face detector work \cite{Jones:03}, \cite{Ramirez:08}. Thus, it is customary to limit the analysis
of ``frontal faces'' to a maximum rotation of $30^0$.

\subsection{Lateral Illumination Detection}

To increase the solution robustness to lateral illumination, we automatically separate such cases.
The motivation for the split lies in the fact that side illumination significantly alters the shape
of the projections in the eye region, thus decreasing the performance of the classification part.

The lateral illumination detection relies on computing the average values on the eye patch
previously selected. The following ratios are considered:

\begin{equation}
    \begin{array}{ccc}
    L_{ratio} = \frac{ \overline{L_{top}} }{ \overline{L_{bot}} }; &
    R_{ratio} = \frac{ \overline{R_{top}} }{ \overline{R_{bot}} }; &
    H_{ratio} = \frac{ \overline{ L_{top}} + \overline{L_{bot}} }{ \overline{R_{top}} + \overline{R_{bot}}};
    \end{array}
\end{equation}
where ($\overline{L_{top}}$) and ($\overline{L_{bot}}$) are the average gray levels on the upper
and lower halves of the left eye and ($\overline{R_{top}}$ and $\overline{R_{bot}}$) are their
correspondents on the right eye.

The lateral illumination case is considered if any of the computed ratios, $L_{ratio}$,
$R_{ratio}$, $H_{ratio}$, is outside the $[0.5 ;  1.75]$ range.

We designed this block such that an illumination that do not produce significant shadows on the eye
region is detected as frontal, and as lateral otherwise. In terms of illumination angle, the cases
with shadows on the eye region imply an absolute value of azimuth angle higher than $40^0$ or an
elevation angle value higher than $10^0$. Negative elevation (light from below) with low azimuth
value does not produce shadows on the eye region. The interval $[0.5; 1.75]$ mentioned above  has
been found by matching the mentioned cases with the ratios values on the training database.

Indeed 98.54\% of the images from the BioID database are detected as frontal illuminated, while the
results on Extended Yale B are presented in table \ref{Tab:IllumDetectYaleB+}. Extended Yale B
database has images with various illumination angles as it will be discussed in section
\ref{Sect:Results}.5.

\begin{table}
    \center
    \caption{Percentage of frontal illumination detected cases on the Extended Yale B database.   \label{Tab:IllumDetectYaleB+} }
    \begin{tabular}{|c|c|c|}
        \hline
        \begin{tabular}{c}  Azimuth \\ \hline Elevation\\
         \end{tabular}
                          & $\pm [40^0:130^0]$  & $\pm[0^0 :  35^0]$  \\ \hline
         $-40^0 : 0^0 $   & { } 26.11\% { }       &{ } 88.28\% { }        \\ \hline
         $ 10^0 :  90^0 $ & { } 17.26\% { }       &{ } 36.35\% { }        \\ \hline
    \end{tabular}
\end{table}

\begin{figure*}[tb]
\center
    \begin{tabular}{ccc ccc}
        \includegraphics[width=0.12 \textwidth]{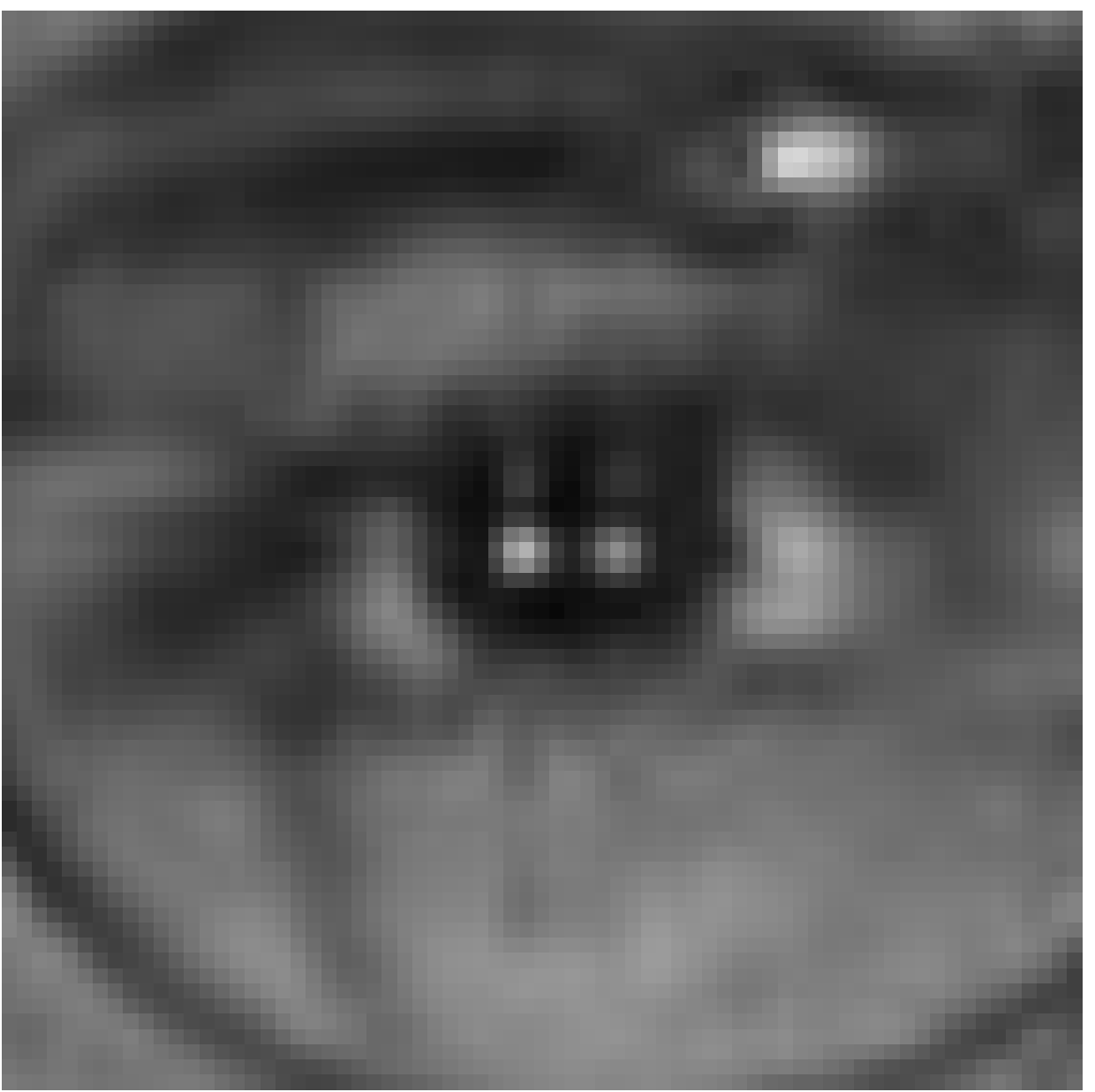} &
        \includegraphics[width=0.12 \textwidth]{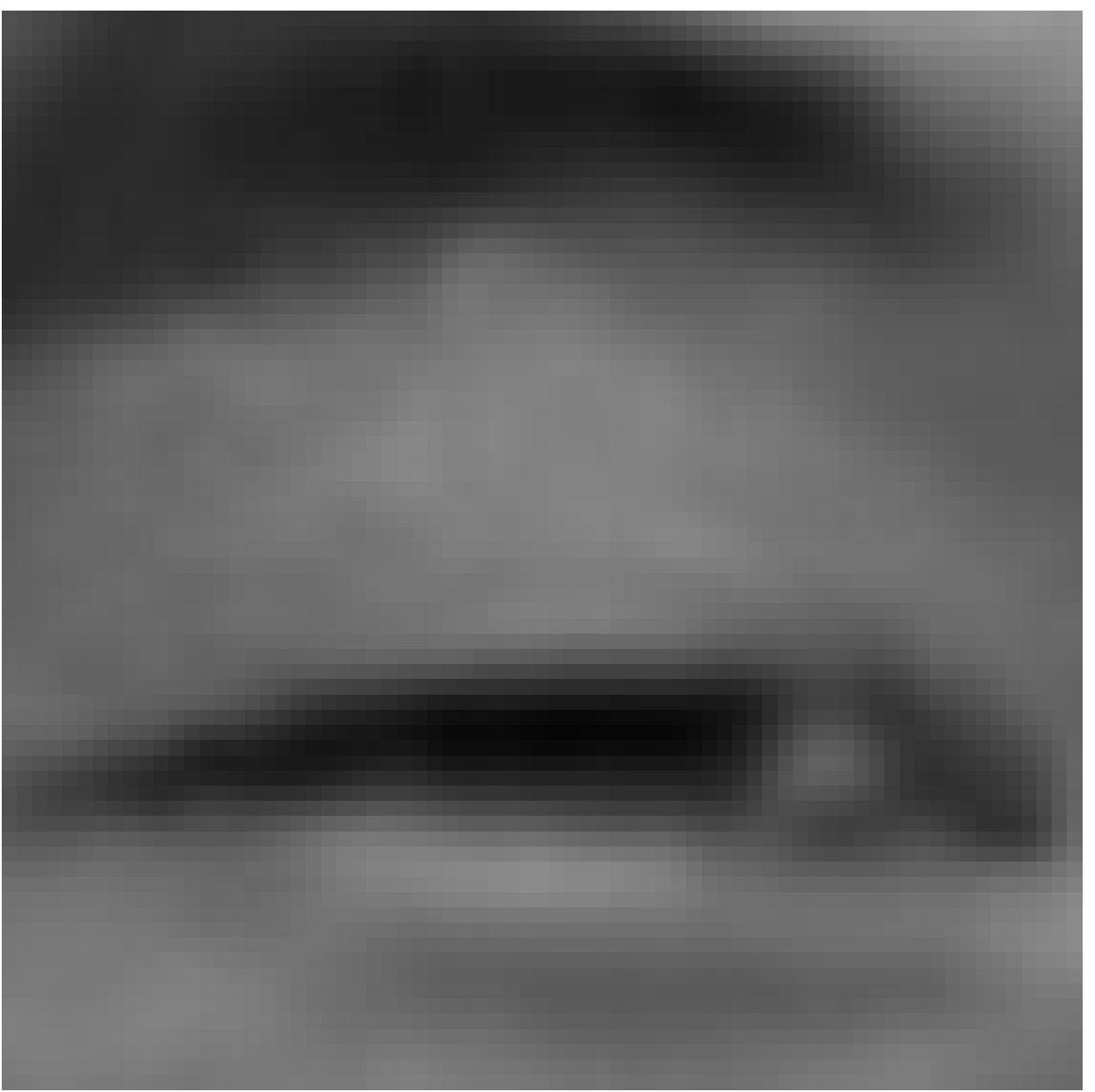} &
        \includegraphics[width=0.12 \textwidth]{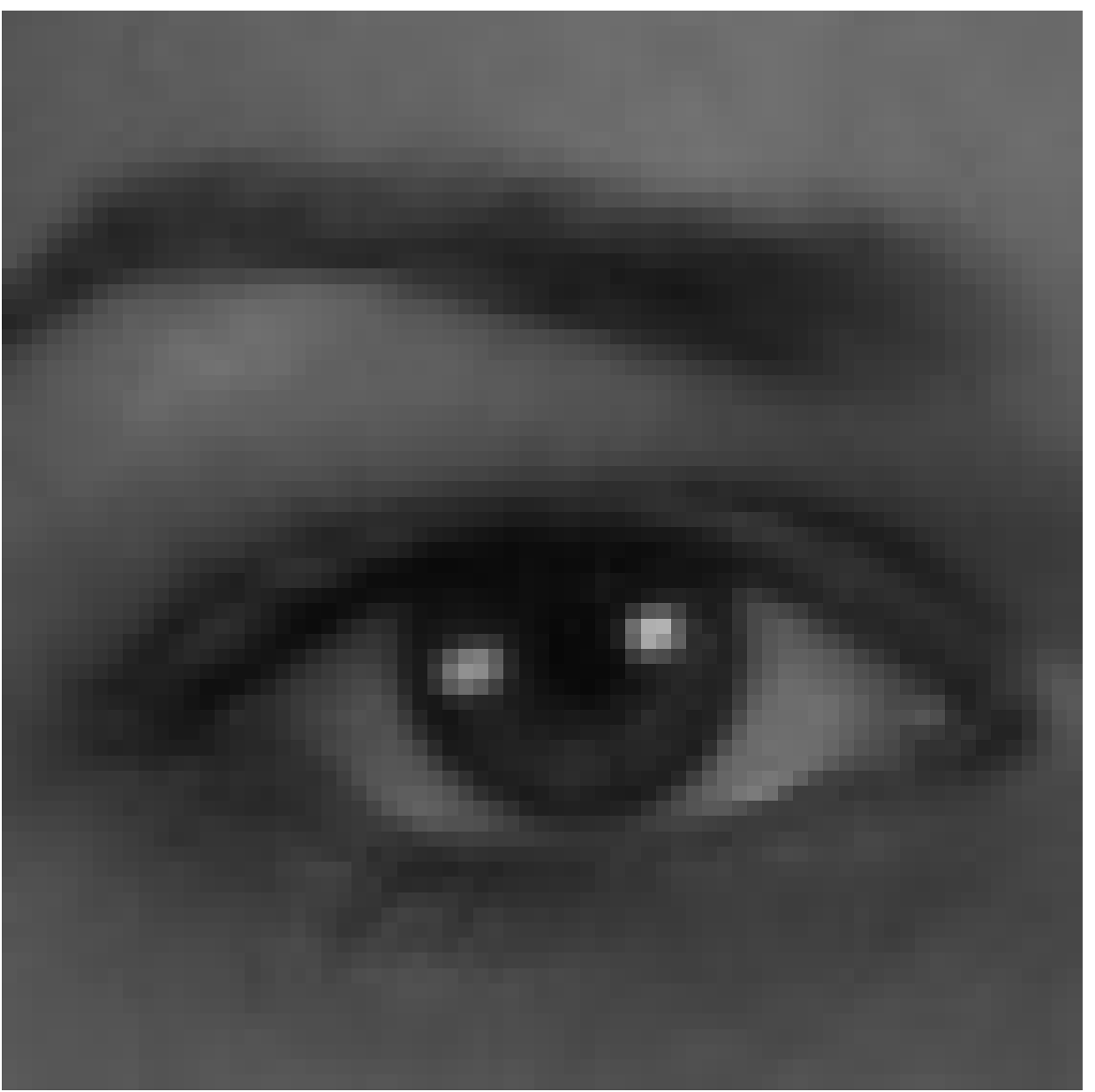} &

        \includegraphics[width=0.12 \textwidth]{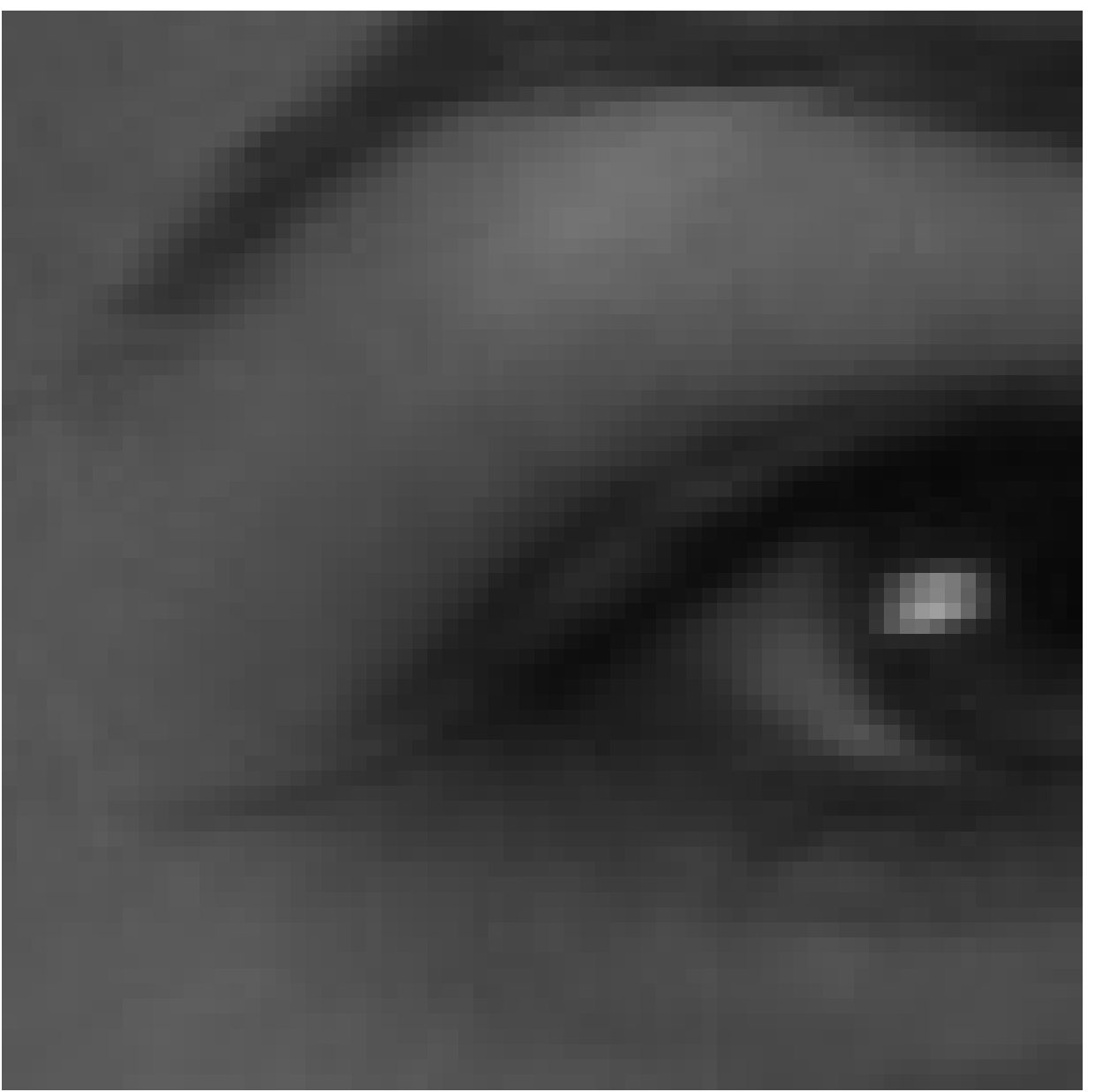} &
        \includegraphics[width=0.12 \textwidth]{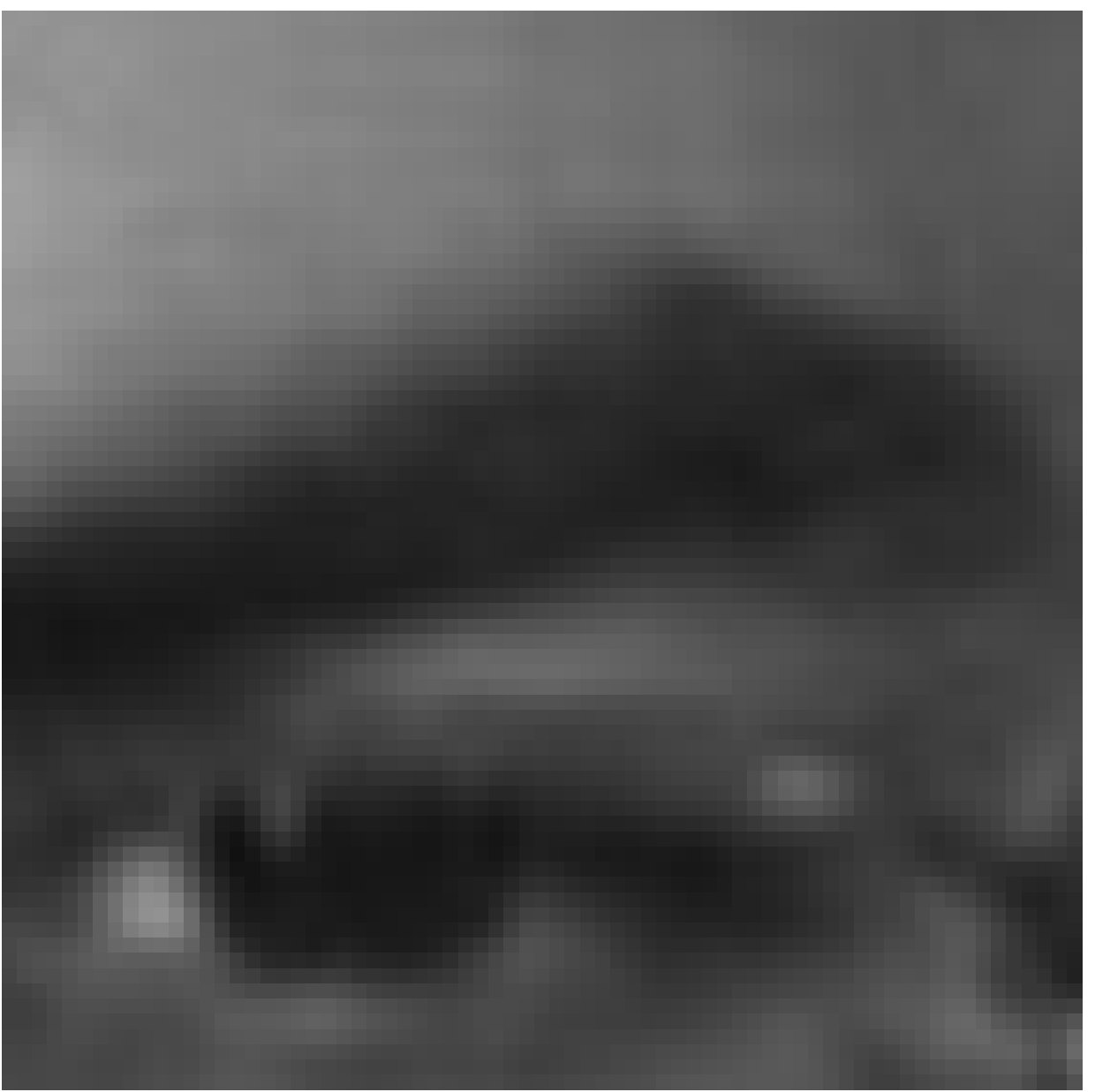} &
        \includegraphics[width=0.12 \textwidth]{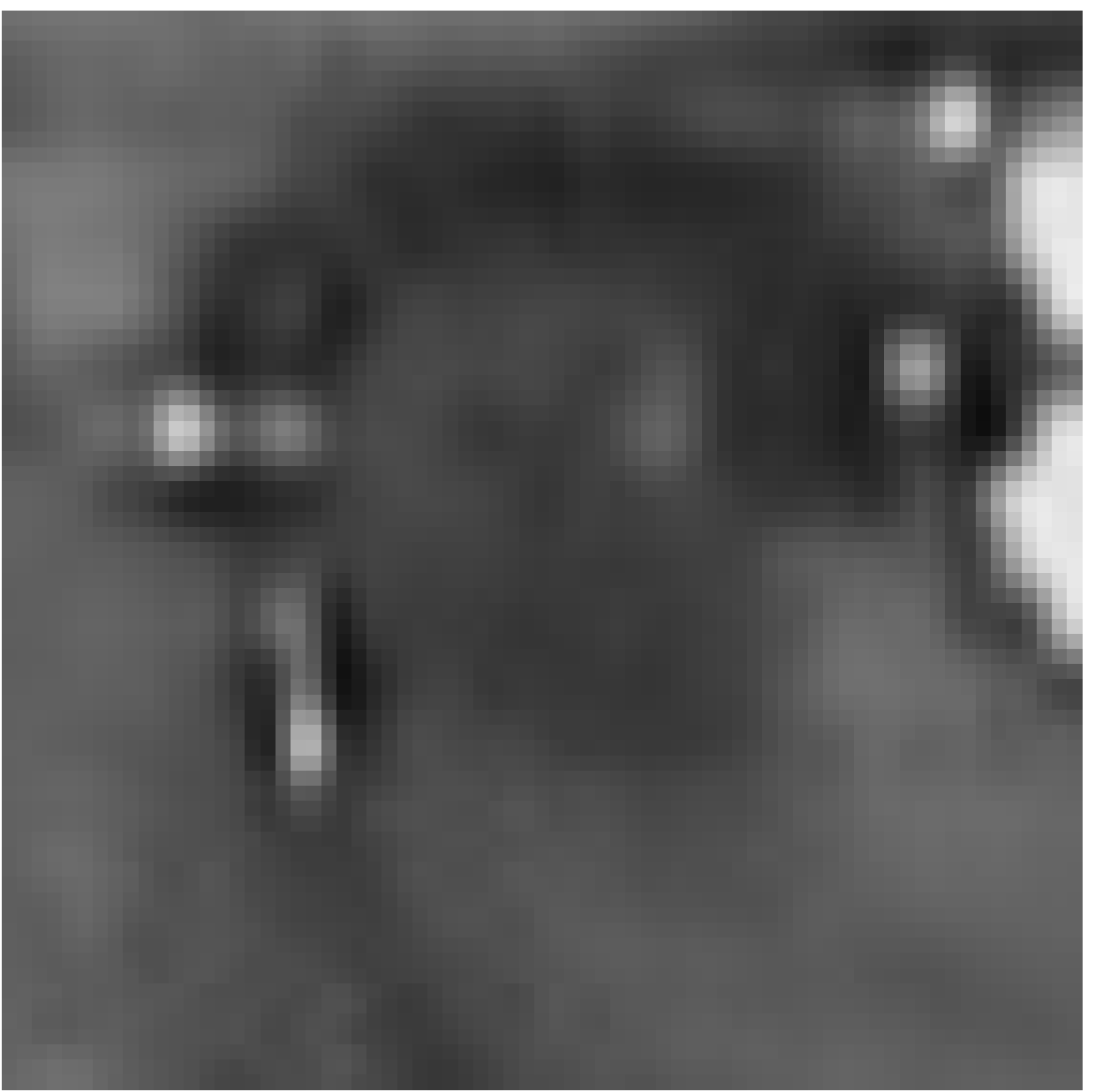}
    \end{tabular}
    \caption{Examples of eye (first three images) and non-eye (last three images) crops; it should
    be noted that the examples include eye expressions and glasses} \label{Fig:Eye_Train}
\end{figure*}

\subsection{Training}

Once the ZEP features are determined,  the extracted data is feed into a  Multi-Layer Perceptron
(MLP), having one input layer, one hidden layer and one output layer, trained with the
back-propagation algorithm. In our implementation, the number of neurons from the hidden layer is
chosen to be half the size of the ZEP feature as it was empirically determined as a reasonable
trade-off between performance (higher number of hidden neurons) and speed.

In the preferred implementation, each projection is encoded with 5 epochs, leading to 60 elements
in the ZEP feature (and 60 inputs to the MLP). If more epochs are provided by projections (which is
very unlikely for eye localization - less than 0.1\% in the tested cases), the last ones are simply
removed.

\begin{figure}[tb]
\center
    \begin{tabular}{cc}
        \includegraphics[width=0.20 \textwidth]{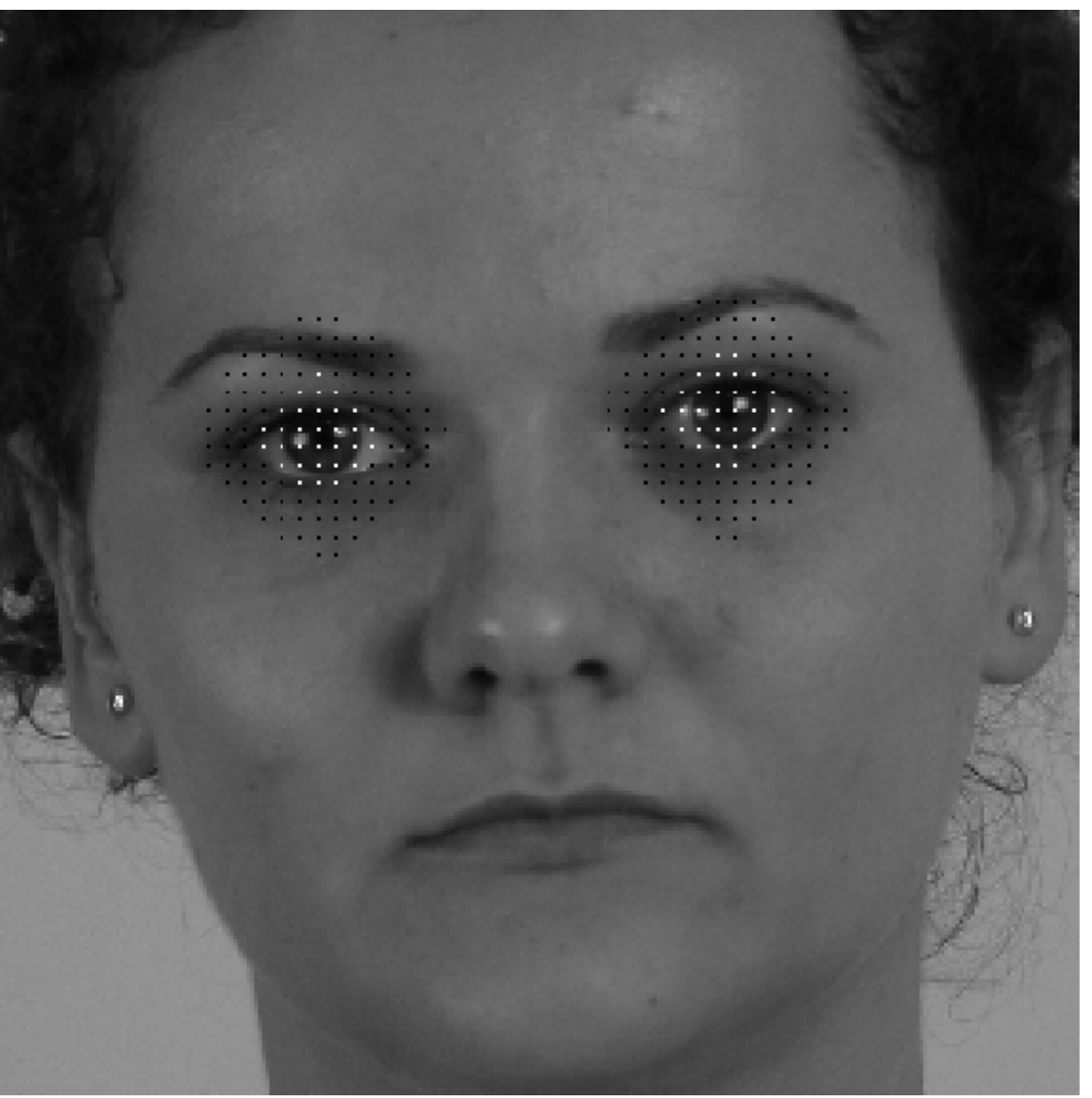} &
        \includegraphics[width=0.20 \textwidth]{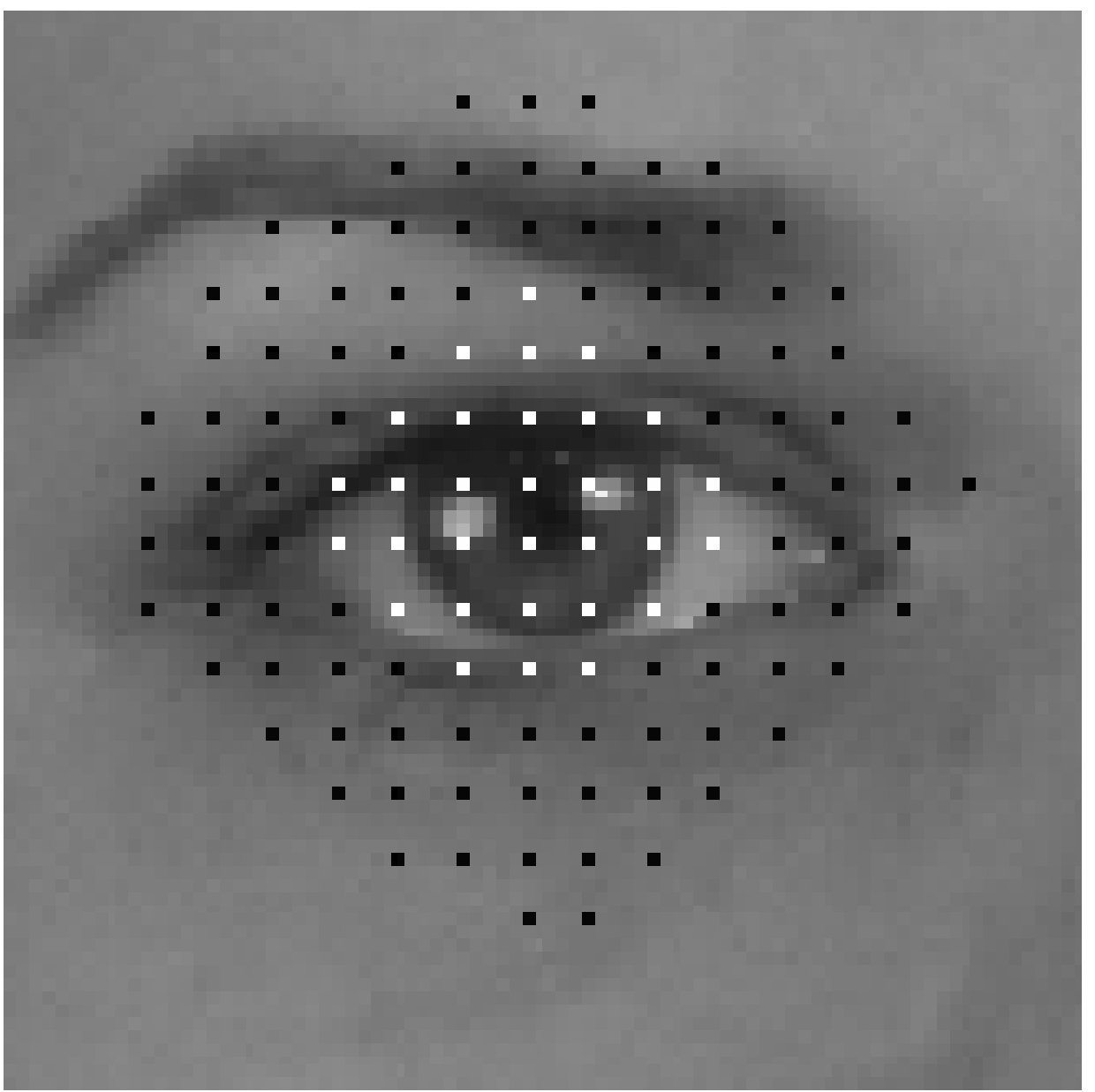}
    \end{tabular}
    \caption{Localization of the image patch centers used as positive
    examples (white) and negative examples (black). The right hand
    image is zoomed from the left hand one.} \label{Fig:EyeExamples}
\end{figure}

The training of the MLP is performed with crops of eyes and non-eyes of $71 \times 71$ pixels, as
shown in figure \ref{Fig:Eye_Train}, while the preferred face size is $300 \times 300$ pixels. The
positive examples are taken near the eye ground truth: the eye rectangle overlaps more than 75\%
with the true eye rectangle. The patches corresponding to the negative examples overlap with the
true eye between 50\% and 75\%, thus leading to a total of 25 positive examples and 100 negative
ones from each eye  in a single face image. Positive and negative locations are showed in figure
\ref{Fig:EyeExamples}. This specific choice of positive and negative examples yields to high
performance in localization.

In total there were 10,000 positive examples and as much negative ones, taken from the authors' Eye
Chimera database, from the Georgia Tech database \cite{Nefian:00} and from the neutral poses
selected from the YaleB database \cite{Georghiades:01}. We have considered two training variants
corresponding to the two types of illumination (frontal or lateral).

One training procedure uses images from our data set (40\%) and from Georgia Tech (60\%) and
focuses on frontal  illumination, eye expressions and occlusions. In this case, the MLP is trained
to return the L2 distance from a specific patch center to the true eye center. Thus the MLP performs
regression.

The second training procedure (for lateral illumination) uses images only from the frontal pose of
the Yale B database and it is used for improving performance against illumination. In this case the
training set was labelled with $-1$ (non-eye -- between 50\% and 75\% overlapping onto the centered
eye patch) or $1$ (eye -- more than 75\% overlapping).

As many machine learning algorithms are available, we have performed a short study on examples
extracted from the training databases. Given the number of images in the databases, 20,000 examples
were used for training the networks and approximately  200,000 were used for classifier validation.
For the classification problem, a Support Vector Machine (SVM) produced 93.7\% correct detection
rate, the used MLP 92.6\% and an ensemble of 50 bagged decision trees 91.5\% detection rate. For
the regression case, a SVM for Regression lead to an approximation error of 0.090, the regression
MLP 0.096 and bagged ensemble of regression trees only 0.115. Taking into account the achieved
values, there is no significant performance difference among the various machine learning systems
tested (conclusion which matches the findings from \cite{Everingham:06}), thus our decision on
using the MLP was based more on speed issues.

\subsection{Preprocessing and Postprocessing}
\label{Subsect:PrePostprocessing}

The conceptual steps in both illumination cases of the actual eye localization procedure are the same:
preprocessing, machine learning and postprocessing.

\begin{table}
\centering
    \caption{Details of the two merged solutions of the actual eye localization \label{Tab:AlgDetails}}
    \begin{tabular} {|p{2.0cm}|p{2.05cm}|p{2.75cm}|}
    \hline
        \textbf{ Case} & \begin{tabular}{l}\textbf{Frontal}\\ \textbf{illumination}\end{tabular} &
                    \begin{tabular}{l}\textbf{Lateral}\\ \textbf{illumination}\end{tabular} \\ \hline
        \begin{tabular}{l} Darkness \\ preprocess. \\threshold  \end{tabular} &
            \begin{tabular}{l}$0.15$\end{tabular} &
             \begin{tabular}{l}$0.3$\end{tabular}  \\ \hline
        \begin{tabular}{l} Training \\database  \end{tabular} &
            \begin{tabular}{l} GeorgiaTech \\ + Authors' \end{tabular} &
             \begin{tabular}{l} Yale B \end{tabular} \\ \hline
        \begin{tabular}{l} Training \\scheme  \end{tabular} &
            \begin{tabular}{l} Regression\\ L2 dist \end{tabular} &
            \begin{tabular}{l} True/False \end{tabular}  \\ \hline
        \begin{tabular}{l} ZEP+MLP \\ threshold  \end{tabular} &
             \begin{tabular}{l}$0$\end{tabular} &
              \begin{tabular}{l}$-0.5$\end{tabular}  \\ \hline
        \begin{tabular}{l} eye area \\ selection  \end{tabular} &
             \begin{tabular}{l} largest \\ lower region  \end{tabular}  &
             \begin{tabular}{l} largest region \end{tabular} \\ \hline
        \begin{tabular}{l} eye center \\ localization  \end{tabular} &
            \begin{tabular}{l} weighted \\ center of\\ mass \end{tabular}&
            \begin{tabular}{l} geometrical center \\ of the  rectangle\\ circumscribed to \\the eye region   \end{tabular} \\ \hline
    \end{tabular}
\end{table}

A simple preprocessing is applied for each eye candidate region to accelerate the localization
process. Following \citeyear{Wu:03}, we note that the eye center (associated with the pupil) is
significantly darker than the surrounding; thus the pixels that are too bright with respect to the
eye region (and are not plausible to be eye centers) are discarded. The ``too bright''
characteristic is encoded as gray--levels higher than a percentage (so called darkness
preprocessing threshold in table \ref{Tab:AlgDetails}) from the maximum value of the eye region. In
the lateral illumination case, this threshold is higher due to the deep shadows that can be found
on the skin area surrounding the eye.

\begin{figure}[tb]
\center
    \begin{tabular}{cc cc}
        \includegraphics[width=0.10 \textwidth]{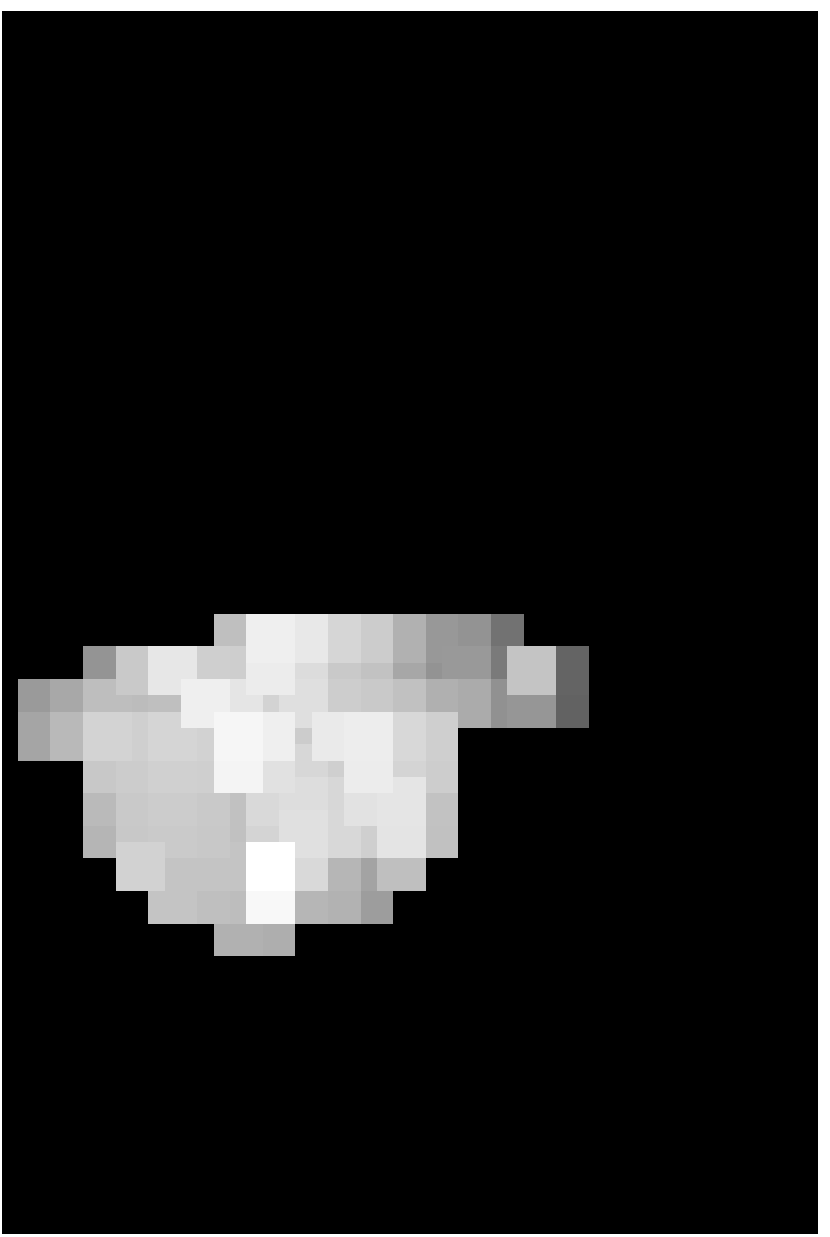} &
        \includegraphics[width=0.10 \textwidth]{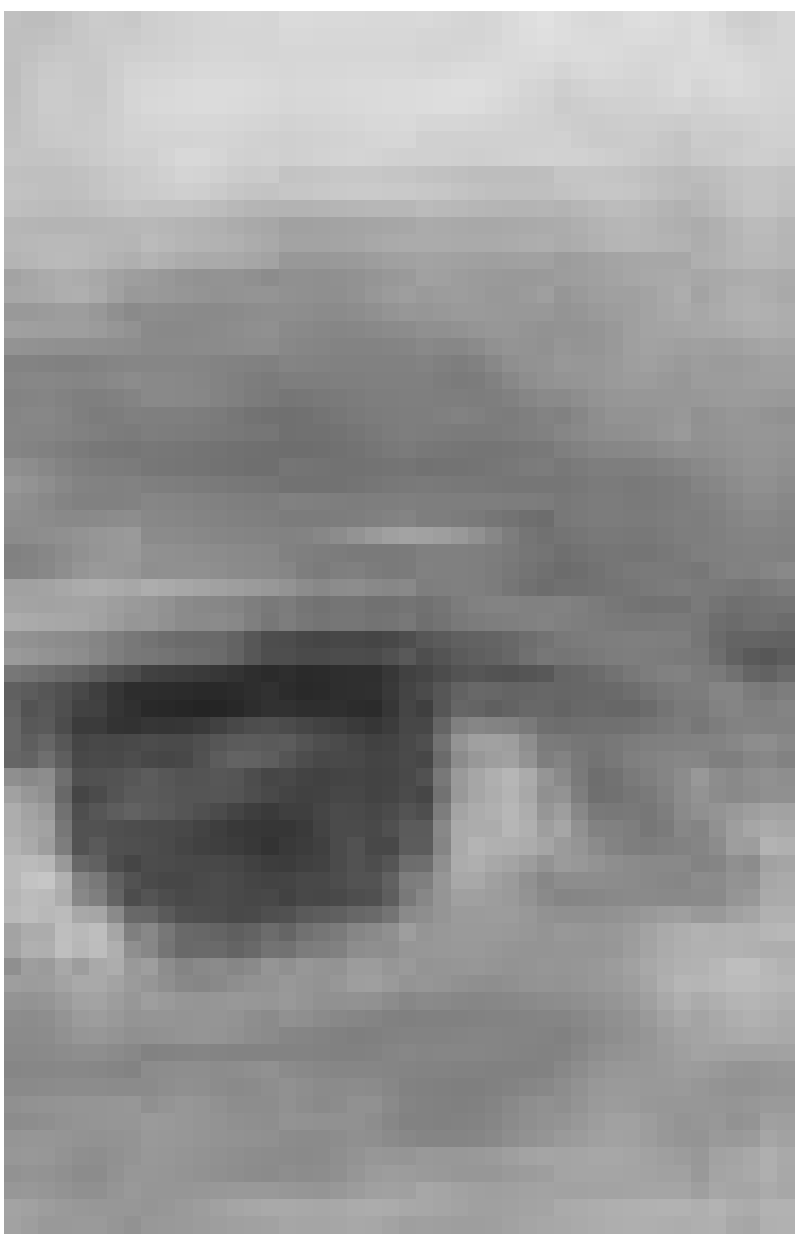} &
        \includegraphics[width=0.10 \textwidth]{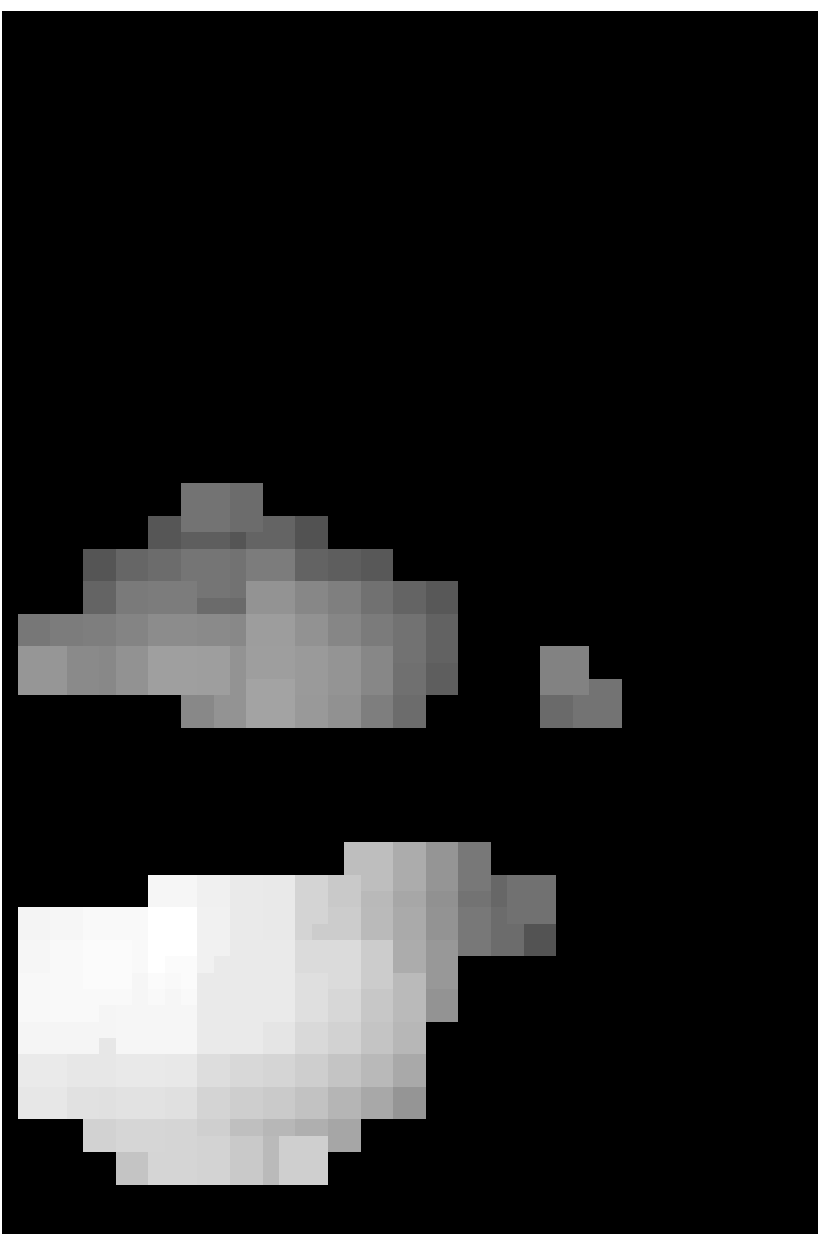} &
       \includegraphics[width=0.10 \textwidth]{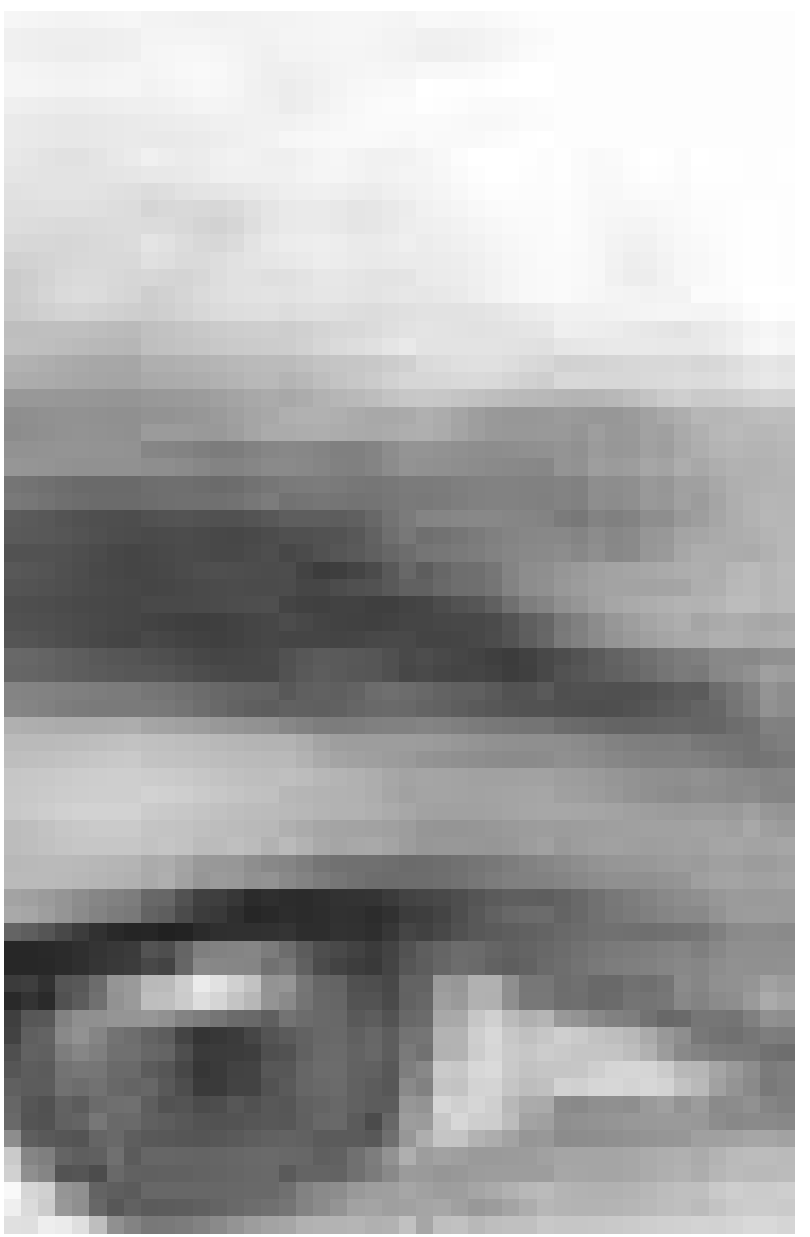}\\
                (a) & (b) &    (c) & (d) \\
    \end{tabular}
    \caption{The ZEP image ((a) and (c)) resulting after processing with ZEP+MLP (so called ZEP image) the original
        eye patch image (b), (d) (under frontal illumination case). The first two images (a), (b) show a simple case
        (a) when only patches around eye center were selected by the MLP. The right hand two images,
        (c) and (d), contain a more difficult case (c) where also the eyebrow was identified as
        possible eye region. Higher gray--levels in the ZEP image signal higher confidence, as estimated by MLP,
        in having the eye center at that specific location. } \label{Fig:ZEPImage}
\end{figure}

In the area of interest, using a step of 2 over a sliding image patch of $71\times71$ pixels, we
investigate by the proposed ZEP+MLP all the plausible locations. We consider as positive results
the  locations where the value given by the MLP is higher than an experimentally found threshold
(see table \ref{Tab:AlgDetails}). These positive results are recorded in a separate image (the ZEP
image, shown in figure \ref{Fig:ZEPImage}) which is further post--processed for eye center
extraction.

Since closed eyes (that were included in the training set) are similar with eyebrows, one may get
false eye regions given by the eyebrow in the ZEP image. Thus the ZEP image is segmented, labelled
and the lowest and largest regions are associated with the eye. This step will discard, for
instance, the regions given by the eyebrow in figure \ref{Fig:ZEPImage} (c).

For the frontal illumination case, due to training with L2 distance as objective, one expects a
symmetrical shape around the true eye center. Thus the final eye location is taken as the weighted
center of mass of the previously selected eye regions. For the lateral illumination, the binary
trained MLP is supposed to localize the area surrounding the eye center and the final eye center is
the geometrical center of the  rectangle circumscribed to the selected region. We note that in both
cases, the specific way of selecting the final eye center is able to deal with holes (caused by
reflections or glasses) in the eye region.

An overview of how each step is implemented in the two illumination cases considered is shown in
table \ref{Tab:AlgDetails}.


\section{Results and Discussions}
\label{Sect:Results}

We will discuss first the influence of various system parameters onto the overall results. For this
purpose we will use the BioID database\footnote{\url{http://www.bioid.com/downloads/software/
bioid-face-database.html}}. This database contains  1521 gray-scale, frontal facial images of size
$384 \times 286$, acquired with frontal illumination conditions in a complex background. The
database contains 16 tilted and rotated faces, people that wear eye-glasses and, in very few cases,
people that have their eyes shut or pose various expressions. The database was released with
annotations for iris centers. Being one of the first databases that provided facial annotations,
BioID became the most used database for face landmarks localization accuracy tests, even that it
provides limited variability and reduced resemblance with real-life cases. We will use BioID as a
starting point in discussing the achieved results (for giving an inside on the system's various
parameters and selecting the most performing state of the art systems) so that later to continue
the evaluation under other stresses like eye expression, illumination angle or pose. Yet the most
relevant test is on real-life cases, which are acquired in the Labelled Faces in Wild database that
will be presented later on.

The localization performance is evaluated according to the stringent localization criterion
\cite{Jesorsky:01}. The eyes are considered to be correctly determined if the specific localization
error $\epsilon$, defined in equation (\ref{Eq:StringCrit}) is smaller than a predefined value.

\begin{equation}
    \label{Eq:StringCrit}
    \epsilon = \frac{\max\{\varepsilon_L, \varepsilon_R \} }{D_{eye}}
\end{equation}

In the equation above, $\varepsilon_L$ is the Euclidean distance between the ground truth left  eye
center and determined left eye center, $\varepsilon_R$ is the corresponding value for the right
eye, while $D_{eye}$ is the distance between the ground truth eyes centers. Typical error
thresholds are $\epsilon= 0.05$ corresponding to eyes centers found inside the true pupils,
$\epsilon= 0.1$ corresponding to eyes centers found inside the true irises, and $\epsilon= 0.25$
corresponding to eyes centers found inside the true sclera. This criterion identifies  the
worst case scenario.

We note that, while the BioID image size leads to approximately a $ 40 \times 40 $ size for the eye
patch, because our target are HD video frames (for which we will also provide duration), we upscale
the face square to $300\times 300$, thus having an eye square of $71\times 71$.

\begin{figure*}[tb]
\center
    \begin{tabular}{cc}
        \includegraphics[width=0.35 \textwidth]{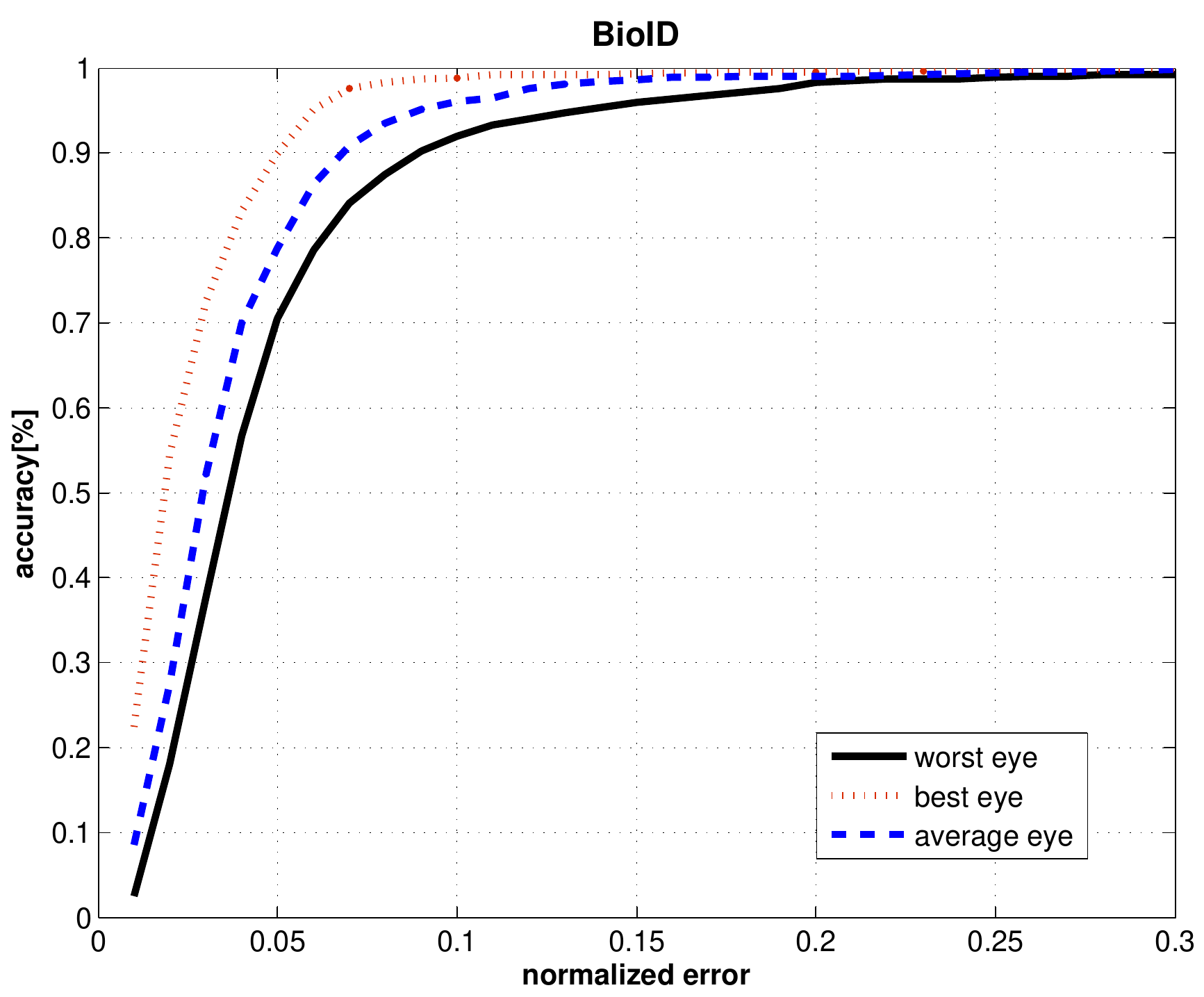}&
        \includegraphics[width=0.35 \textwidth]{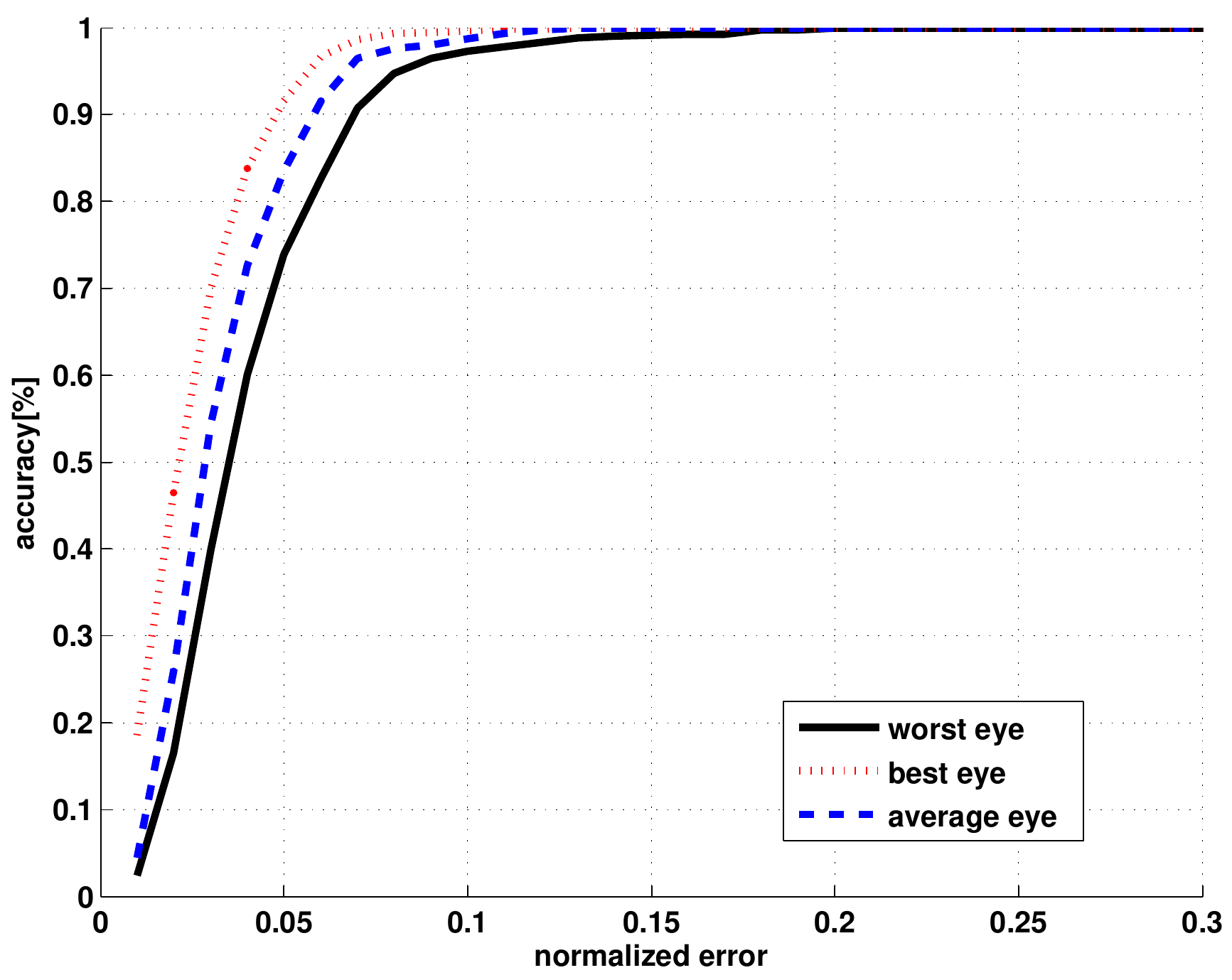}\\
        (a) & (b)
    \end{tabular}
    \caption{Maximum (best case - red dotted line), average (blue dashed line) and minimum  (worst
    eye - black solid line) accuracy for eye localization on the BioID database (left - (a)) and
    Cohn-Kanade database (right - (b)). } \label{Fig:BioID_Results}
\end{figure*}

The results on the BioID database are shown in figure \ref{Fig:BioID_Results} (a), where we
represented the maximum (better localized eye), average and minimum (worst localized eye) accuracy
with respect to various values of the $\varepsilon$ threshold.

\subsection{The Influence of ZEP Parameters}

We investigated the performance of the proposed system when only one type of projection is used.
The results are presented in table \ref{Tab:Rez_onlyIPF}. The computation time dropped to $\approx
53\%$ of the full algorithm time if only one projection type is used. The performance drops with
$\approx 14\%$ in the case of integral projection and with $\approx 17\%$ in the case of edge
projections. Using the proposed encoding it is possible to keep both the dimensionality of the
feature and the time duration low enough in order to use more than one type of projection. This
supplementary information helps to increase the results accuracy when compared with the method in
\cite{Turkan:08}.

\begin{table}[tb]
 \centering
    \caption{Percentage of correct eye localization of the proposed algorithm (integral -- IPF
     and edge projection functions -- EPF) compared with IPF-only and EPF-only implementations}
    \label{Tab:Rez_onlyIPF}

    \begin{tabular}{| c | c | c | c |}
    \hline
        \textbf{Projection Type} & \emph{IPF+EPF} & IPF & EPF \\ \hline
        Accuracy,  $\epsilon < 0.05$ &{ } \colorbox{lightgray}{70.46} { }&{ }  56.61 {  }&{ }  53.17 {  }\\ \hline
        Accuracy,  $\epsilon < 0.10$ &{ } \colorbox{lightgray}{91.94} { }&{ }  87.58 {  }&{ }  84.02  {  }\\ \hline
        Accuracy,  $\epsilon < 0.25$ &{ } \colorbox{lightgray}{98.87} { }&{ }  98.81 {  }&{ }  96.50  {  }\\ \hline
    \end{tabular}

\end{table}

Alternatives to the eye crop size and resulting values are presented in table
\ref{Tab:Rez_EyeSize}. The experiment was performed by re-training the MLP with eye crops of the
target size. As one can notice, the results are similar, thus proving the scale invariance of the
ZEP feature. Slight variation is due to the pre- and post processing.

\begin{table}[tb]
\centering
 \caption{The variation of the results with respect to the size of the eye analysis window.  } \label{Tab:Rez_EyeSize}

\begin{tabular}{| c | c | c | c | }
  \hline
    \textbf{Crop size}            & $36\times36$ & $71\times71$         & $100\times100$
     \\ \hline
     Accuracy,  $\epsilon < 0.05$ & { } 64.00 { } & { } \colorbox{lightgray}{70.46} { }&{ } 64.20 { }\\ \hline
     Accuracy,  $\epsilon < 0.10$ & { } 90.36 { } & { } \colorbox{lightgray}{91.94} { }&{ } 90.22 { }\\ \hline
     Accuracy,  $\epsilon < 0.25$ & { } 98.61 { } & { } \colorbox{lightgray}{98.87} { }&{ } 99.14 { }\\ \hline
\end{tabular}

\end{table}


\subsection{The Dimensionality Reduction }
\label{Subsect:DimReduct}

The main visible effect of the proposed encoding is the reduction of the size of the concatenated
projections. Yet, as we have adapted the encoding technique to the specific of the projection
functions applied on the eye area, its performance is higher than of other methods. To see the
influence of this encoding technique, we compared the achieved results with the ones obtained by
reducing the dimensionality with PCA (as being the most foreknown such technique) by the same
amount as the proposed one. The rest of the algorithm remains the same. The comparative results may
be seen in table \ref{Tab:Rez_PCA}. We also report the results when no reduction was performed.


The results indicate that both methods are lossy compression techniques and lead to decreased
accuracy. The proposed method is able to extract the specifics of the eye from the image
projections, as discussed in subsection \ref{Subsect:ZepOnEye}, being marginally better then the
PCA compression.

Furthermore, we take into account that the dimensionality reduction with PCA requires, for each
considered location, a matrix multiplication to project the initial vector of size $N_p$
($N_p=284$) onto the final space (with size $M_p=60$), thus having the complexity
$\mathcal{O}(N_pM_p) = \mathcal{O}(284\times 60)$. In comparison, the determination of the epochs
parameters is done in a single cross of the initial vector (i.e. with complexity $\mathcal{O}(N_p)
= \mathcal{O}(284)$), thus we expect the proposed method to be significantly  faster.

Indeed, the average value for computation time increases from 6 msec (using the proposed method) to
11 msec (almost double) using PCA on a $300 \times 300$ face square. The lack of compression
increases the duration to 24 msec per face square.

\begin{table}[tb]
 \centering
    \caption{Percentage of correct eye localization of the proposed algorithm (ZEP encoding)
    compared with dimensionality reduction with PCA.}
    \label{Tab:Rez_PCA}

    \begin{tabular}{| c | c | c | c | c |}
    \hline
              \textbf{Encoding Type} & \emph{Proposed}         & PCA     & None\\ \hline
        Accuracy,  $\epsilon < 0.05$ &{ } \colorbox{lightgray}{70.46} { }&{ }  69.66 { }&{ }  72.97 { }\\ \hline
        Accuracy,  $\epsilon < 0.10$ &{ } \colorbox{lightgray}{91.94} { }&{ }  92.70 { }&{ }  93.52 { }\\ \hline
        Accuracy,  $\epsilon < 0.25$ &{ } \colorbox{lightgray}{98.89} { }&{ }  98.87 { }&{ }  99.07 { }\\ \hline
    \end{tabular}
\end{table}


\subsection{Robustness to Noise}
\begin{figure}[tb]
\center
        \includegraphics[width=5cm]{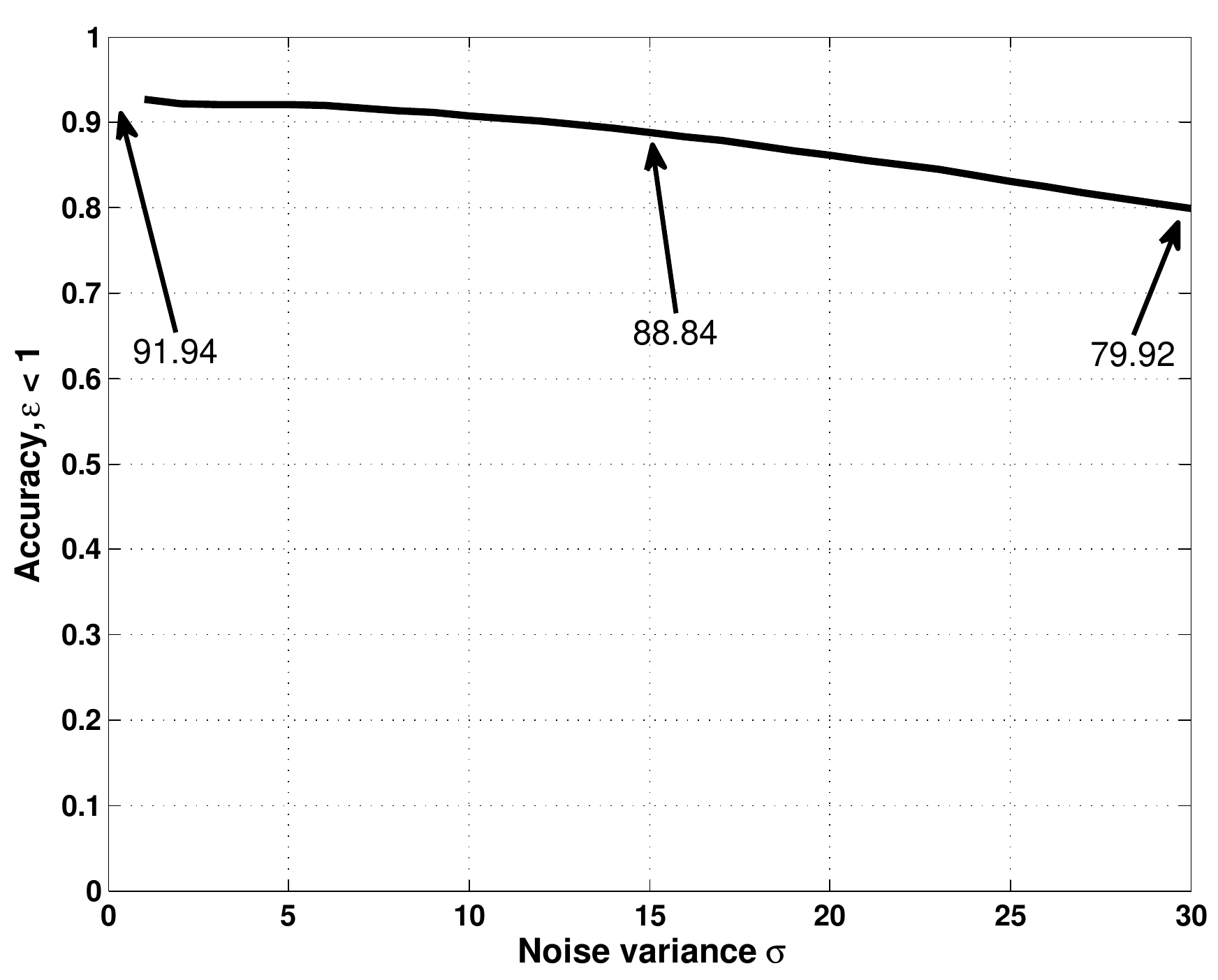}
        \caption{Variation of the localization performance for $\epsilon < 0.1$ accuracy with
        respect to gaussian noise variance used for image degradation. }
   \label{Fig:NoiseEye}
\end{figure}

\begin{figure}[tb]
\center
    \begin{tabular}{ccc }
        \includegraphics[width=0.14\textwidth]{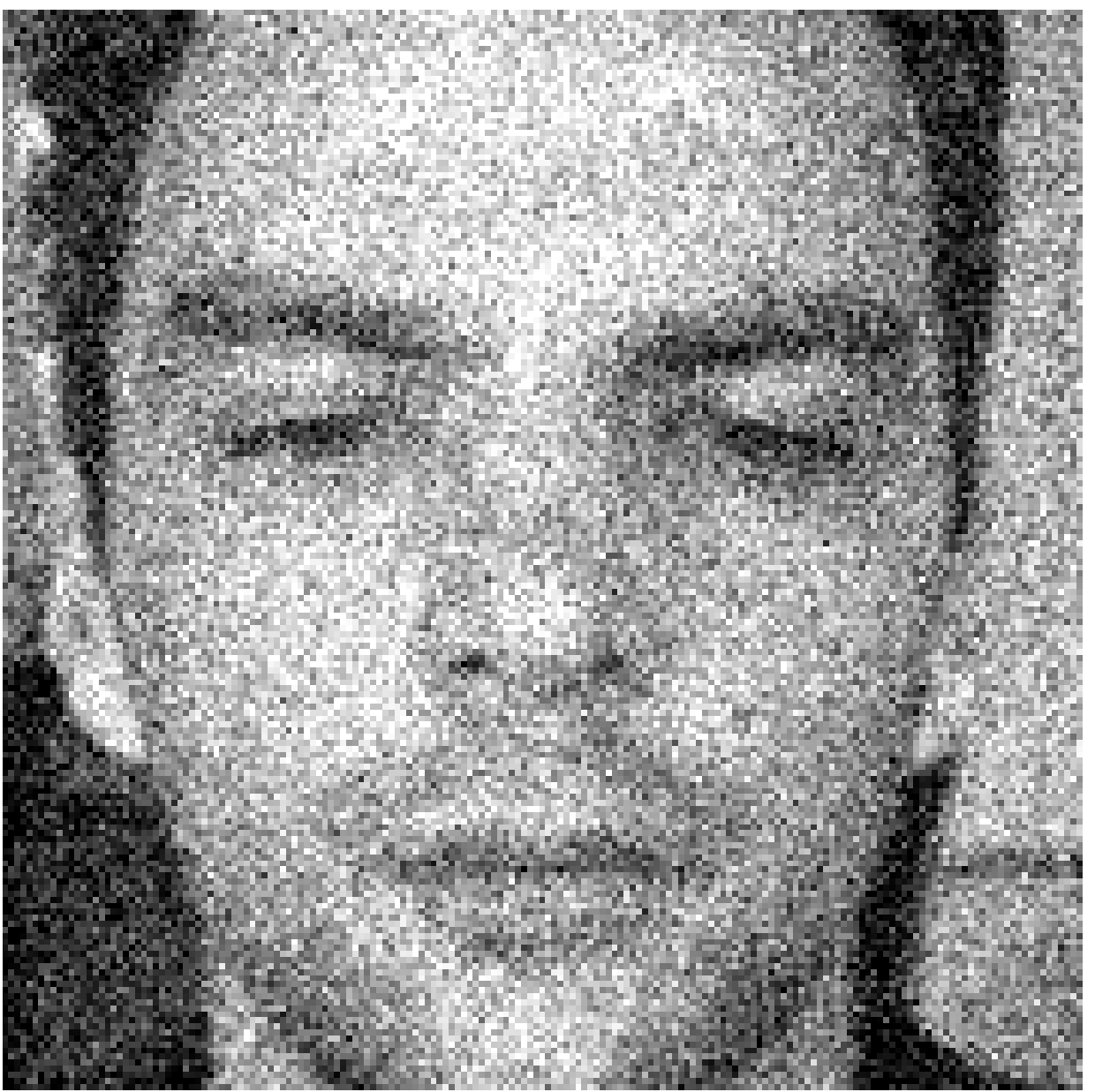}&
        \includegraphics[width=0.14\textwidth]{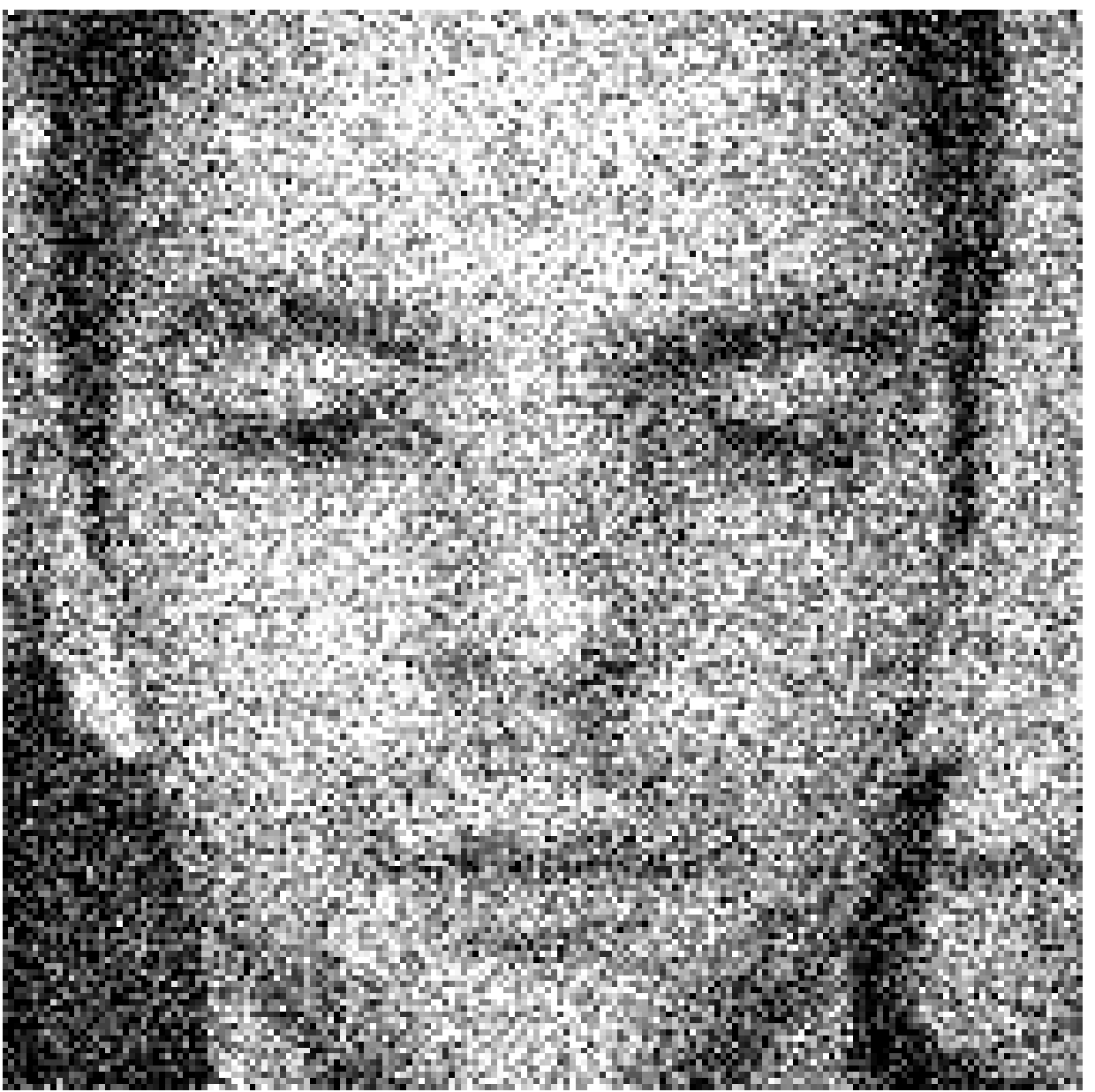}&
        \includegraphics[width=0.14\textwidth]{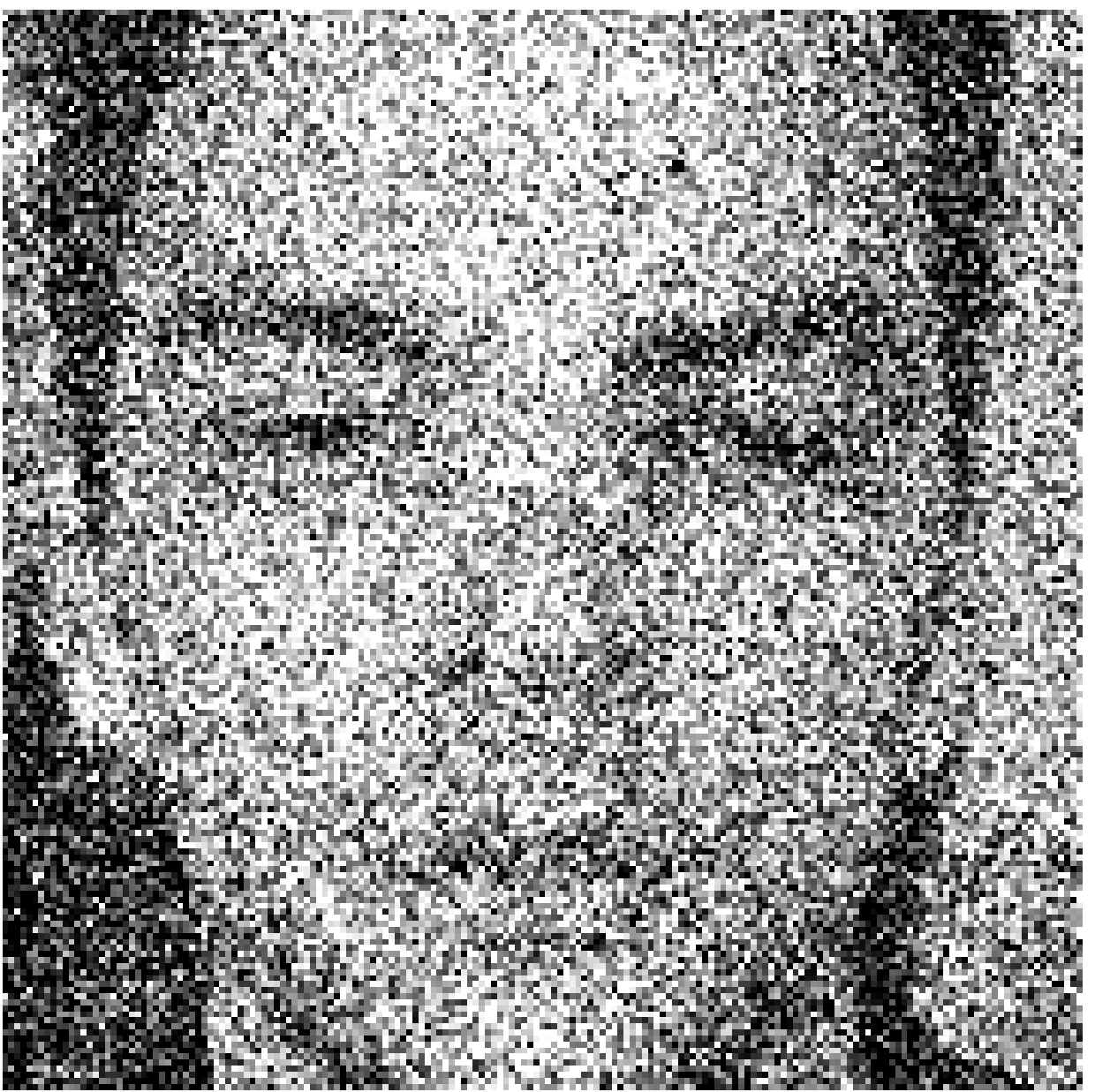}\\
        $\sigma = 5$ & $\sigma = 15$ & $\sigma = 30$
    \end{tabular}
   \caption{Image from the BioID database degraded by noise. For a human observer it is quite difficult
        to determine the eye centers. Yet the results showed in figure \ref{Fig:NoiseEye}
        prove that our method performs remarkably well.}
   \label{Fig:NoiseEyeExamples}
\end{figure}

An image projection represents a gray-scale average, hence it is reasonable to expect that the
proposed method is very robust to noise. To study robustness to noise we have artificially added
Gaussian noise to the BioID images and we subsequently measured the localization performance for
$\epsilon < 0.1$ accuracy. Indeed, while the noise variance increases from 0 to 30, the average
accuracy decreases from $91.94\%$ to only $79.92\%$. The variation of the accuracy with respect to
the added noise standard deviation  may be seen in figure \ref{Fig:NoiseEye}. Examples of images
degraded by noise may be seen in figure \ref{Fig:NoiseEyeExamples}.

\subsection{Results on the BioID }

\begin{figure}
\center
    \begin{tabular}{ccc }
        \includegraphics[width=0.14 \textwidth]{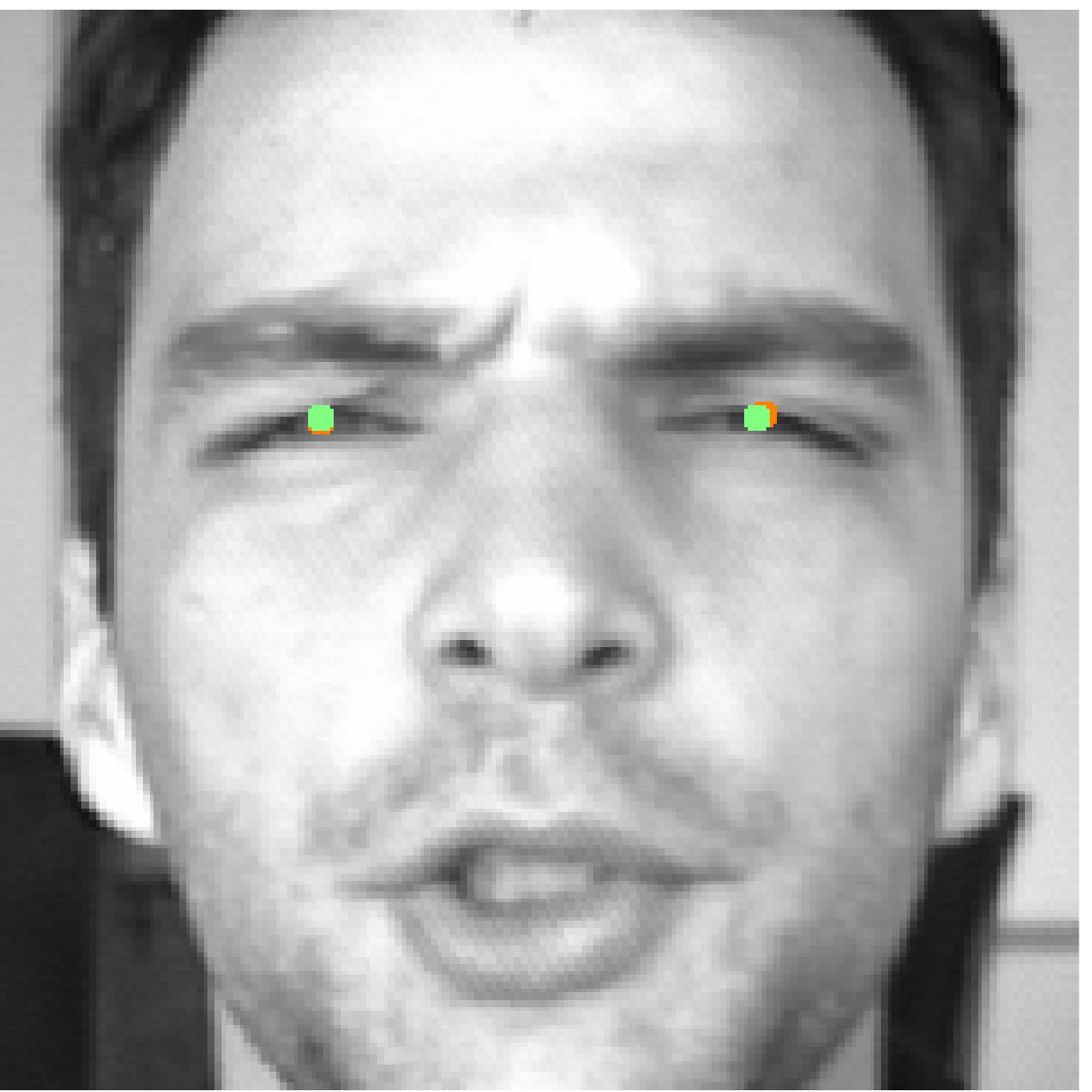}&
        \includegraphics[width=0.14 \textwidth]{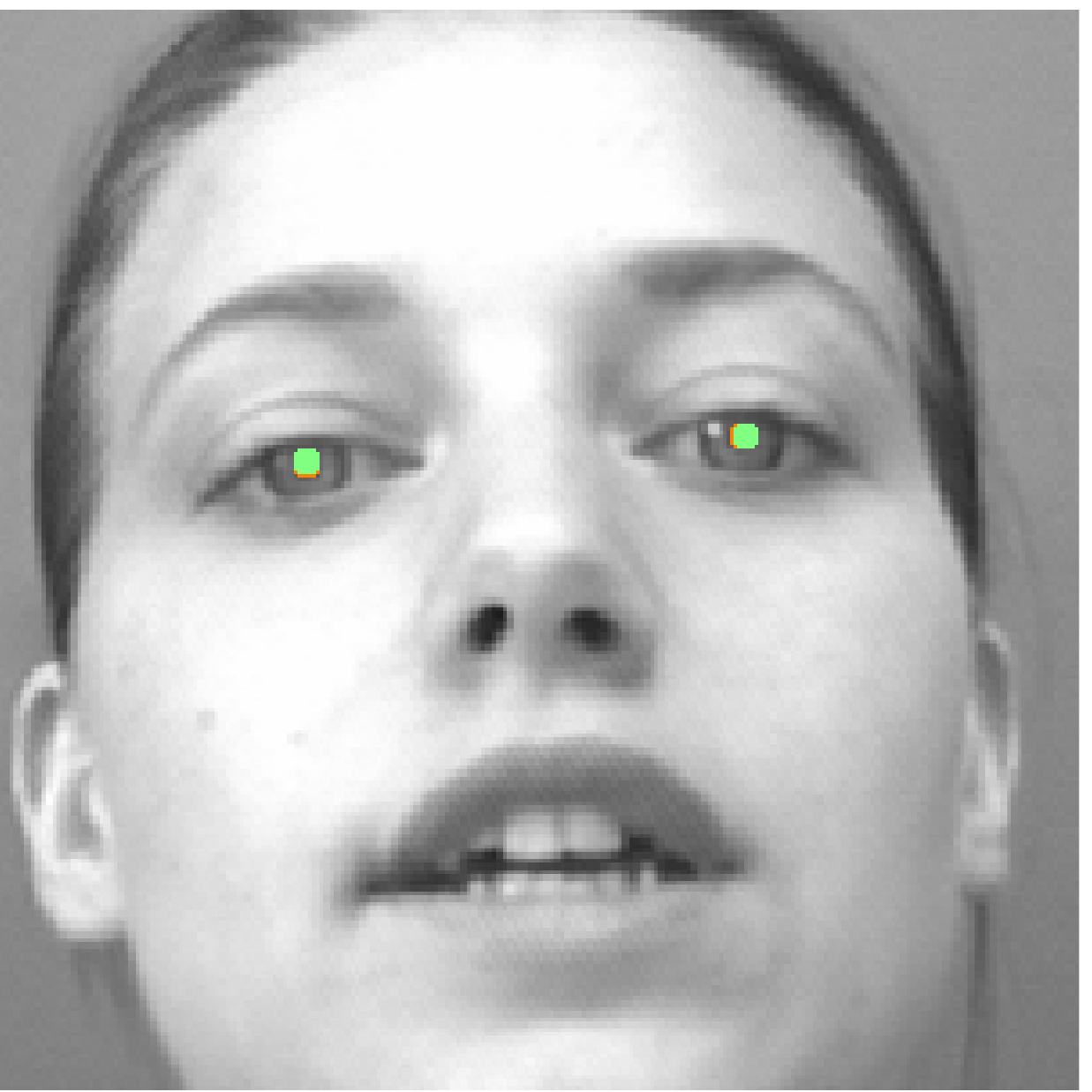}&
        \includegraphics[width=0.14 \textwidth]{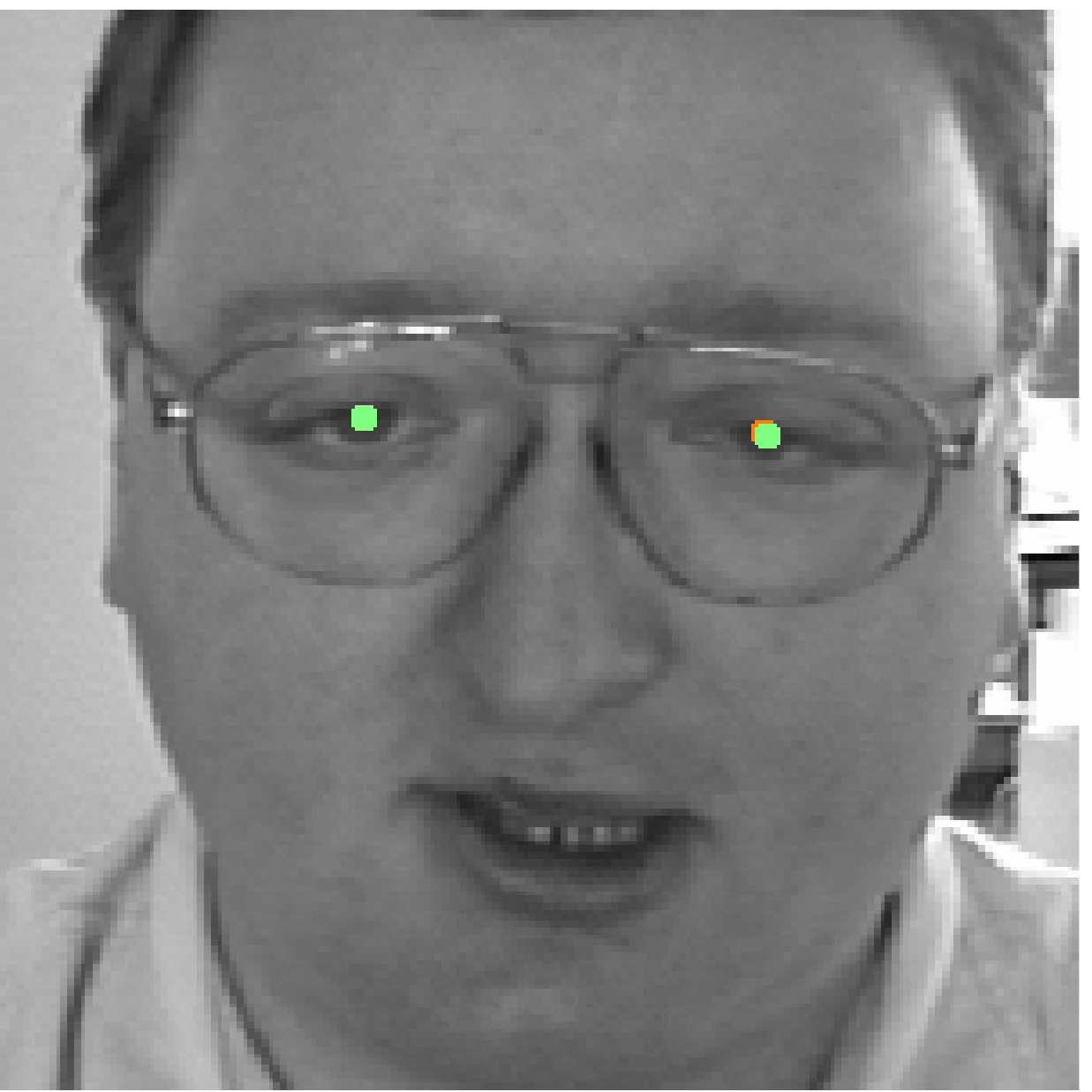}\\
        \includegraphics[width=0.14 \textwidth]{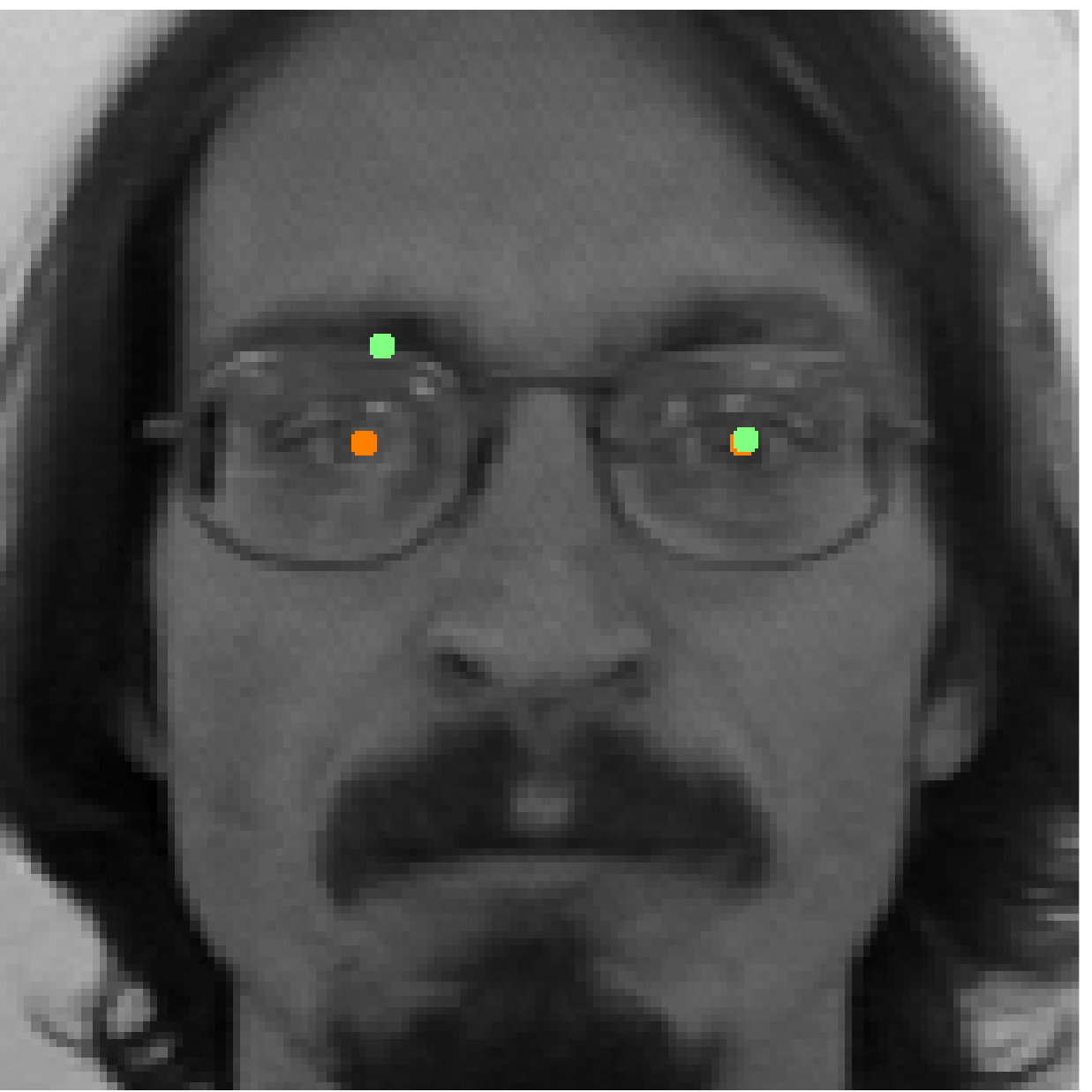}&
        \includegraphics[width=0.14 \textwidth]{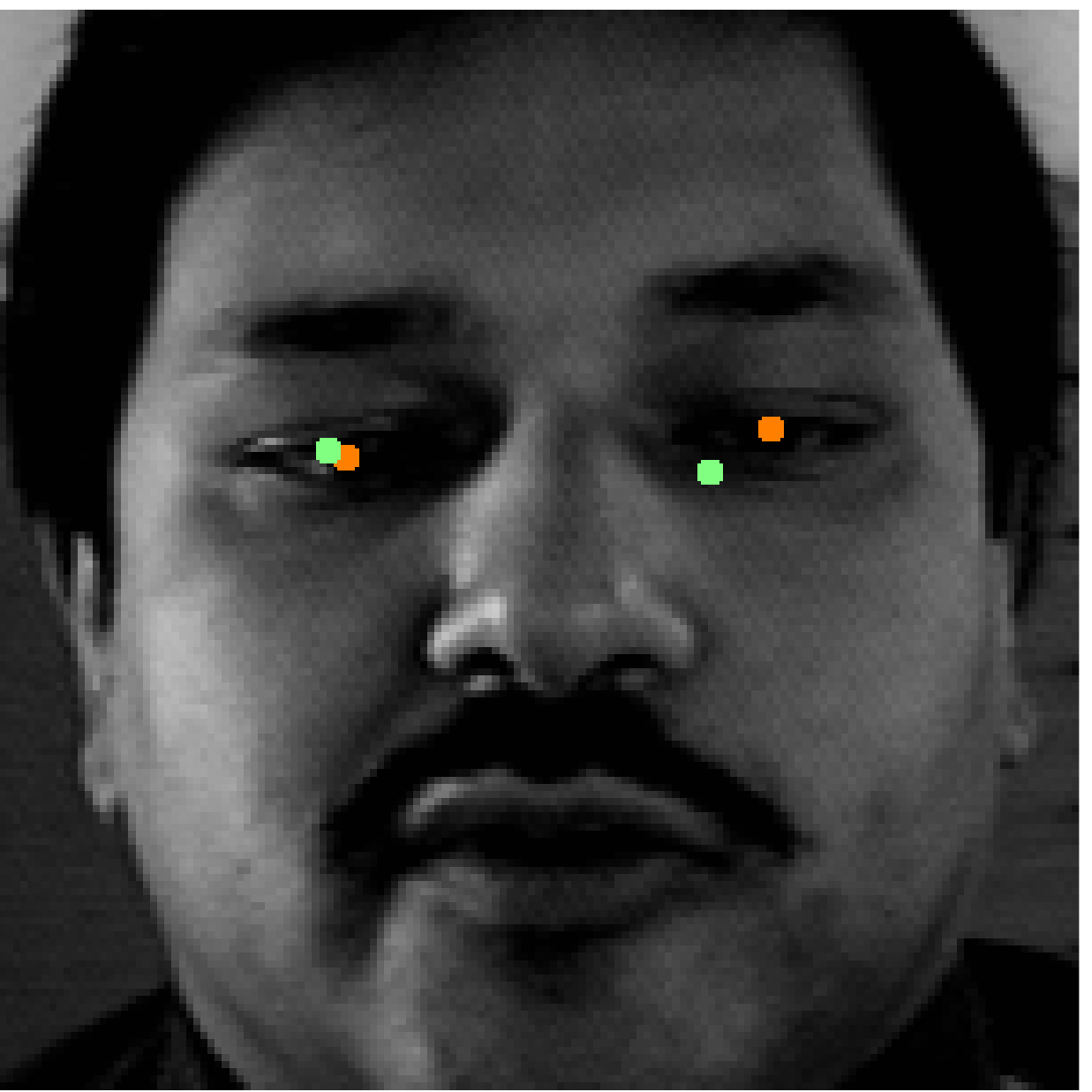}&
        \includegraphics[width=0.14 \textwidth]{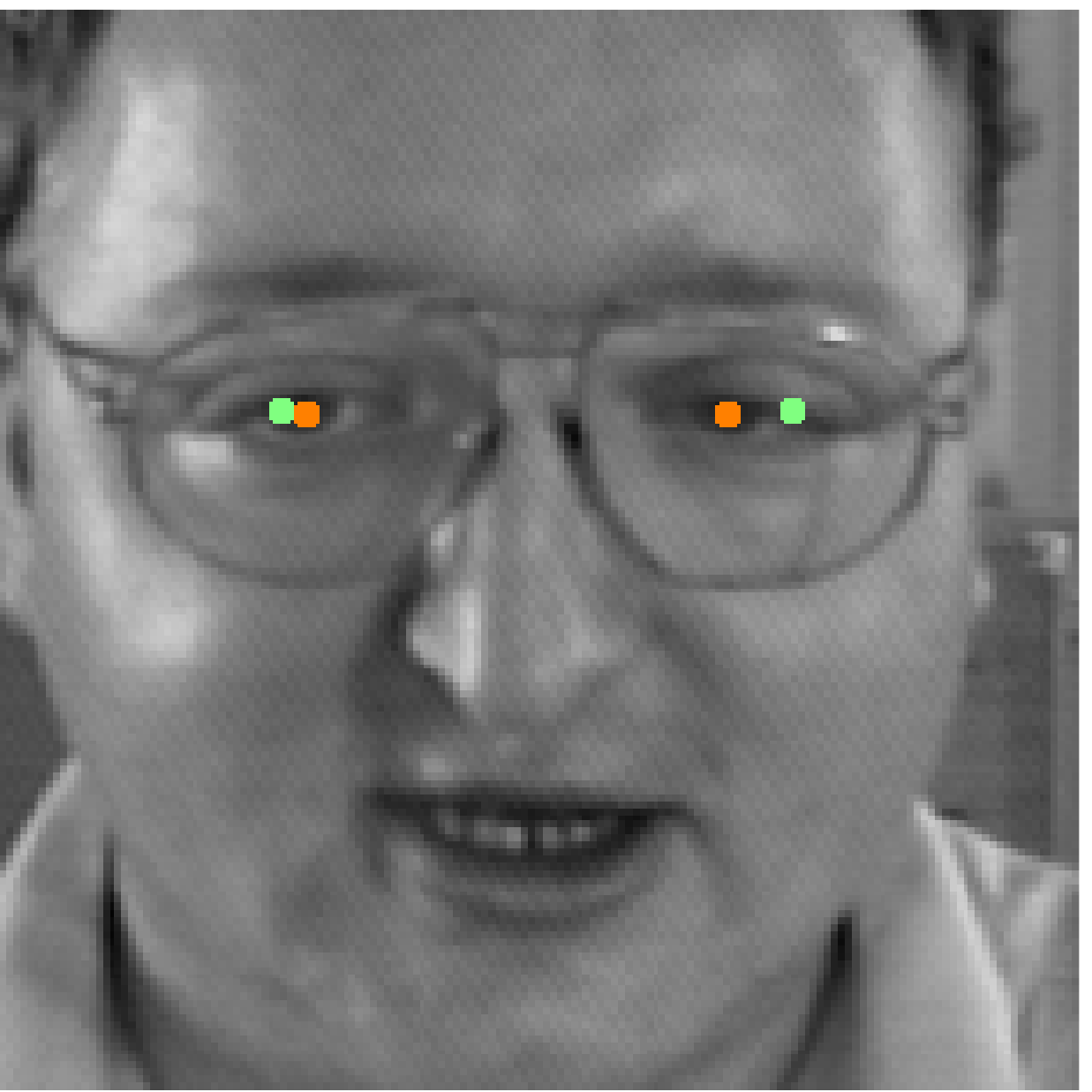}\\
    \end{tabular}
     \caption{Face cropped images from BioID database. Top row shows images with eyes correctly
      localized, while bottom row shows failure cases.}
\label{Fig:Eye_BioID}
\end{figure}

To give an initial overview of the problem  in state of the art, we consider the results reported
by other methods on the BioID database. Other solutions for eye localization are summarized in
table \ref{Tab:Rez_BioID}. The results of the methods for localization of face fiducial points are
showed in table \ref{Tab:Rez_BioIDFiduc}. Visual examples of images with localized eyes produced by
our method are shown in figure \ref{Fig:Eye_BioID}.

\begin{table*}
\centering
  \caption{ Comparison with state of the art (listed in chronological appearance) in terms of
   localization accuracy on the BioID database. The correct localization presents
   results reported by authors; values marked with ''*'' were
   inferred from authors plot. While \citeyear{Zhou:04} reports only the value for $\varepsilon <0.25$,
   the rest is reported by \cite{Ciesla:12}. The method marked with $\dag$ relied on a
   10-fold training/testing scheme, thus, at a step, using 9 parts of the BioID database for training. }
\label{Tab:Rez_BioID}

\begin{tabular}{| c || c |}
  \hline
    { }& \textbf{Accuracy}  \\ \hline 
    \begin{tabular}{c}
        { \emph{Method}} \\ \hline
        \textbf{Proposed}                   \\ \hline
        \cite{Jesorsky:01}        \\ \hline
        \cite{Wu:03}                    \\ \hline
        \cite{Zhou:04}                \\ \hline
        \cite{Cristinacce:04}  \\ \hline
        \cite{Campadelli:06}    \\ \hline
        \cite{Hamouz:05}           \\ \hline
        \cite{Turkan:08}            \\ \hline
        \cite{Kroon:08}              \\ \hline
        \cite{Asteriadis:09}    \\ \hline
        \cite{Asadifard:10}      \\ \hline
        \cite{Tim:11}                \\ \hline
        \cite{Valenti:12}+MS          \\ \hline
        \cite{Valenti:12}+SIFT $\dag$      \\ \hline
        \cite{Florea:12}           \\
    \end{tabular} &

    \begin{tabular}{c|c|c}
     $\epsilon < 0.05$   & $\epsilon < 0.1$ & $\epsilon < 0.25$  \\ \hline
      \textbf{70.46}  & \textbf{91.94 } & \textbf{98.89} \\ \hline
        40.0   & 79.00  & 91.80  \\ \hline 
        10.0*  & 54.00* & 93.00* \\ \hline 
        47.7   & 74.5   & 97.9   \\ \hline 
        55.00* & 96.00  & 98.00  \\ \hline 
        62.00  & 85.20  & 96.10  \\ \hline 
        59.00  & 77.00  & 93.00  \\ \hline 
        19.0*  & 73.68  & \colorbox{lightgray}{99.46}\qquad \\ \hline 
        65.0   & 87.0   &  98.8  \\ \hline 
        62.0*  & 89.42  & 96.0   \\ \hline 
        47.0   & 86.0   & 96.0   \\ \hline  
        82.50  & \colorbox{lightgray}{93.40}  & 98.00  \\ \hline 
        81.89  & 87.05  & 98.00  \\ \hline 
        \colorbox{lightgray}{86.09}\qquad  & 91.67  & 97.87  \\ \hline 
        57.13  & 88.97 & 98.48 
    \end{tabular} \\ \hline
\end{tabular}

\end{table*}

\begin{table*}
\centering
  \caption{ Comparison with methods that localize a multitude ($\approx 20$)  of points on
  the face on the BioID database. All the results were extrapolated from authors graphs. We note that in these
  cases eye localization was not the major objective and authors report an \emph{average} error
  for the entire set of points; thus we also used \emph{average} achieved error. The reported time is
  for determination of the entire set of points.} \label{Tab:Rez_BioIDFiduc}

\begin{tabular}{| c || c || c |}
  \hline
    { }& \textbf{Accuracy}  & \textbf{Time performance} \\ \hline 
    \begin{tabular}{c | c}
        { \emph{Method}}                &\emph{ No. pts.} \\ \hline
        \textbf{Proposed}               & \textbf{2}        \\ \hline
        \cite{Vukadinovic:05}& 20                \\ \hline
        \cite{Milborrow:08}    & 17                \\ \hline
        \cite{Valstar:10}        & 22                \\ \hline
        \cite{Belhumeur:11}    & 29                \\ \hline
        \cite{Mostafa:12}       & 17                \\ 
    \end{tabular} &

    \begin{tabular}{c|c}
     $\epsilon < 0.05$   & $\epsilon < 0.1$   \\ \hline
      \colorbox{lightgray}{\textbf{78.80}}  & \colorbox{lightgray}{\textbf{96.04 }}  \\ \hline
        15.0  & 78.0  \\ \hline 
        66.0  & 95.0    \\ \hline 
        74.00 & 95.00  \\ \hline 
        62.00 & 85.20  \\ \hline 
        74.00 & 96.00  \\ 
    \end{tabular} & 
    \begin{tabular}{c|c|c}
      \emph{Duration} & \emph{Platform} & $\frac{Duration}{point}$    \\ \hline
      \colorbox{lightgray}{\textbf{13 msec}} & \textbf{i7 2,7 GHz}& \textbf{6.5 msec}  \\ \hline 
        n/a            & n/a                &  n/a               \\ \hline 
        230 msec        & P4 3Ghz            &  13.53 msec        \\ \hline 
        n/a           & n/a                 &  n/a               \\ \hline 
        $\approx$950 msec & i7 3.06GHz      & 32.75 msec         \\ \hline 
        470 msec      & i7 2.93GHz          & 27.64 msec         \\ 
     \end{tabular} \\
    \hline
\end{tabular}

\end{table*}

Analyzing the performance, first we note that our method significantly outperforms in both time and
accuracy other methods relying on image projections (\cite{Zhou:04}, \cite{Turkan:08}). The
explanation lies in the normalization procedure implied when constructing the ZEP feature.

Comparison with face feature fiducial points localization is not straightforward. While such
methods localize significantly more points than simple eye centers localization, they also rely
strongly on the inter-spatial relation among them to boost the overall performance. Furthermore,
they often do not localize eye centers, but eye corners and the top/bottom of the eye, which in
many cases are more stable than the eye center (i.e. not occluded or influenced by gaze). And yet
we note that our method is comparable in terms of accuracy and significantly faster (if one
normalizes the reported time by the number of detected points).

Regarding other methods for eye localization, the proposed method ranks as one of the top methods
for all accuracy tests, being always close to the best solution. Furthermore taking into account
that on BioID database there are only $\approx 50$ images (3\%) with closed eyes, methods that
search circular (symmetrical) shapes have better circumstances. Because we targeted images with
expressions, we specifically included in our training data set closed eyes. To validate this
assumption we tested with very good results on the Cohn-Kanade database showing that our method is
more robust in that case as showed in figure \ref{Fig:BioID_Results} (b).

Considering as most important criterion the accuracy at $\epsilon < 0.05$, we note that
\citeyear{Tim:11} and \citeyear{Valenti:12}  provide higher accuracy. Yet, we must also note and
the highest performance achieved by a variation of the method described in \cite{Valenti:12},
namely Val.\&Gev.+SIFT contains a 10-fold testing scheme, thus using 9 parts of the BioID database
for training. Furthermore, taking into account that BioID database was used for more that 10 years
and provides limited variation, it has been concluded \cite{Belhumeur:11}, \cite{Dantone:12} that
other tests are also required to validate a method.

\citeyear{Valenti:12} provide results on other datasets and made public the associated code for
their baseline system (Val.\&Gev.+MS) which  is not database dependent. \citeyear{Tim:11} do not
provide results on any other database except BioID or source code, yet there is publicly
available\footnote{At
\url{http://thume.ca/projects/2012/11/04/simple-accurate-eye-center-tracking-in-opencv/}.} code
developed with author involvement. Thus, in continuation, we will compare the hereby proposed
method against these two on other datasets. Additionally, we include the comparison against the eye
detector developed by \citeyear{Ding:10} which has also been trained and tested on other database,
thus is not BioID dependent.

\subsection{Robustness to Eye Expressions}

\begin{figure}[tb]
\center
 \begin{tabular}{ccc }
        \includegraphics[width=0.14 \textwidth]{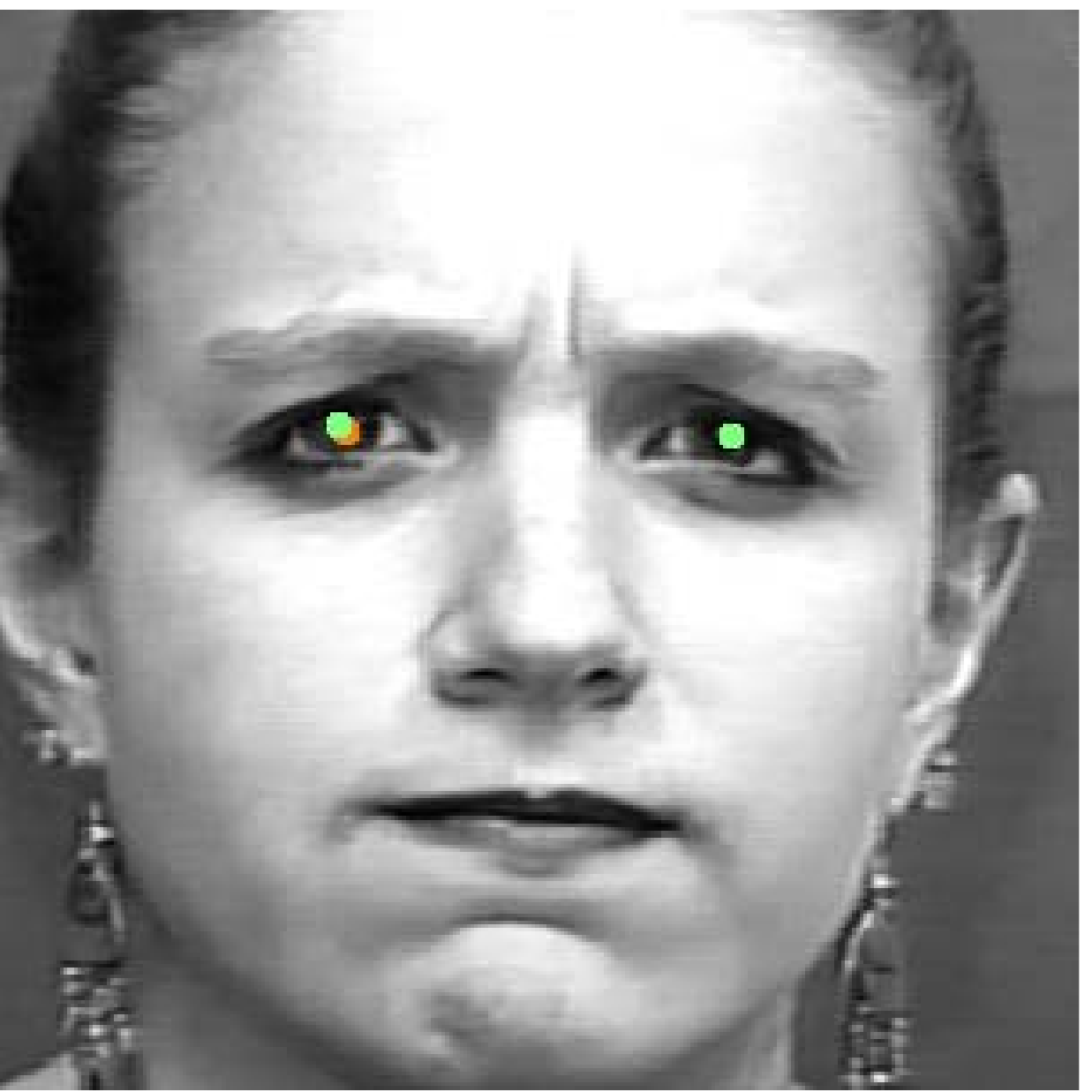}&
        \includegraphics[width=0.14 \textwidth]{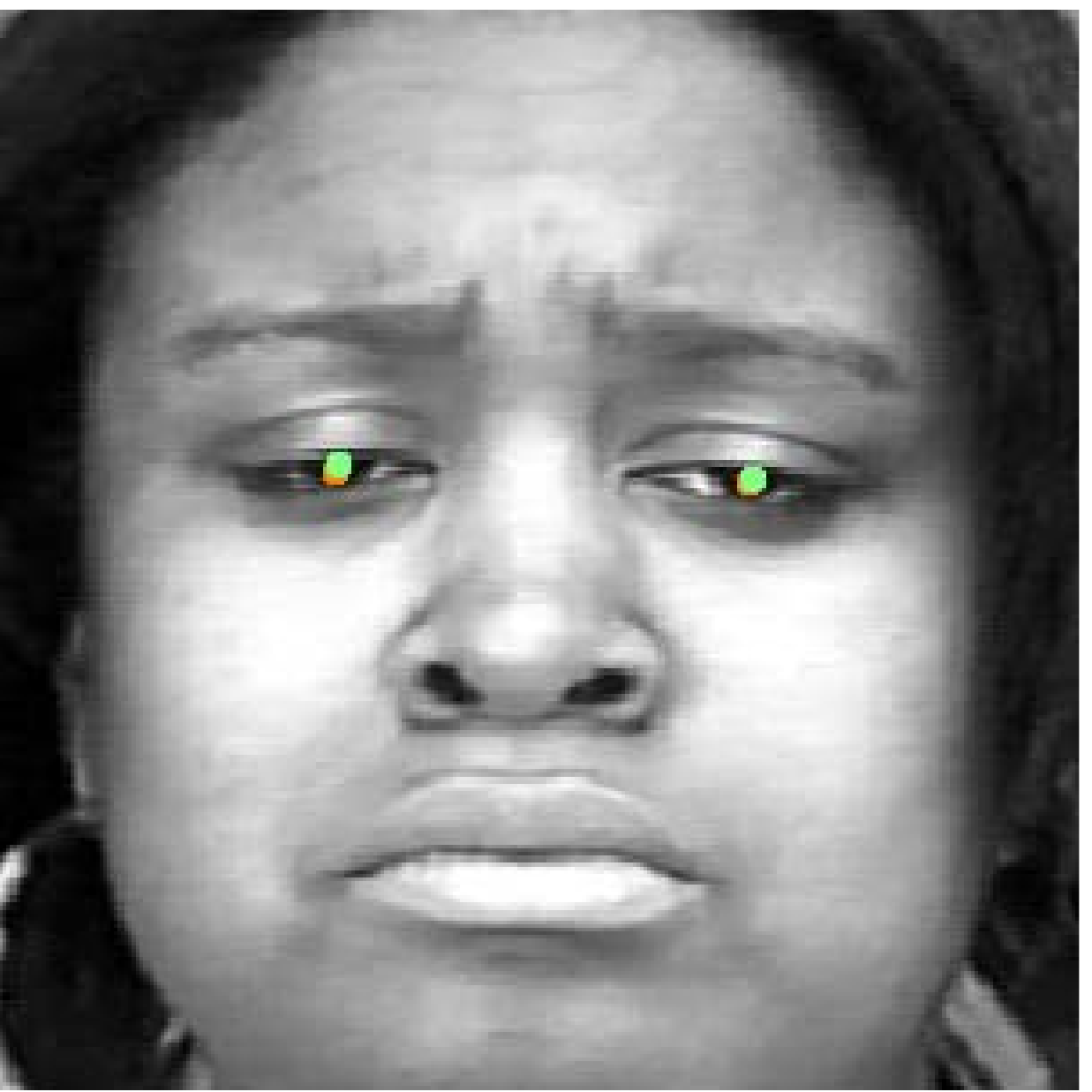}&
        \includegraphics[width=0.14 \textwidth]{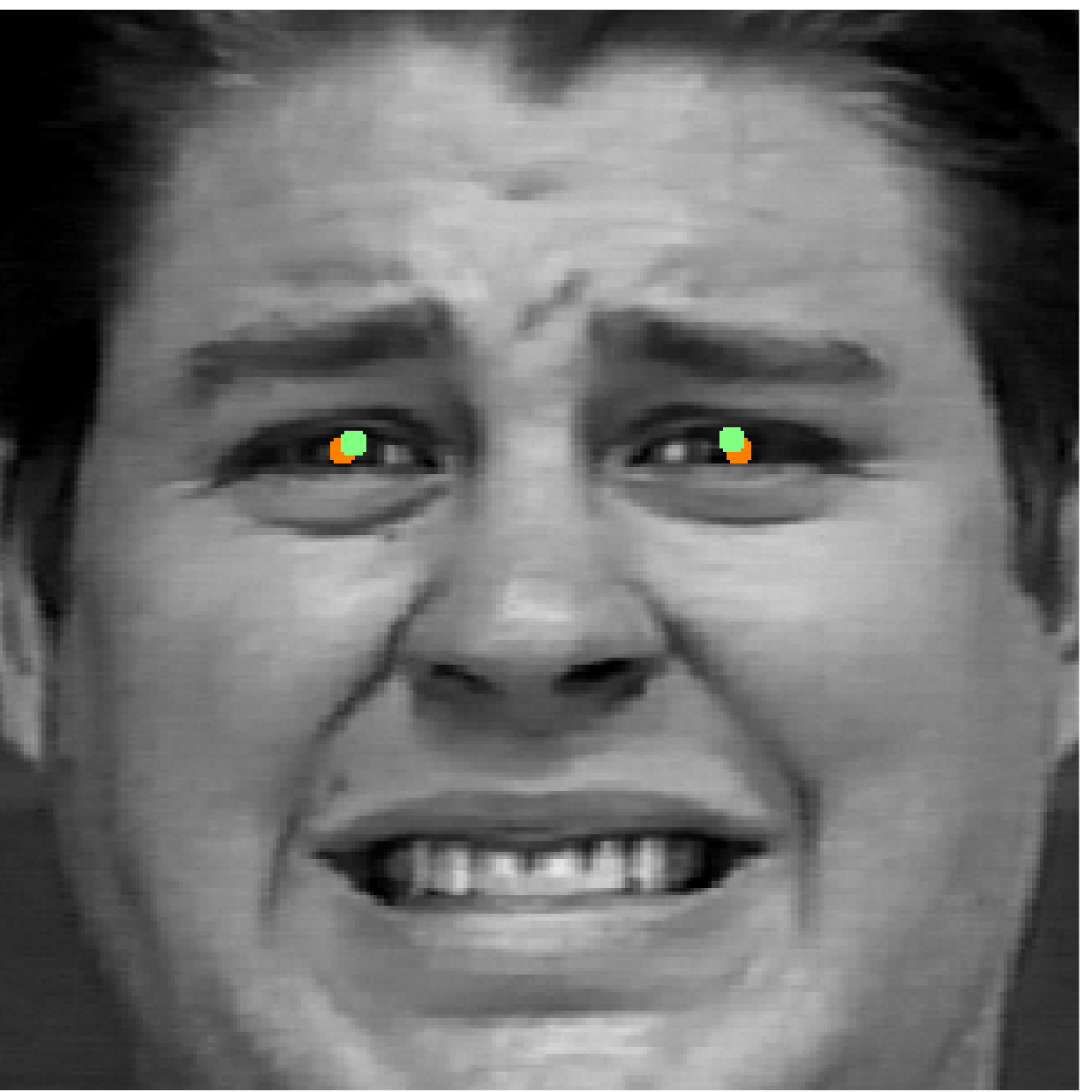}\\

        \includegraphics[width=0.14 \textwidth]{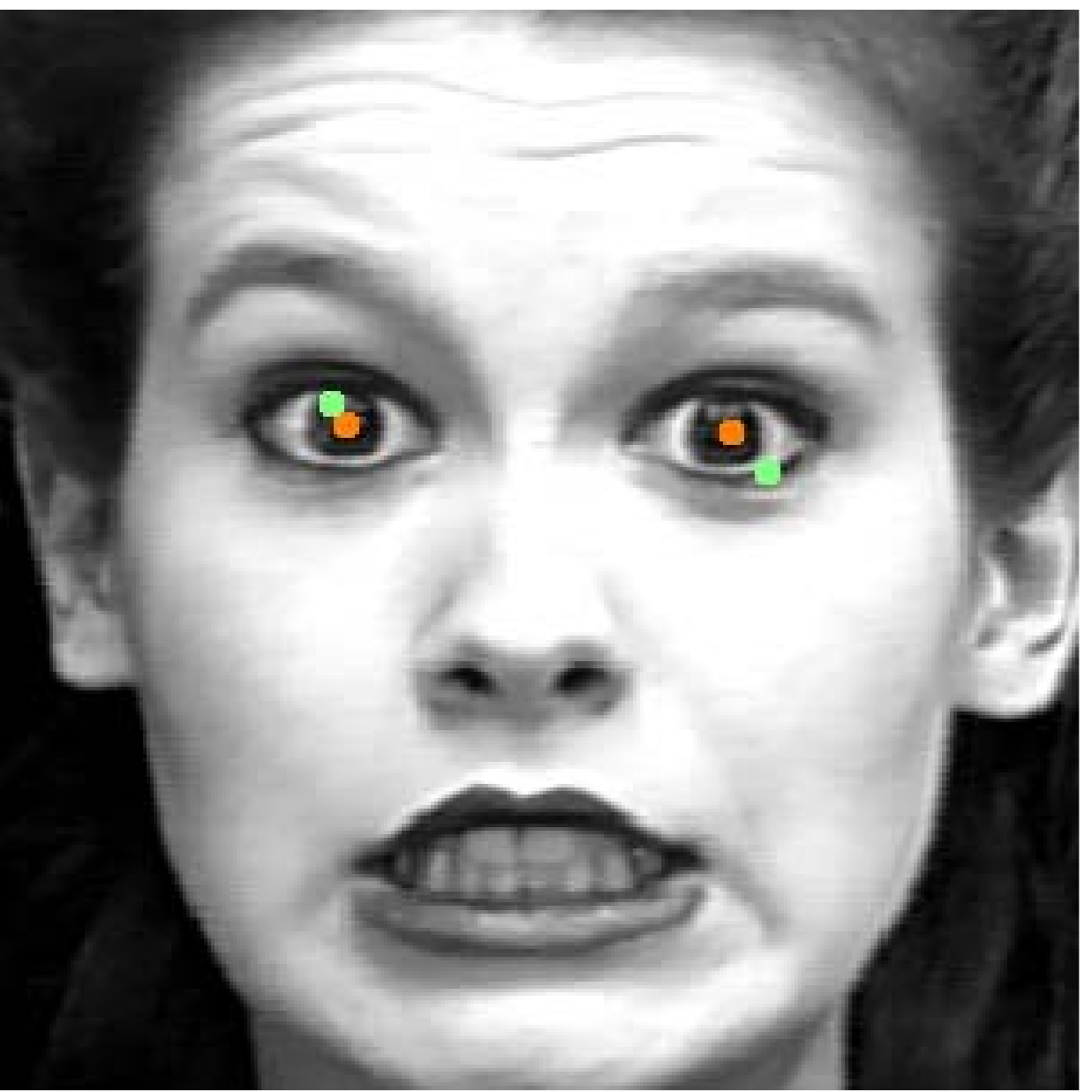}&
        \includegraphics[width=0.14 \textwidth]{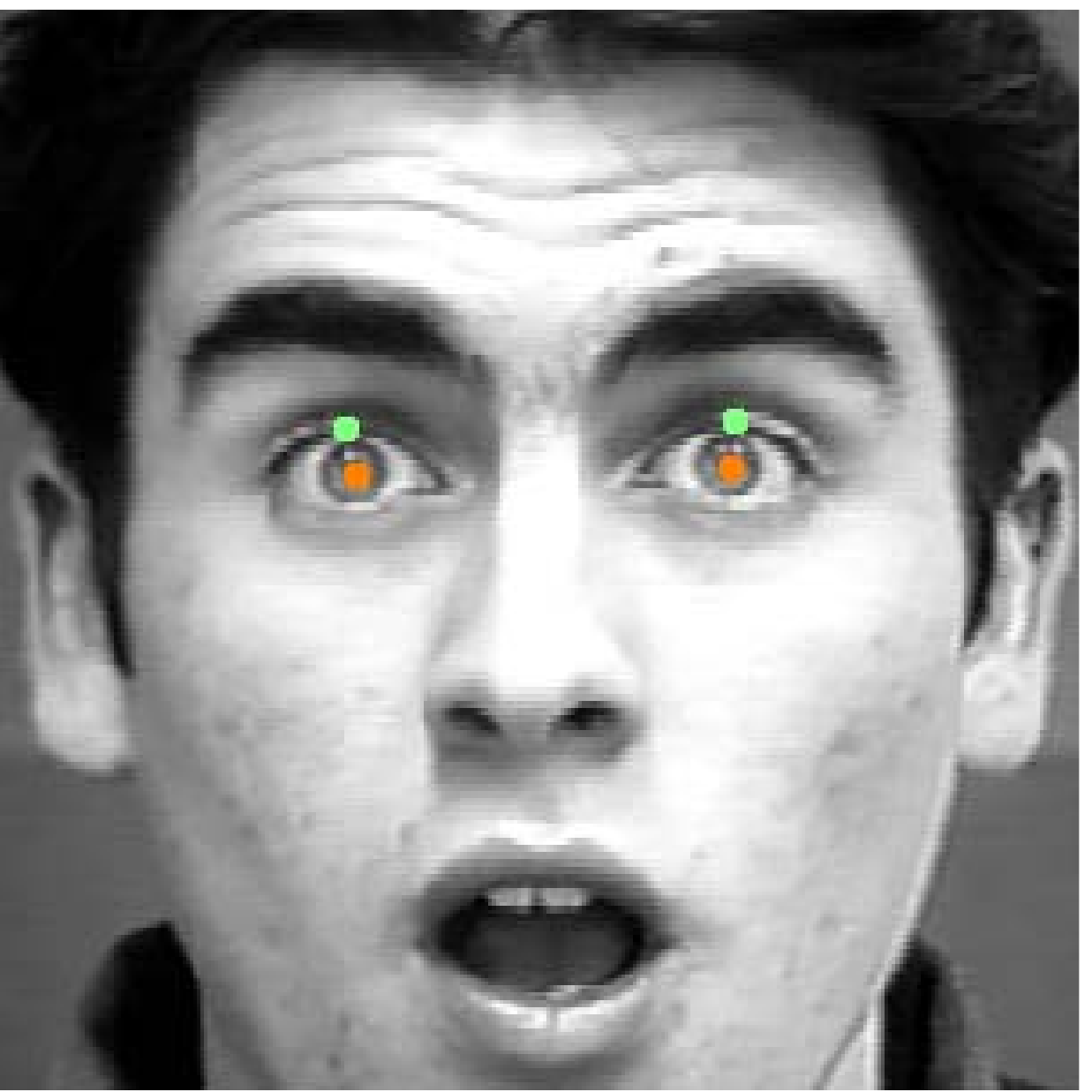}&
        \includegraphics[width=0.14 \textwidth]{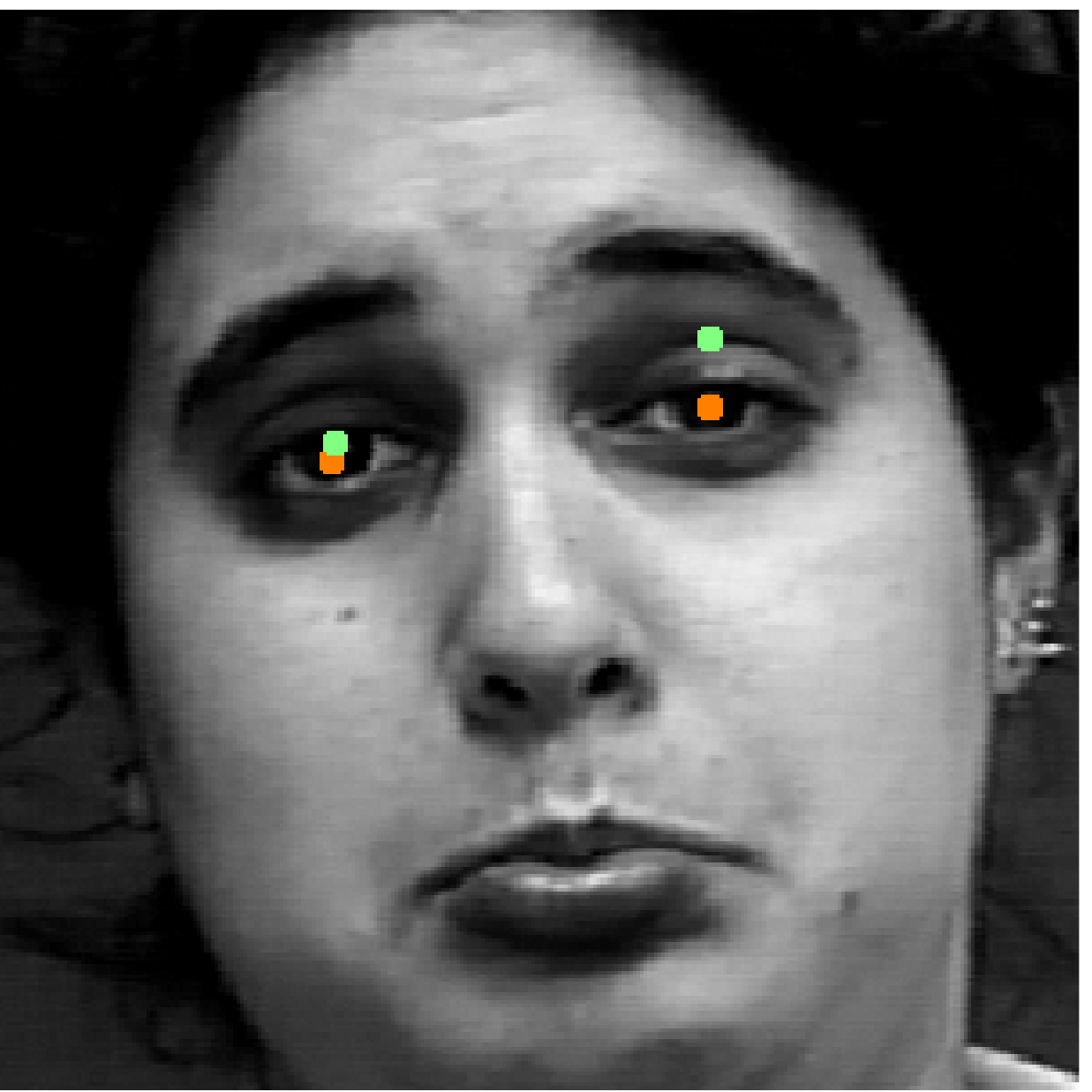}\\
    \end{tabular}
     \caption{Face cropped images from the Cohn-Kanade database. The ground truth eyes are marked with
      red (dark grey), while detected eyes with green (light grey). Top row images show eyes
     correctly localized, while bottom row shows failure cases.} \label{Fig:Eye_CK}
\end{figure}

As mentioned in the introduction, we are specifically interested in the performance of the eye
localization with respect to facial expressions, as real-life cases with fully opened eyes looking
straight are rare. We tested the performance of the proposed method on the Cohn-Kanade database
\cite{Kanade:00}. This database was developed for the study of emotions, contains frontal
illuminated portraits and it is challenging through the fact that eyes are in various poses
(near-closed, half-open, wide-open). We tested only on the neutral pose and on the expression apex
image from each example. The correct eye locations, with standard precisions, are shown in table
\ref{Tab:RezCK}. Typical localization results are presented in figure \ref{Fig:Eye_CK}, while the
maximum, average and minimum errors are plotted in figure \ref{Fig:BioID_Results} (b).

\begin{table}[tb]
 \centering
    \caption{Percentage of correct eye localization on the Cohn-Kanade database. We report results
     on the neutral poses, expression apex and overall and compare against method from \cite{Valenti:12}, \cite{Tim:11} and \cite{Ding:10}.
     We marked with light-gray background the best achieved performance for each accuracy criterion and respectively for each image type.}
    \label{Tab:RezCK}

    \begin{tabular}{| c | c|c|c|c |}
    \hline
         { } & \multicolumn{4}{ |c| }{\textbf{Accuracy}} \\ \hline
         \textbf{Method} &  \emph{Type} & $\epsilon < 0.05$ & $\epsilon < 0.1$ & $\epsilon < 0.25$  \\ \hline
         \multirow{3}{*}{\textbf{Proposed}}&
            Neutral  &{ } \colorbox{lightgray}{\textbf{76.0}}  { }&{ } \colorbox{lightgray}{\textbf{99.0}} { }&{ } \colorbox{lightgray}{\textbf{100}} \\ \cline{2-5}
         { } &Apex     &{ } \colorbox{lightgray}{\textbf{71.9}}  { }&{ } \colorbox{lightgray}{\textbf{95.7}} { }&{ }  \colorbox{lightgray}{\textbf{100}} \\ \cline{2-5}
         { } &Total    &{ } \colorbox{lightgray}{\textbf{73.9}}  { }&{ } \colorbox{lightgray}{\textbf{97.3}} { }&{ } \colorbox{lightgray}{\textbf{100}} \\
                     \hline \hline

       \multirow{3}{*}{\begin{tabular}{c} \cite{Valenti:12}\end{tabular} } &
                           Neutral  &{ } 46.0         { }&{ } 95.7 { }&{ }  99.6 \\ \cline{2-5}
                      { } &Apex     &{ } 35.1         { }&{ } 92.4 { }&{ }  98.8 \\ \cline{2-5}
                      { } &Total    &{ } 40.6         { }&{ } 94.0 { }&{ }  99.2 \\ \hline \hline
       \multirow{3}{*}{\begin{tabular}{c}  \cite{Tim:11} \end{tabular} } &
                           Neutral  &{ } 66.0         { }&{ } 95.4 { }&{ }  99.0 \\ \cline{2-5}
                      { } &Apex     &{ } 61.4         { }&{ } 85.1 { }&{ }  93.4 \\ \cline{2-5}
                      { } &Total    &{ } 63.7         { }&{ } 90.2 { }&{ }  96.2 \\ \hline \hline
       \multirow{3}{*}{\begin{tabular}{c}  \cite{Ding:10} \end{tabular} } &
                           Neutral  &{ } 14.3         { }&{ } 75.9 { }&{ }  \colorbox{lightgray}{100} \\ \cline{2-5}
                      { } &Apex     &{ } 11.8         { }&{ } 72.8 { }&{ }  \colorbox{lightgray}{100} \\ \cline{2-5}
                      { } &Total    &{ } 13.1         { }&{ } 74.4 { }&{ }  \colorbox{lightgray}{100} \\ \hline \hline

    \end{tabular}
\end{table}

We note that solutions that try to fit a circular or a symmetrical shape over the iris, like
\cite{Valenti:12} or \cite{Tim:11}, and thus, perform well on open eyes, do encounter significant
problems when facing eyes in expressions (as it is shown in table \ref{Tab:RezCK}). Taking into
account the achieved results, which are comparable on neutral pose and expression apex images, we
show that our method performs very well under such complex conditions. Achieved results indicate
approximately a doubled accuracy when compared with the foremost state of the art method.

\subsection{Robustness to Illumination and Pose}

We systematically evaluated the robustness of the proposed algorithm with respect to lighting and
pose changes. This was tested onto the Extended Yale Face Database B (B+) \cite{Lee:05}. We stress
that part of the Yale B database \cite{Georghiades:01} was used for training the MLP for  lateral
illumination, thus the training and testing sets are completely different.

The Extended Yale B database contains 16128 gray-scale images of 28 subjects, each seen under 576
viewing conditions (9 poses $\times$ 64 illuminations). The size of each image is $640 \times 480$.
The robustness with respect to pose and with respect to illumination was evaluated separately.

For evaluating the robustness to illumination, we tested the system on 28 faces, in neutral pose,
under changing illumination (64 cases). The results are summarized in table
\ref{Tab:Rez_YaleBIlum}.

\begin{figure}[tb]
\center
    \begin{tabular}{cc c}
        \includegraphics[width=0.14 \textwidth]{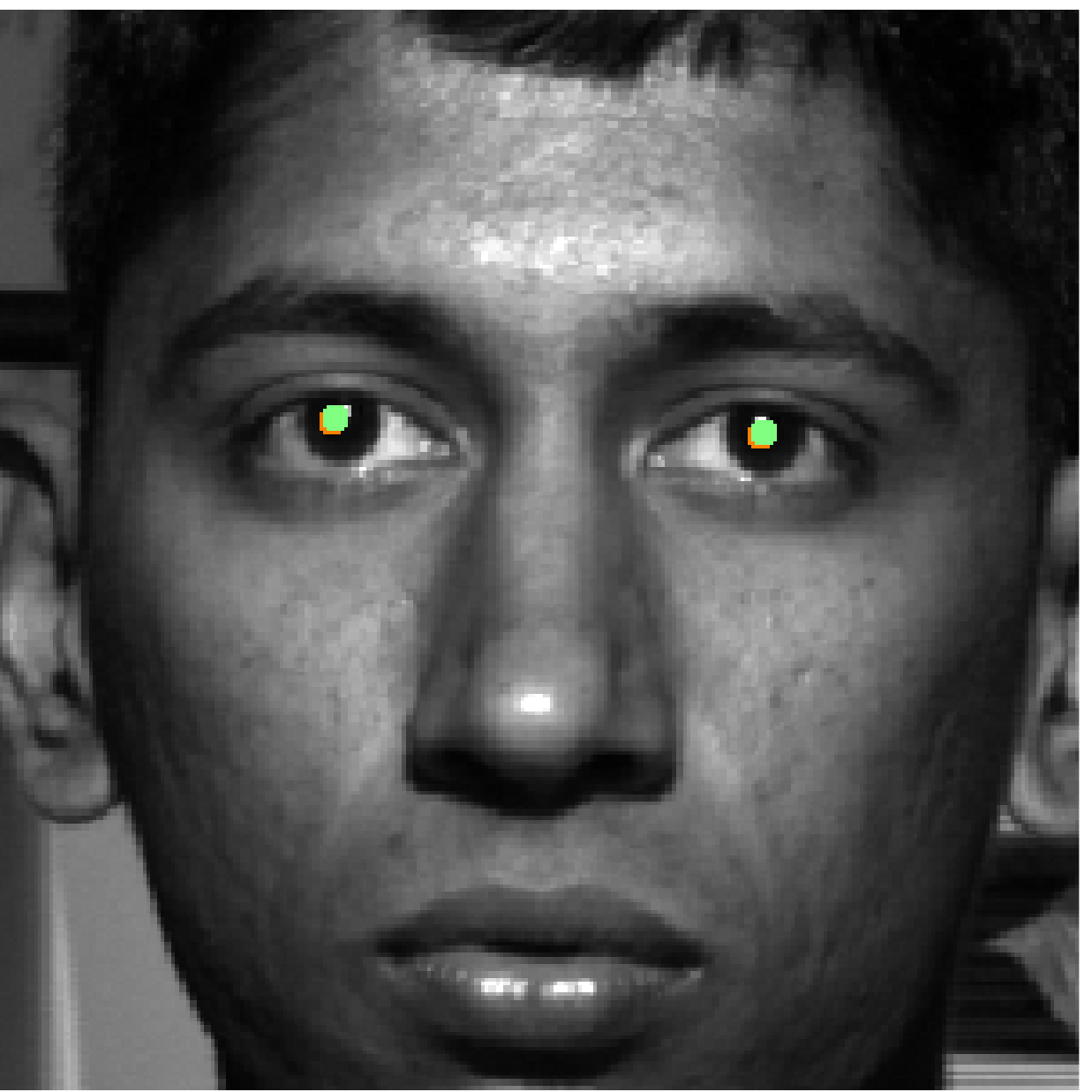}&
        \includegraphics[width=0.14 \textwidth]{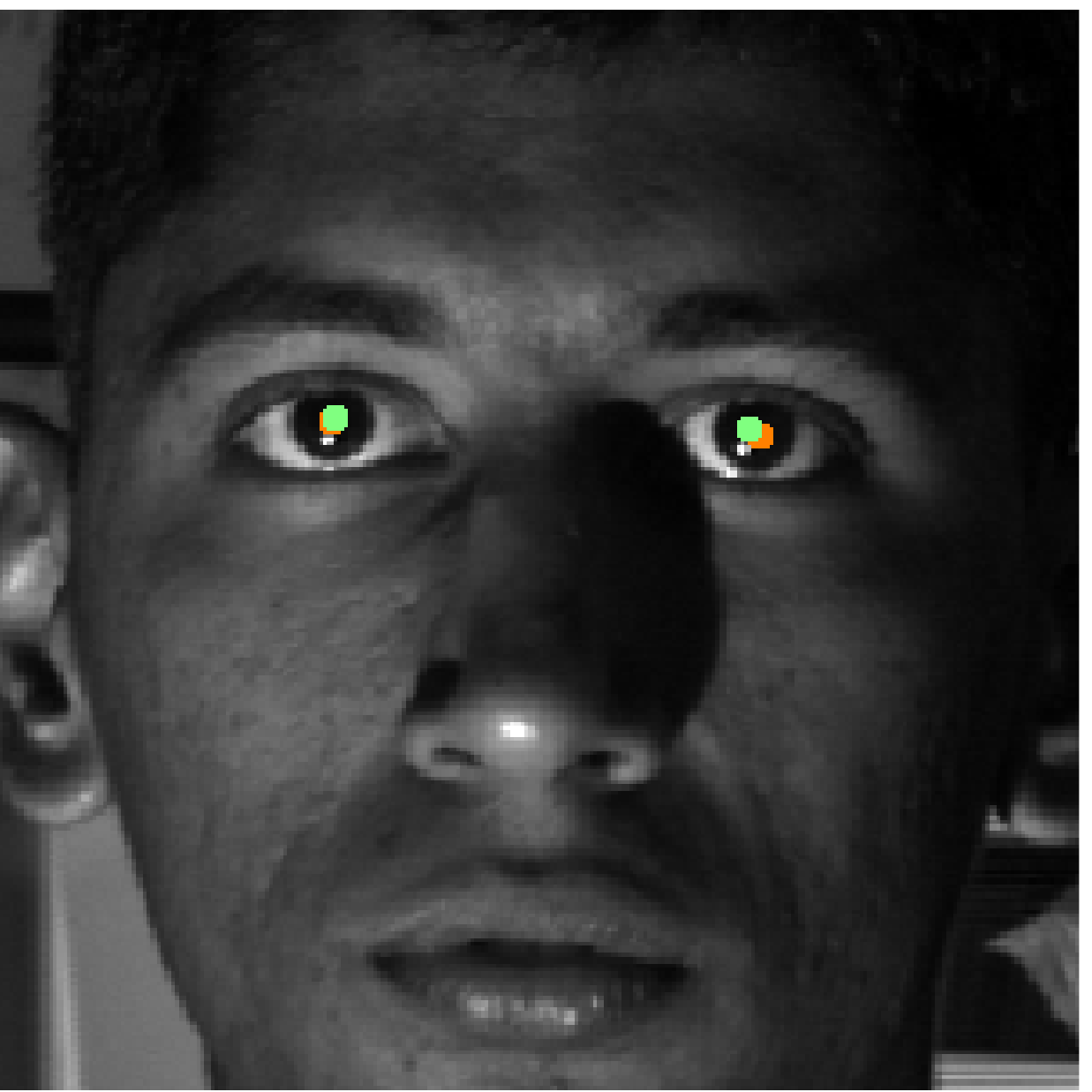}&
        \includegraphics[width=0.14 \textwidth]{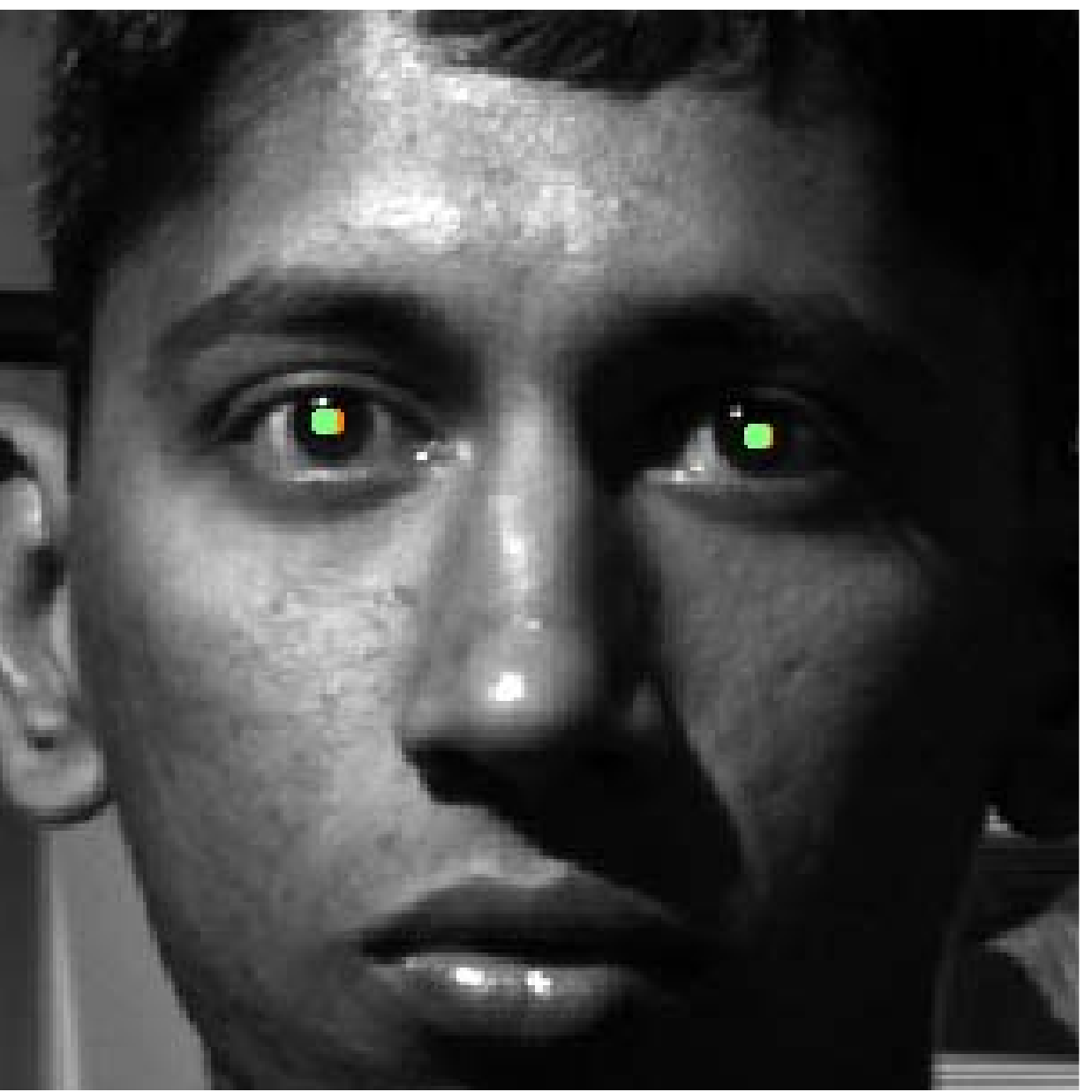}\\

        \includegraphics[width=0.14 \textwidth]{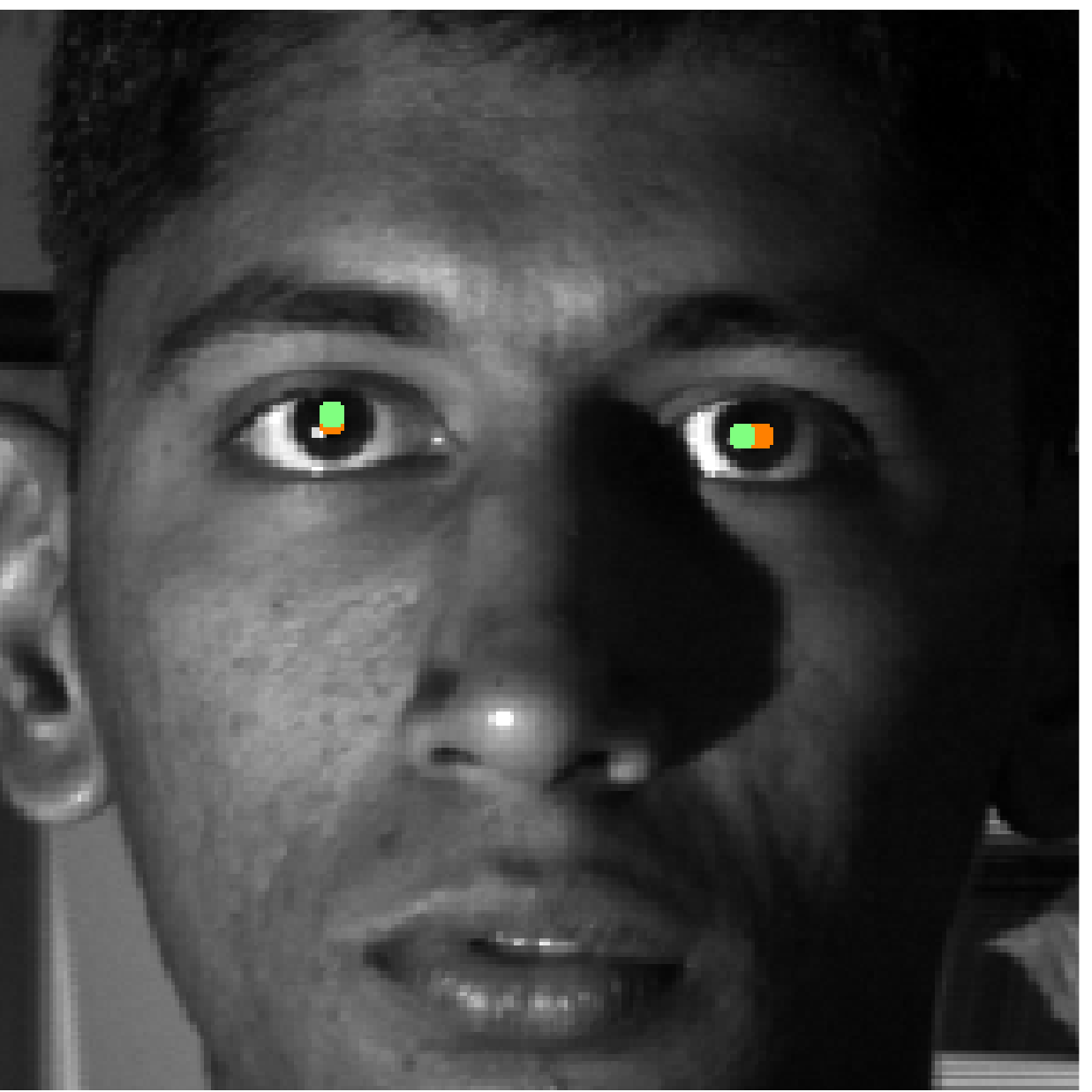}&
        \includegraphics[width=0.14 \textwidth]{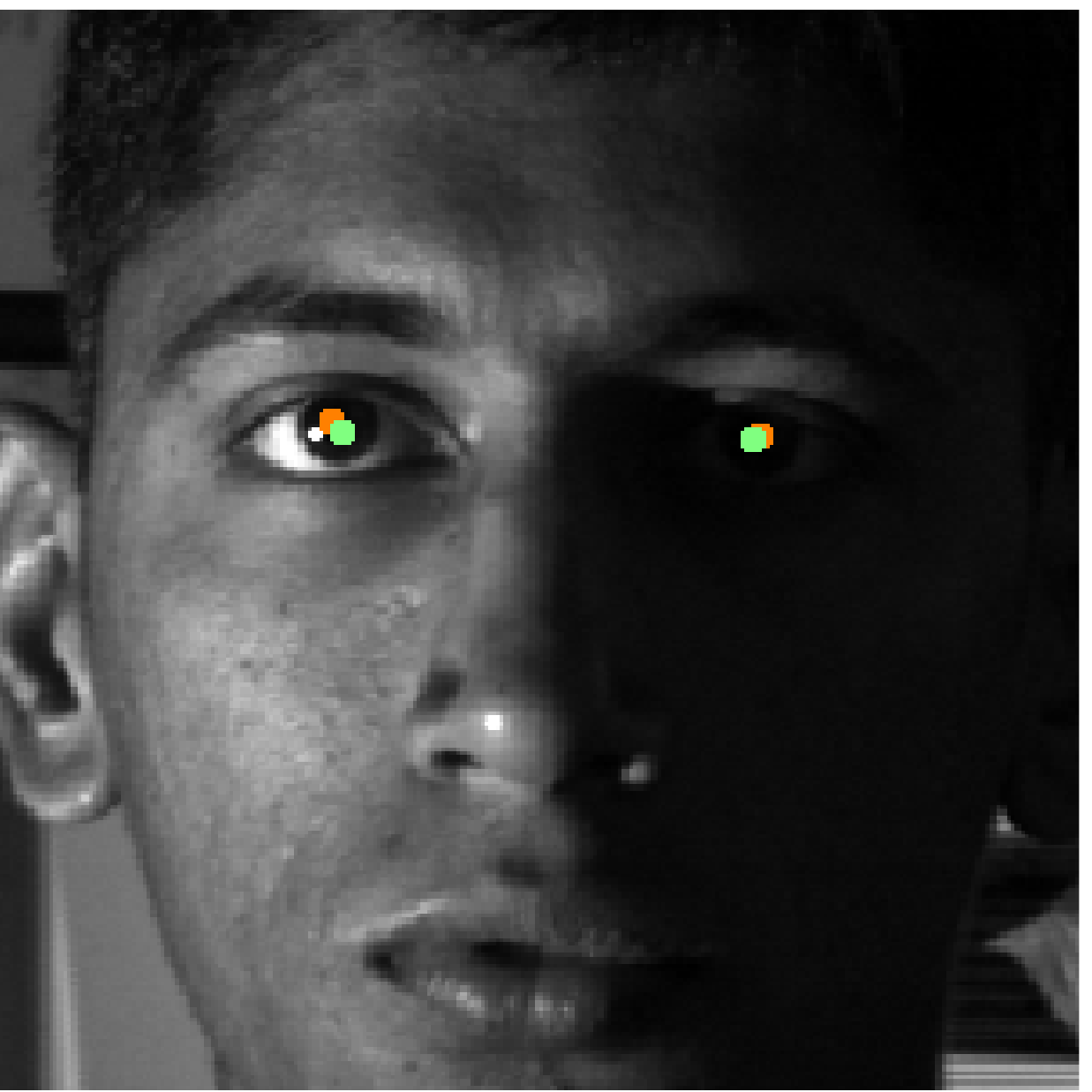}&
        \includegraphics[width=0.14 \textwidth]{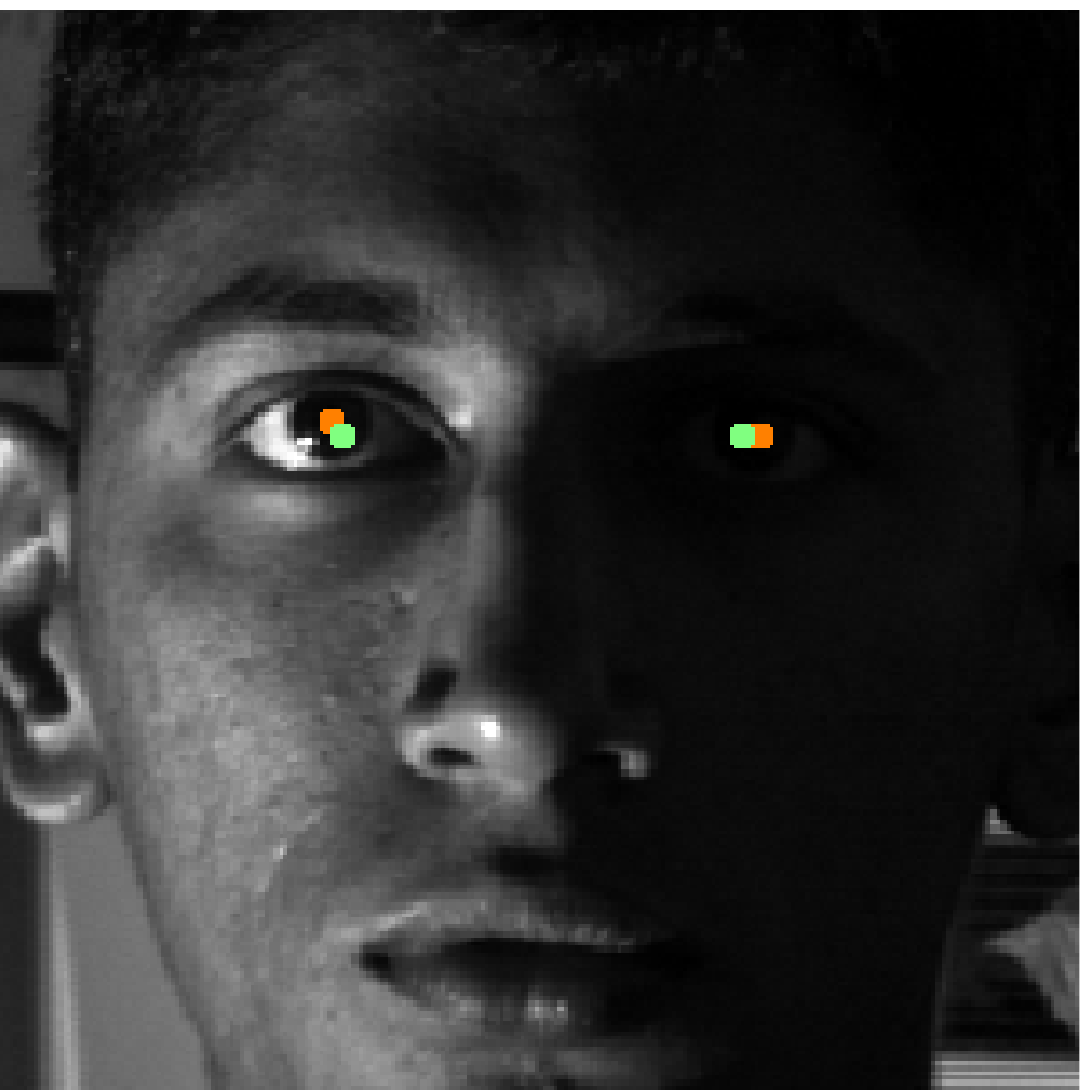}\\
        \includegraphics[width=0.14 \textwidth]{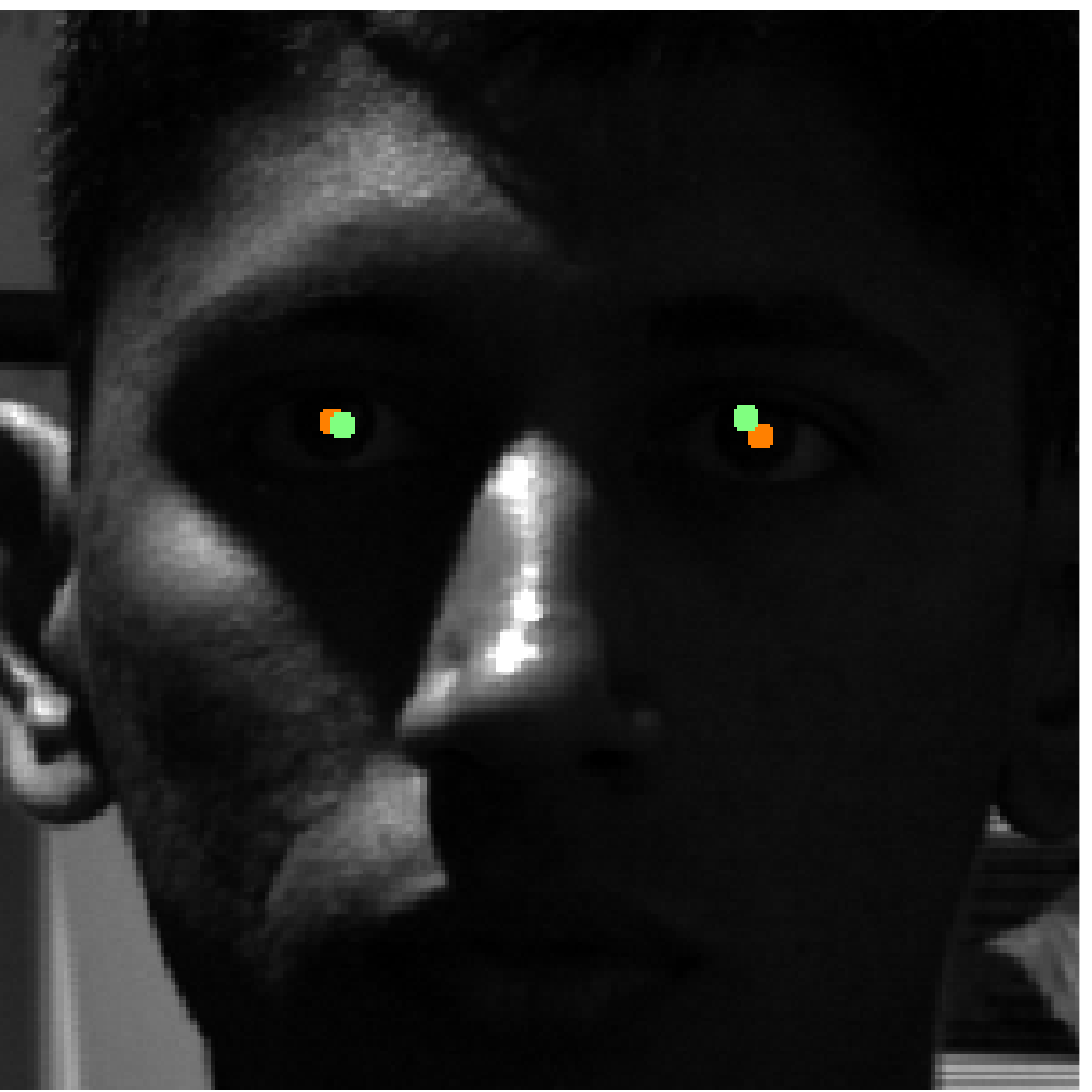}&
        \includegraphics[width=0.14 \textwidth]{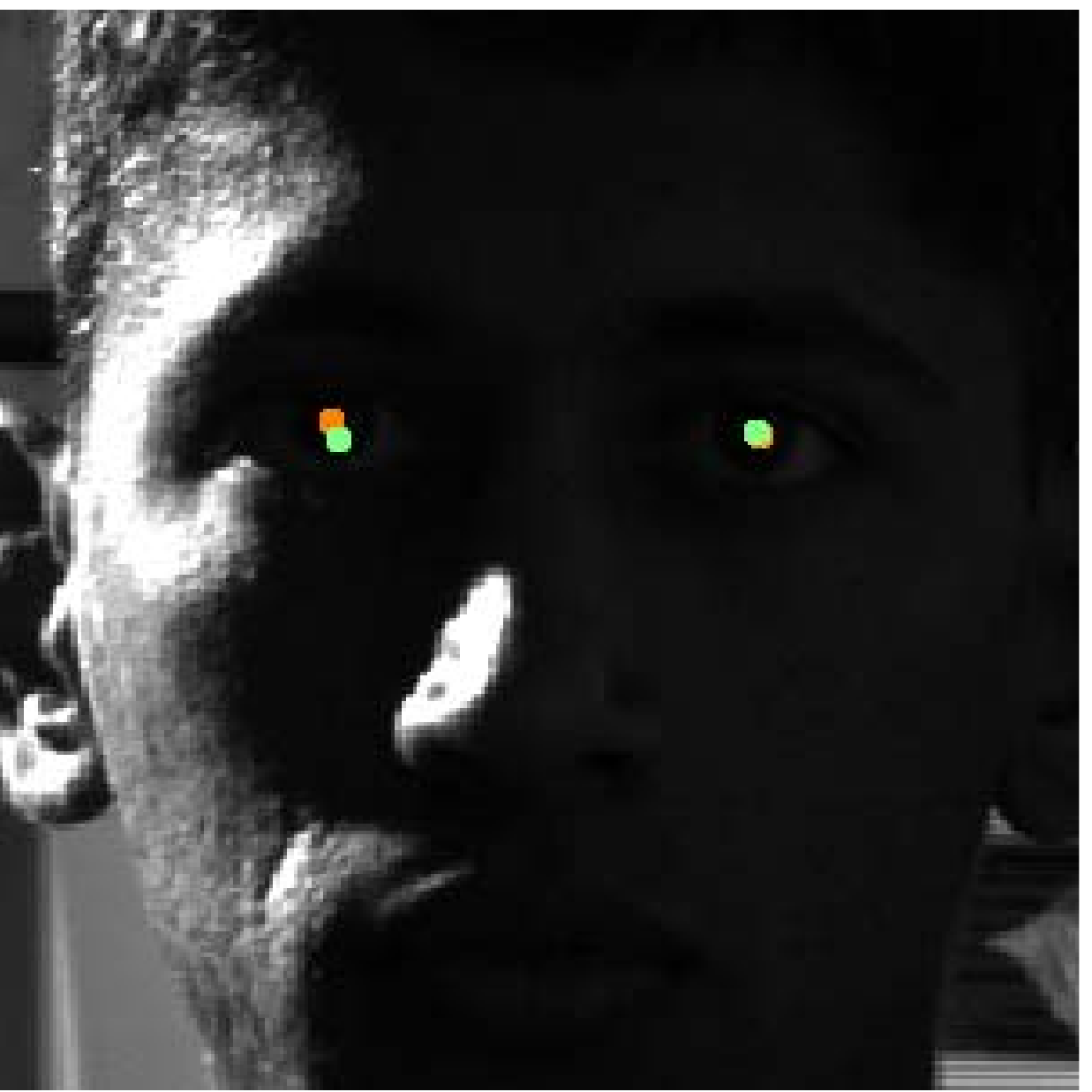}&

        \includegraphics[width=0.14 \textwidth]{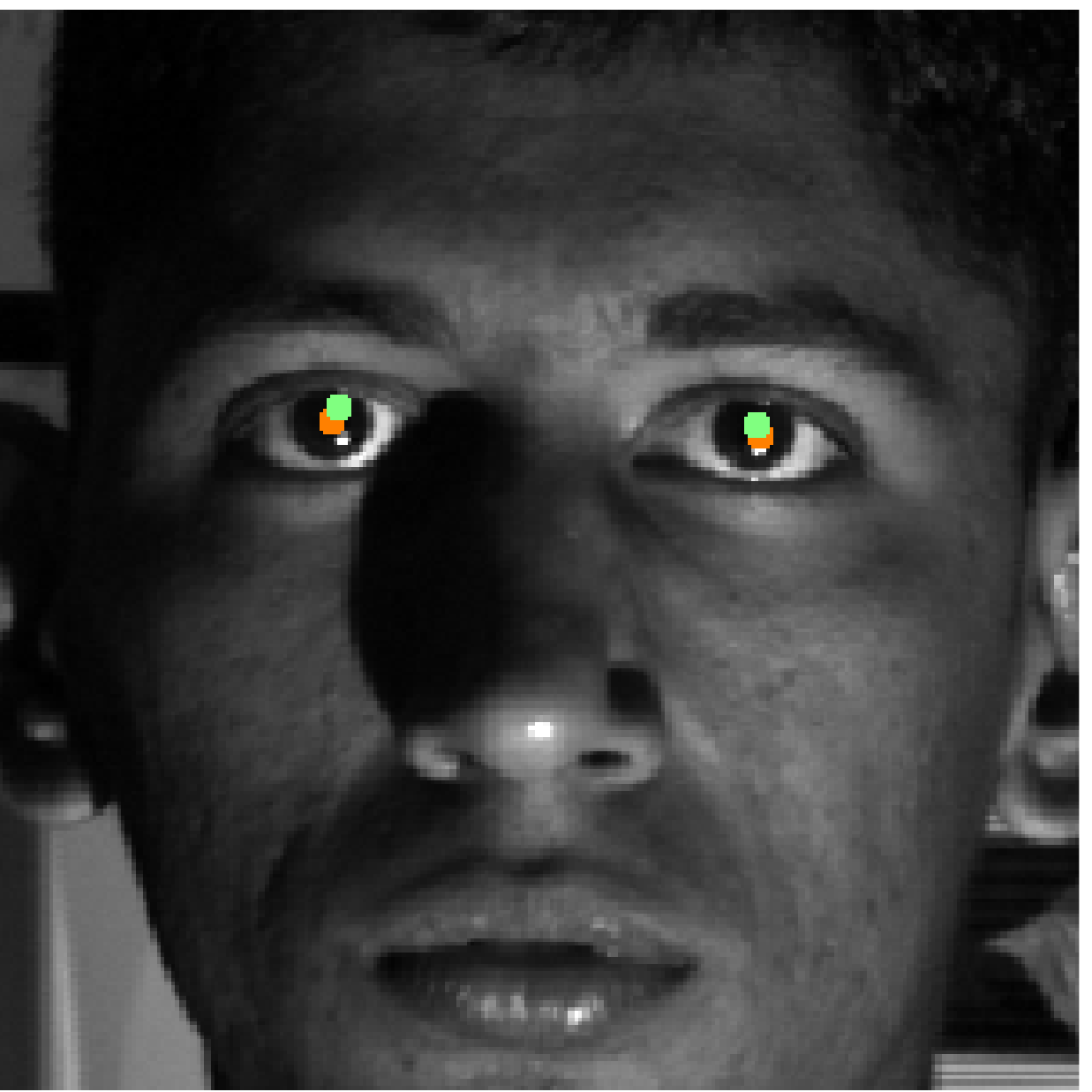}\\
        \includegraphics[width=0.14 \textwidth]{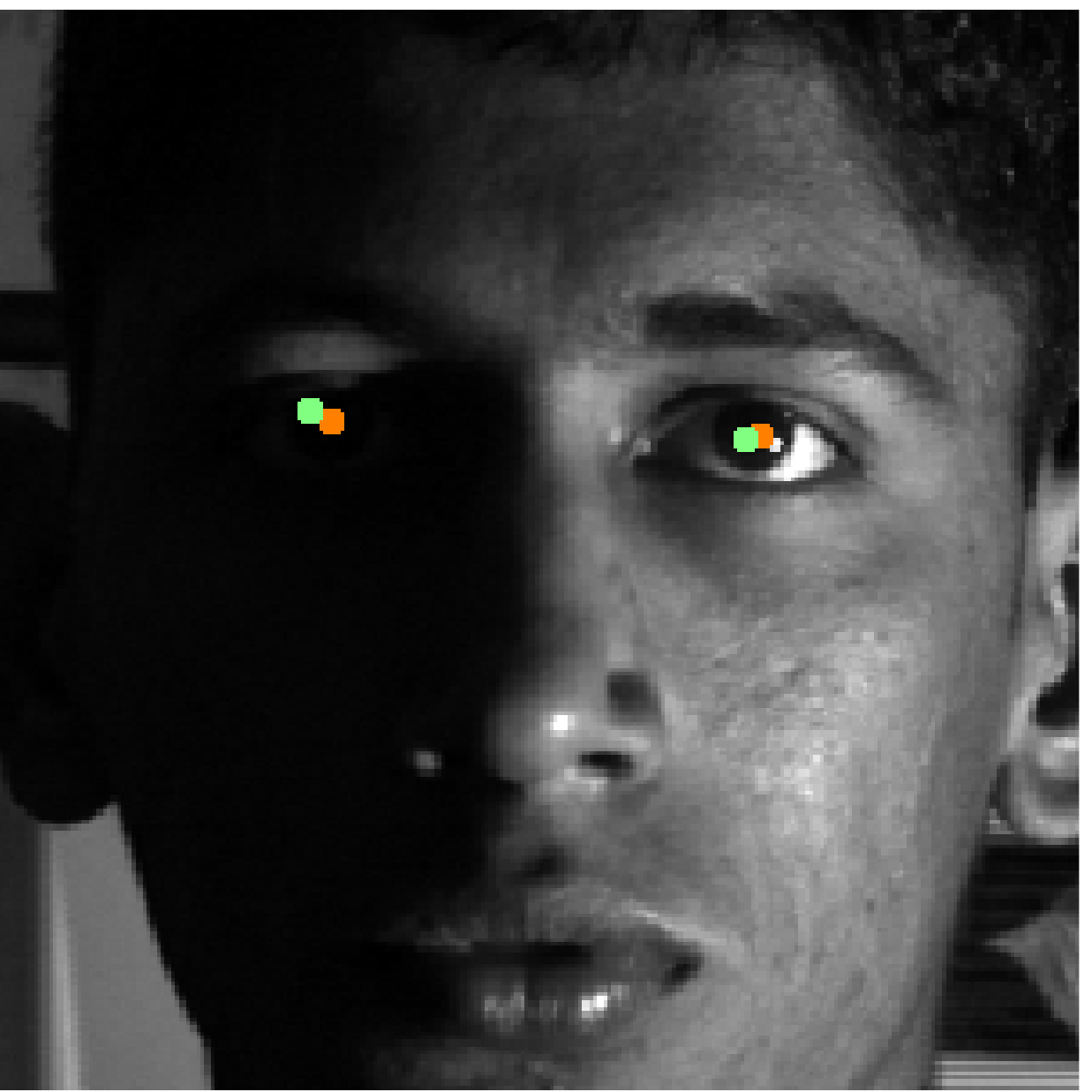}&
        \includegraphics[width=0.14 \textwidth]{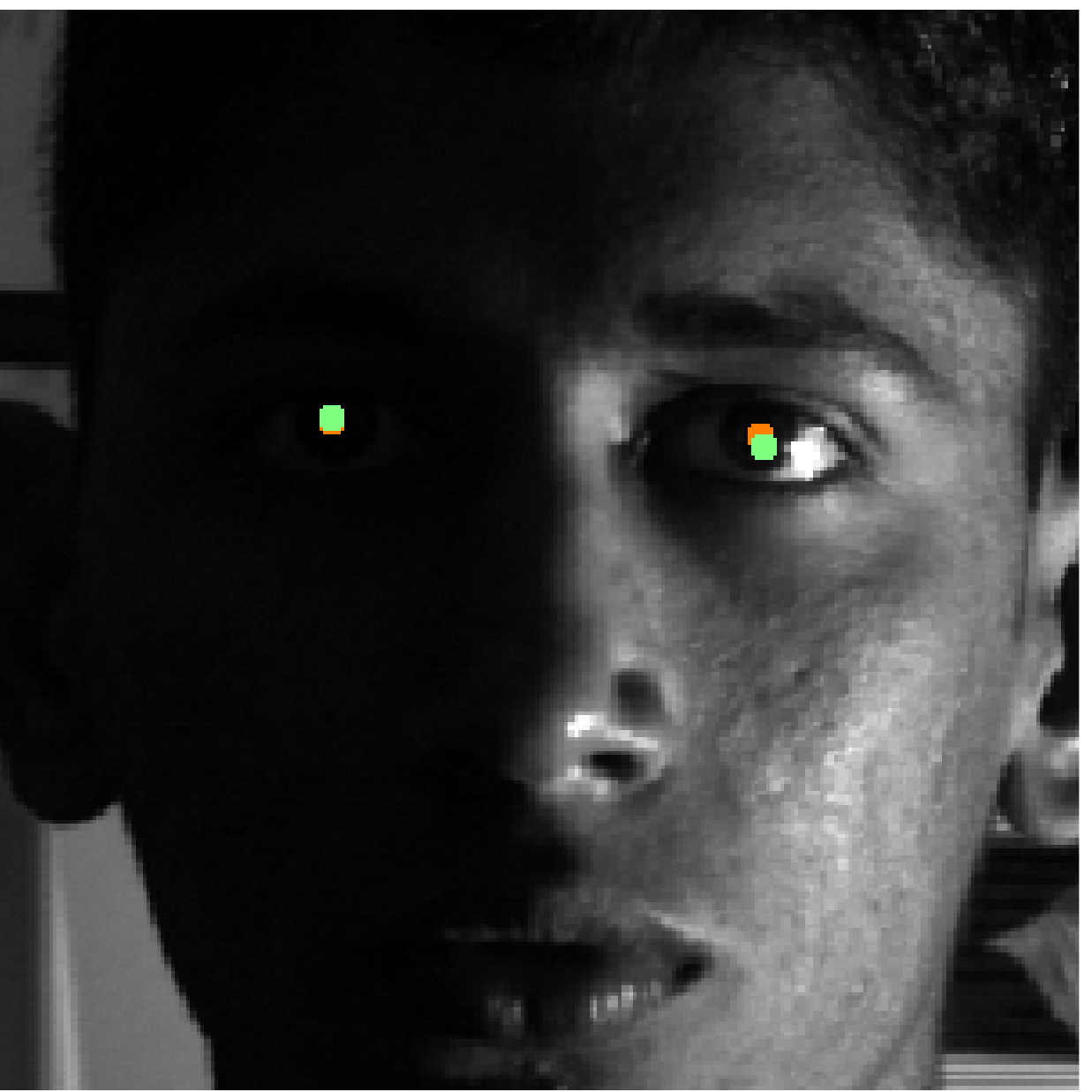}&
        \includegraphics[width=0.14 \textwidth]{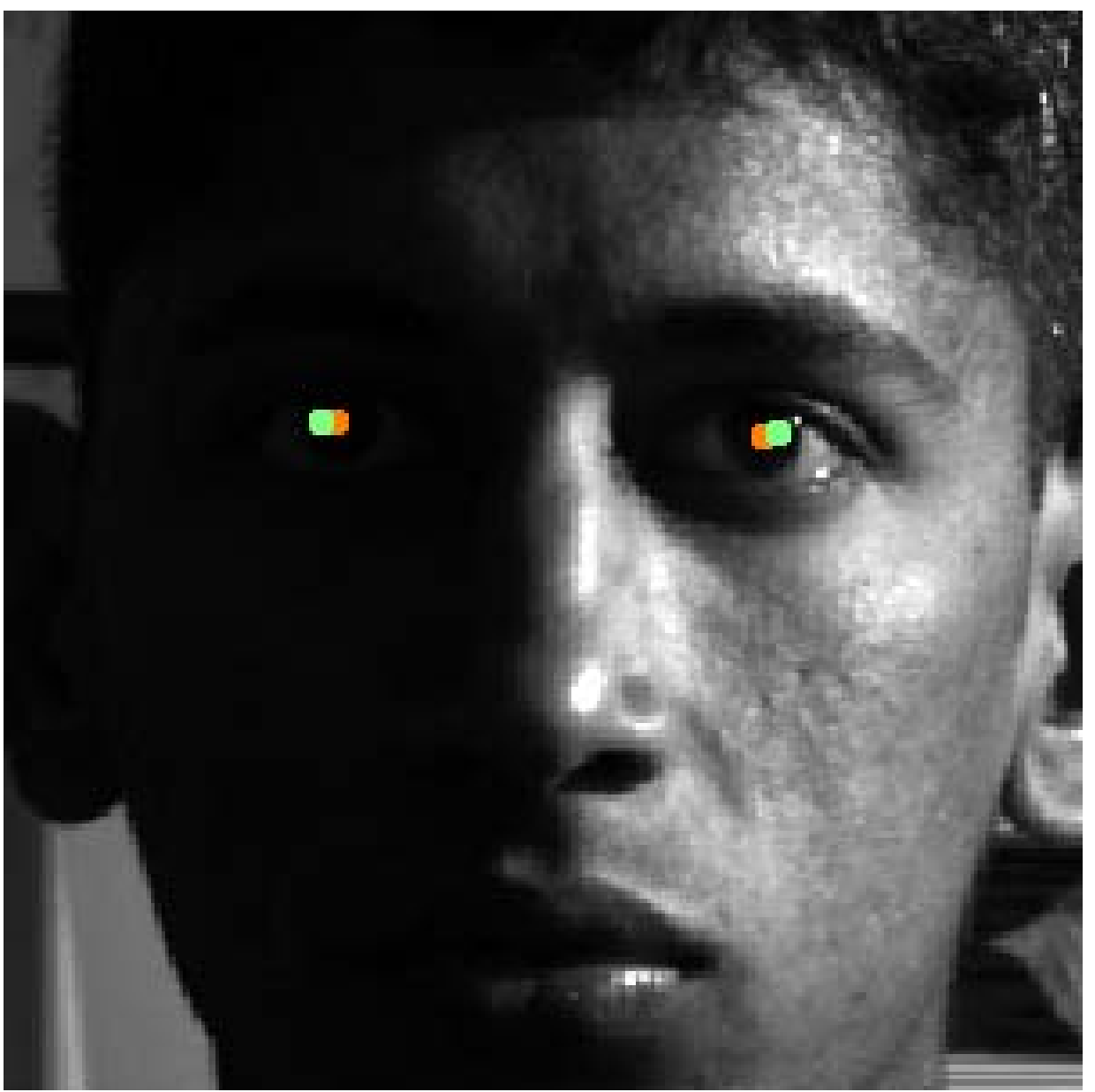}\\

        \\
    \end{tabular}
   \caption{Face cropped images from the Extended YaleB database showing robustness to illumination.}
   \label{Fig:ExampleYaleBIlum}
\end{figure}

The system achieves reasonable results in the cases when even a human observer is not able to
identify the eyes. As long as the illumination is constant over the eye, the system performs very well,
proving the invariance to uniform illumination of the ZEP feature claim. Examples of localization
while illumination varies are presented in figure \ref{Fig:ExampleYaleBIlum}.

\begin{figure}[tb]
\center
\begin{tabular}{ccc}
        \includegraphics[width=0.14 \textwidth]{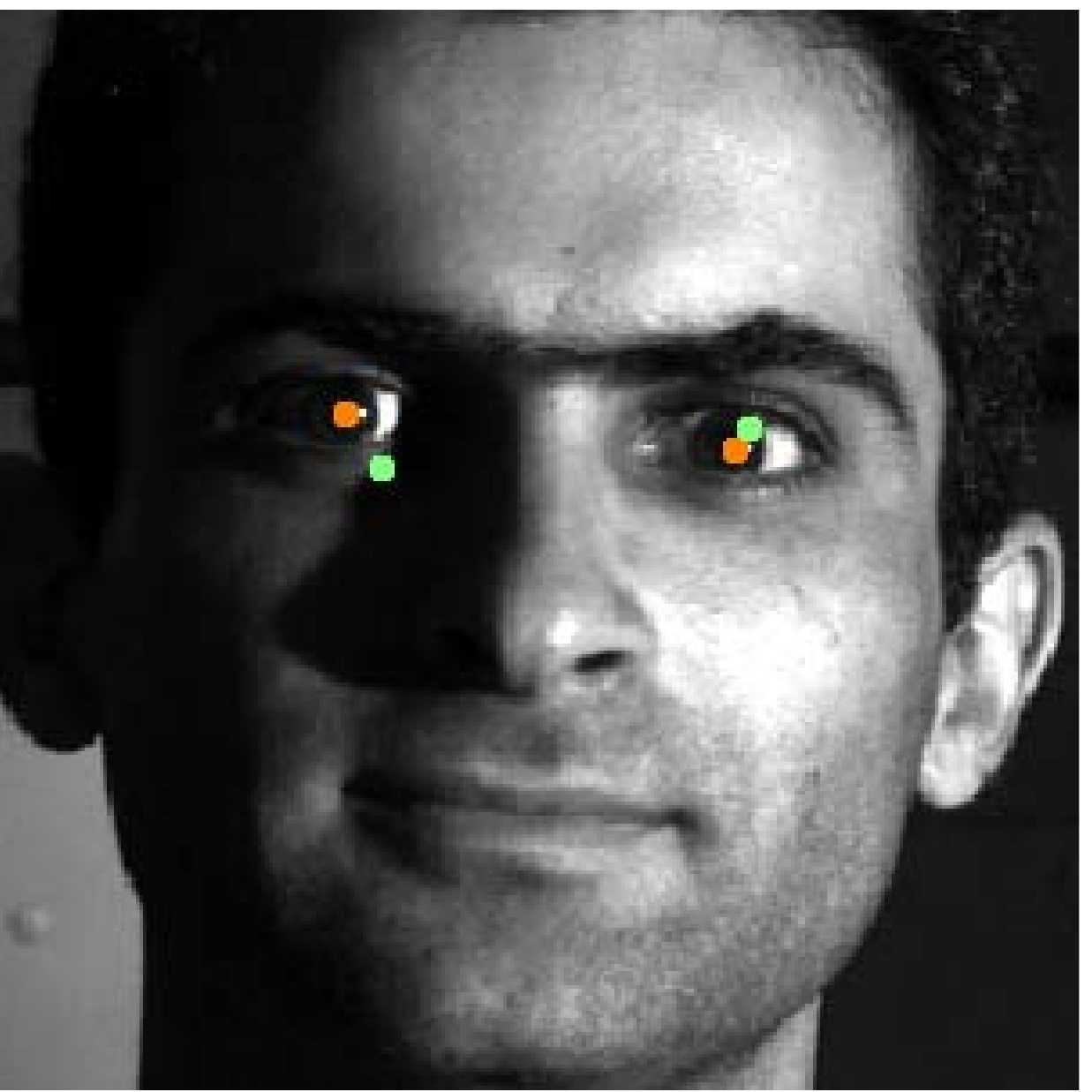}&
        \includegraphics[width=0.14 \textwidth]{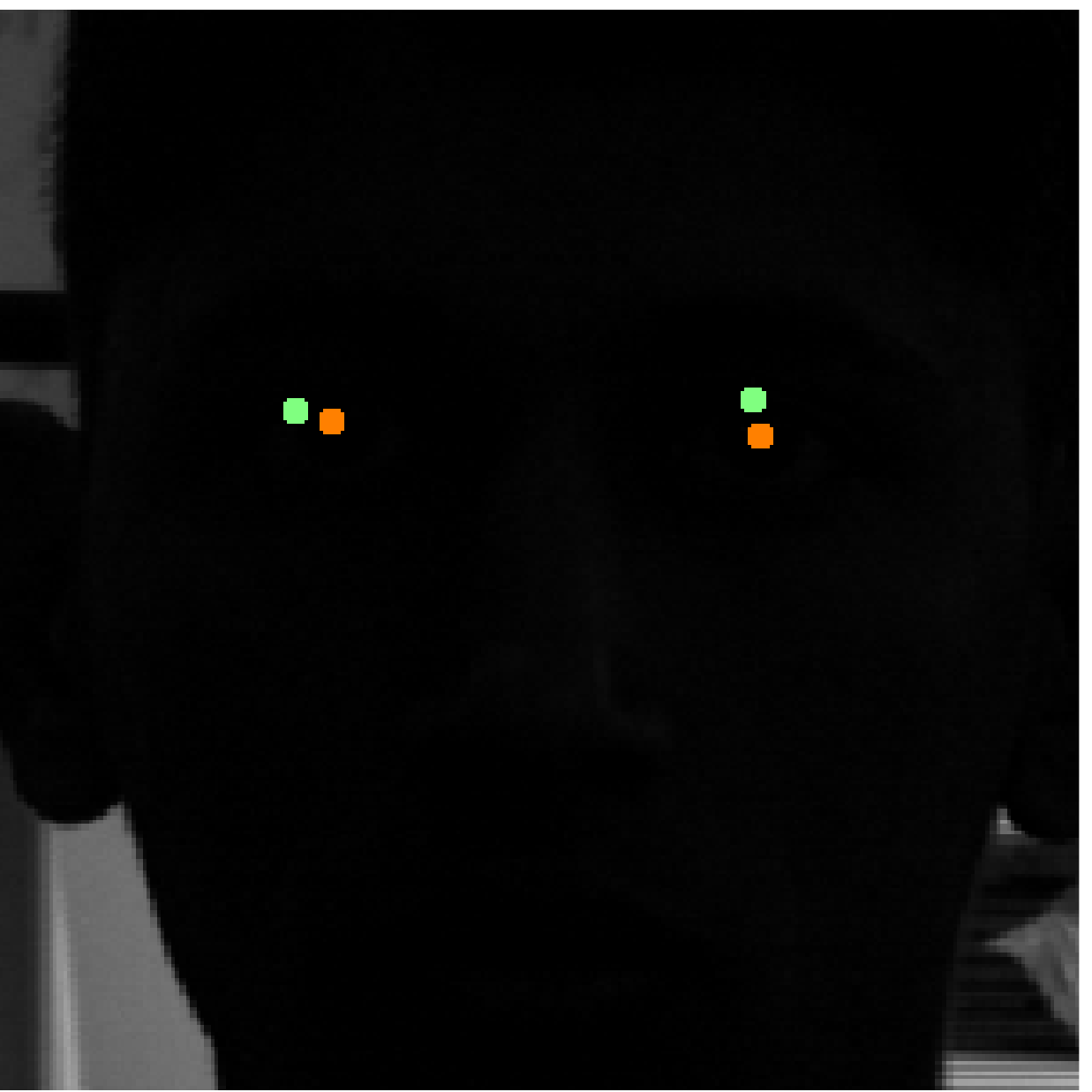}&
        \includegraphics[width=0.14 \textwidth]{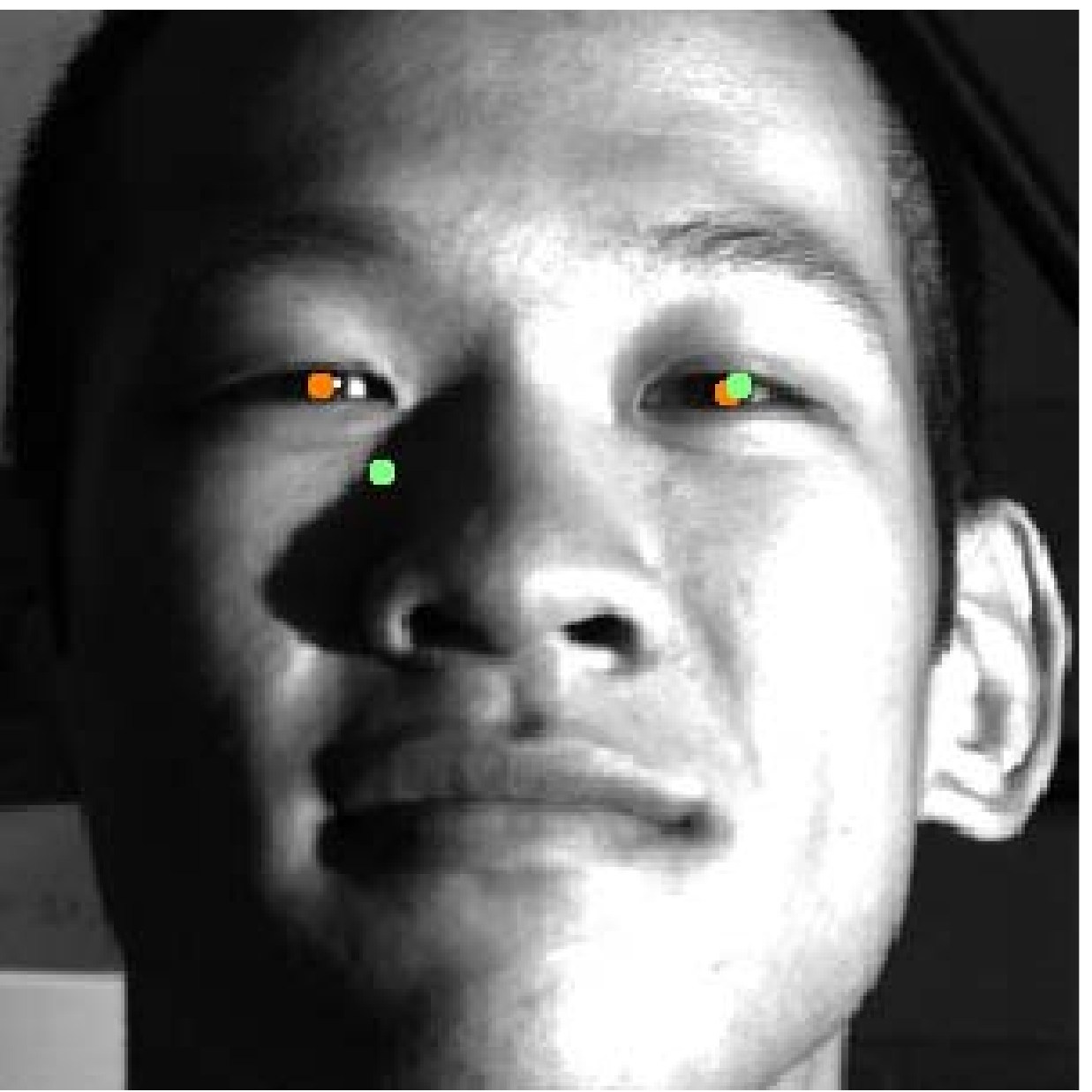}\\

        \begin{tabular}{c} $\mathcal{A} =-50^0$, \\ $\mathcal{E} =0^0$ \end{tabular}  &
        \begin{tabular}{c} $\mathcal{A} =+50^0$, \\ $\mathcal{E} =-40^0$ \end{tabular} &
        \begin{tabular}{c} $\mathcal{A} =-50^0$, \\ $\mathcal{E} =0^0$ \end{tabular}  \\

\end{tabular}
    \caption{Face cropped images from Extended Yale B database showing illumination cases that
    reveal limitations of the method. Specific shapes of projections caused by light and shade
    make the system prone to errors. Illumination angle is given by azimuth $\mathcal{A}$ and
    elevation $\mathcal{E}$. }
   \label{Fig:ExtYaleB_FunnyShade}
\end{figure}

For larger illumination angles, due to the uneven distribution of the shadows, the shape of the
projections is significantly altered and the accuracy decreases. Examples, with cases where the
shades are too strong or inopportunely placed and we reach lower results, are showed in figure
\ref{Fig:ExtYaleB_FunnyShade}.

\begin{table*}[tb]
\centering
 \caption{Illumination variation studied on the Extended Yale B (B+) data set. The numerical  values
  have been obtained for  accuracy  $\epsilon < 0.1 $. If more cases were available in the given
  interval, the average is reported. If the specific case does not exist in the database, ''n/a''
  is reported. }
  \label{Tab:Rez_YaleBIlum}

\begin{tabular}{| c | c | c | c | c | c | c |}
  \hline
    \begin{tabular}{c}
     Azimuth \\ \hline Elevation\\
     \end{tabular}
                        & $\pm [110^0:130^0]$  & $\pm[70^0 :  90^0]$& $\pm[50^0 : 60^0]$
                        & $\pm [20^0 : 35^0]$  & $\pm [5^0 : 15^0]$ & $0^0$   \\ \hline
     $-40^0 : -35^0 $   &{ }   n/a { } &{ } 73.21 { } &{ } 58.93 { }&{ } 60.71 { }& { }   n/a { } & { } 92.86 { }\\ \hline
     $-20^0 : -10^0 $   &{ } 71.43 { } &{ } 76.79 { } &{ } 83.93 { }&{ } 84.82 { }& { } 91.96 { } & { } 96.43 { }\\ \hline
     $ 0^0          $   &{ } 42.86 { } &{ } 75.00 { } &{ } 65.50 { }&{ } 94.64 { }& { } 91.07 { } & { } 92.86 { }\\ \hline
     $ 10^0 :  20^0 $   &{ } 67.86 { } &{ } 76.79 { } &{ } 75.87 { }&{ } 87.50 { }& { } 95.54 { } & { }   100 { }\\ \hline
     $ 40^0 :  45^0 $   &{ } 75.00 { } &{ } 76.79 { } &{ }   n/a { }&{ } 75.00 { }& { }   n/a { } & { } 89.29 { }\\ \hline
     $65^0 :   90^0 $   &{ } 82.14 { } &{ }  n/a  { } &{ }   n/a { }&{ } 64.29 { }& { }   n/a { } & { } 78.57 { }\\ \hline
\end{tabular}
\end{table*}

\begin{figure}[tb]
\center
    \begin{tabular}{ccc }
        \includegraphics[width=0.14 \textwidth]{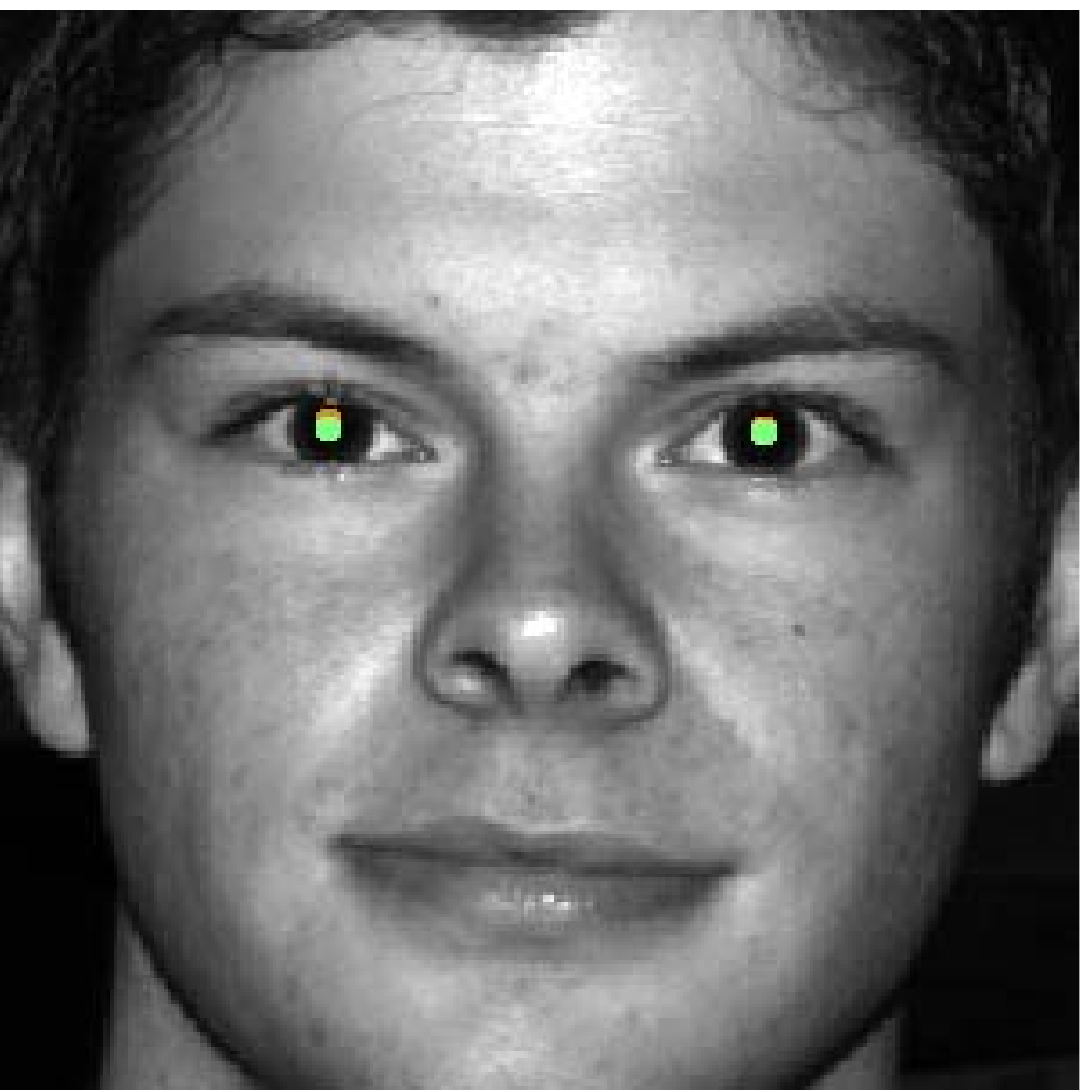}&
        \includegraphics[width=0.14 \textwidth]{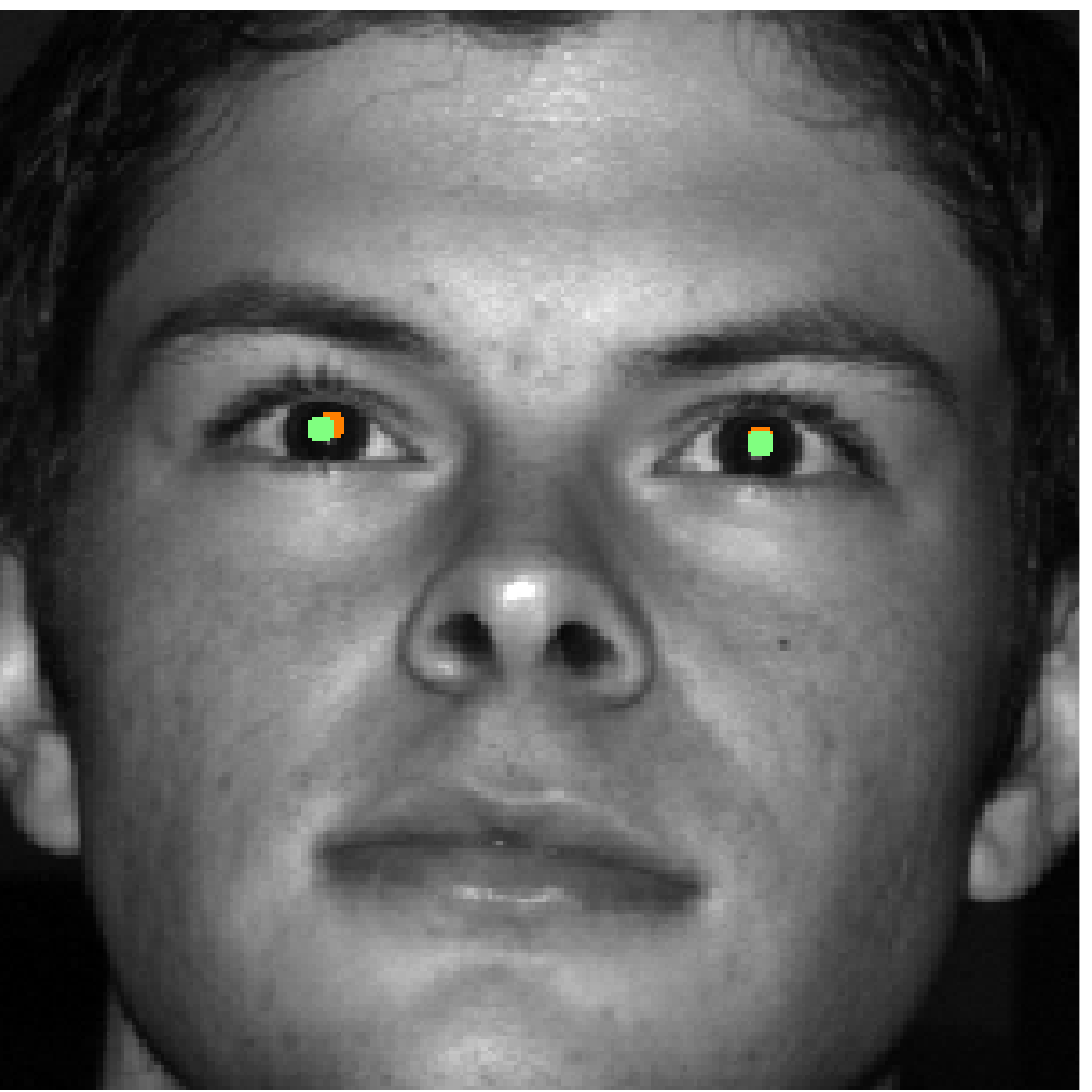}&
        \includegraphics[width=0.14 \textwidth]{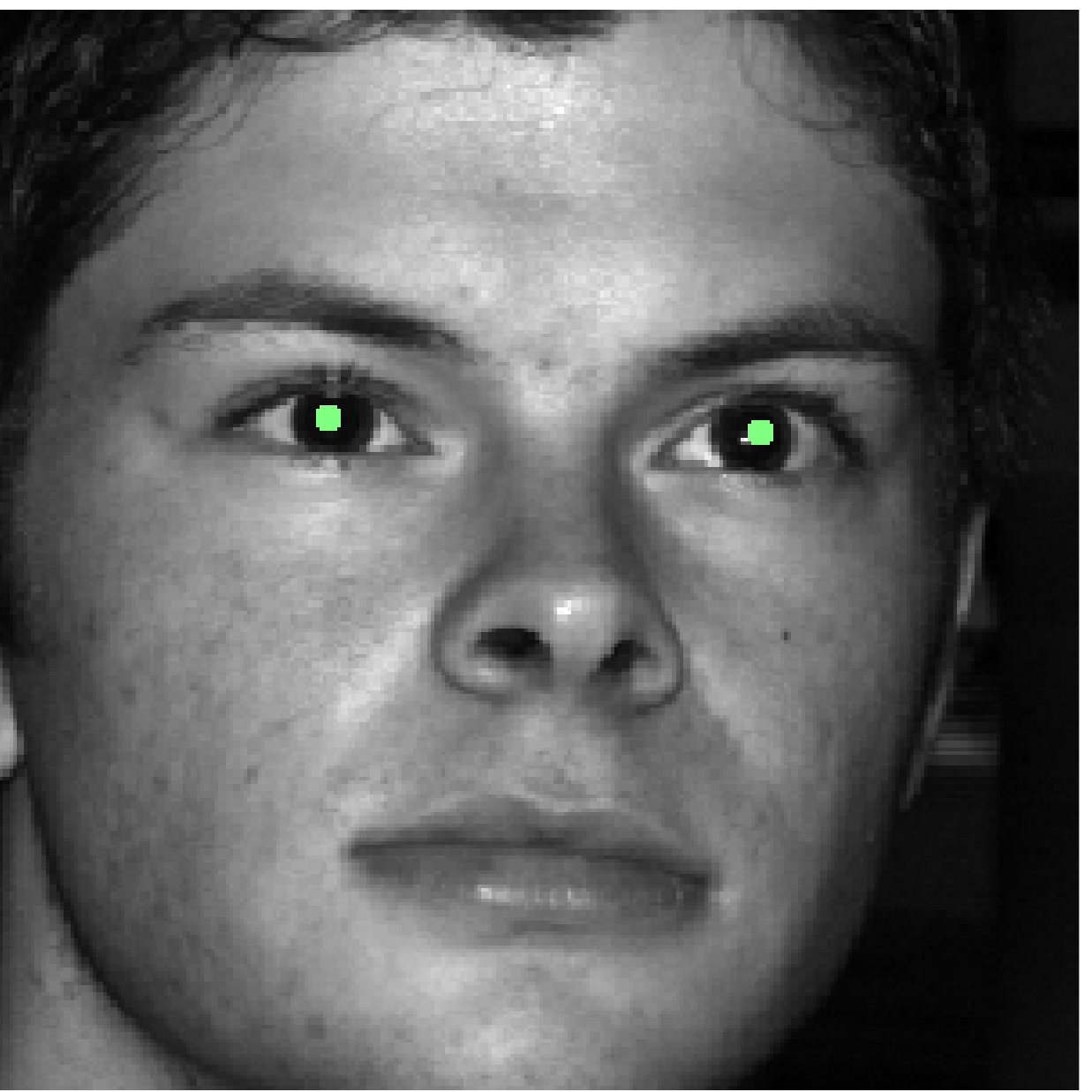}\\

        \includegraphics[width=0.14 \textwidth]{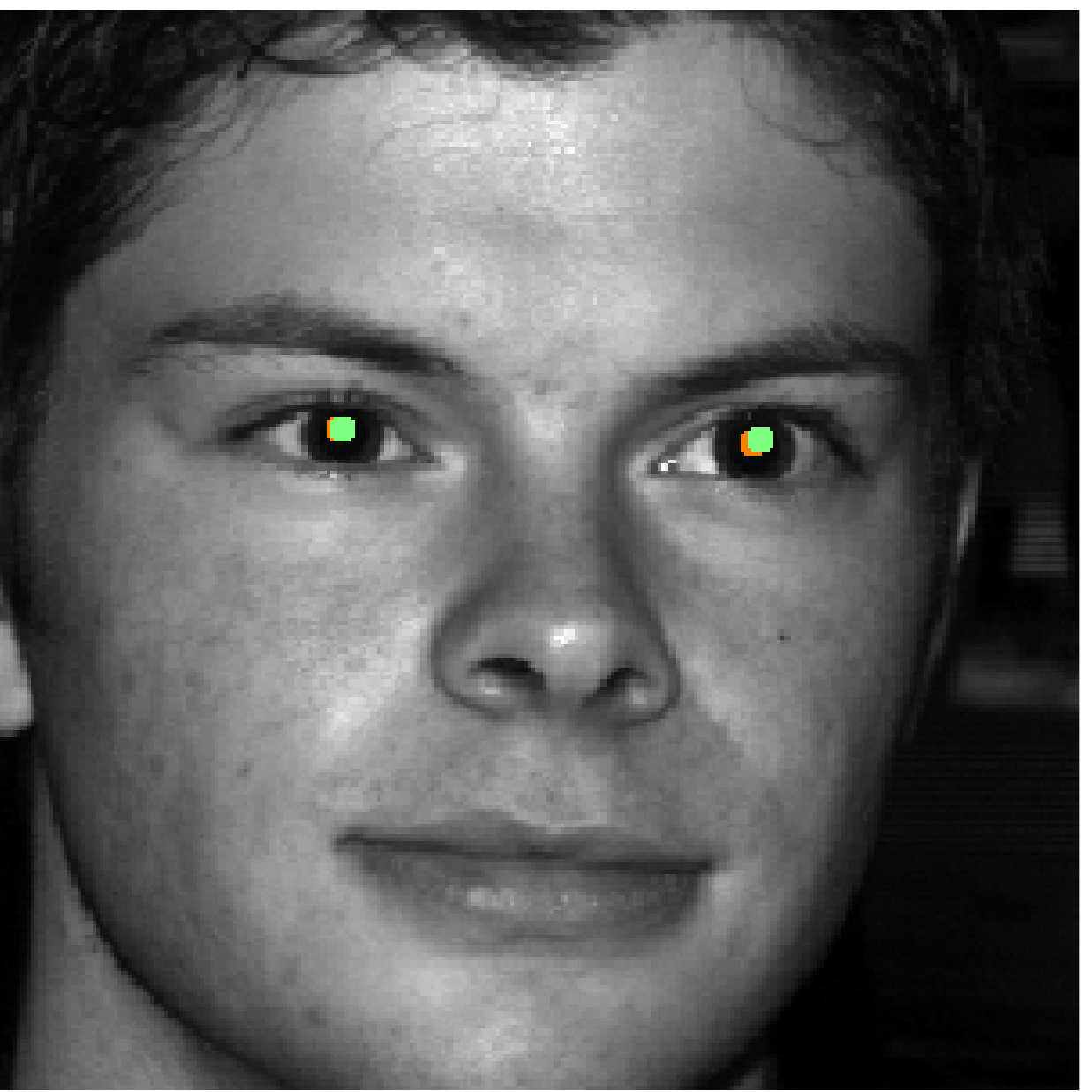}&
        \includegraphics[width=0.14 \textwidth]{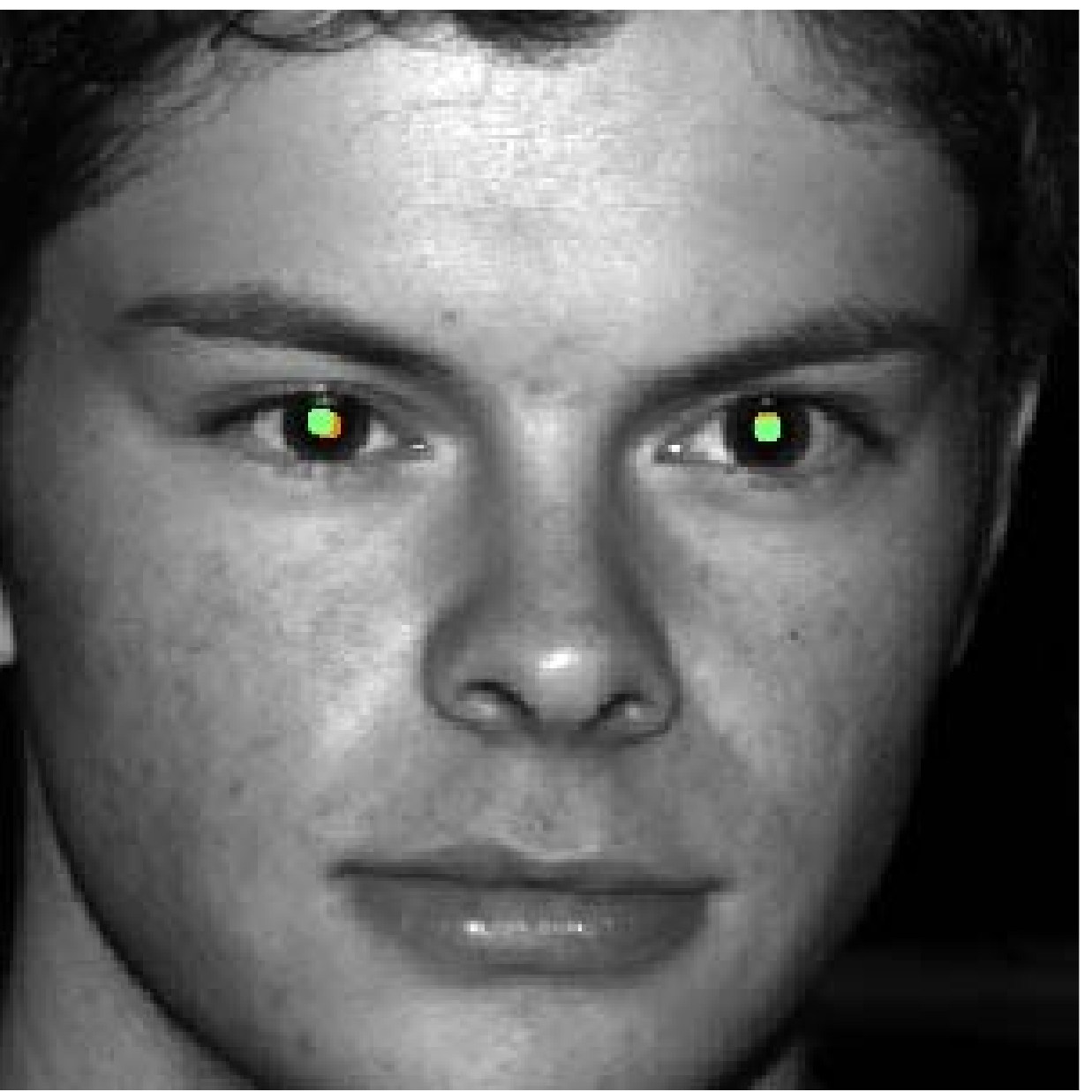}&
        \includegraphics[width=0.14 \textwidth]{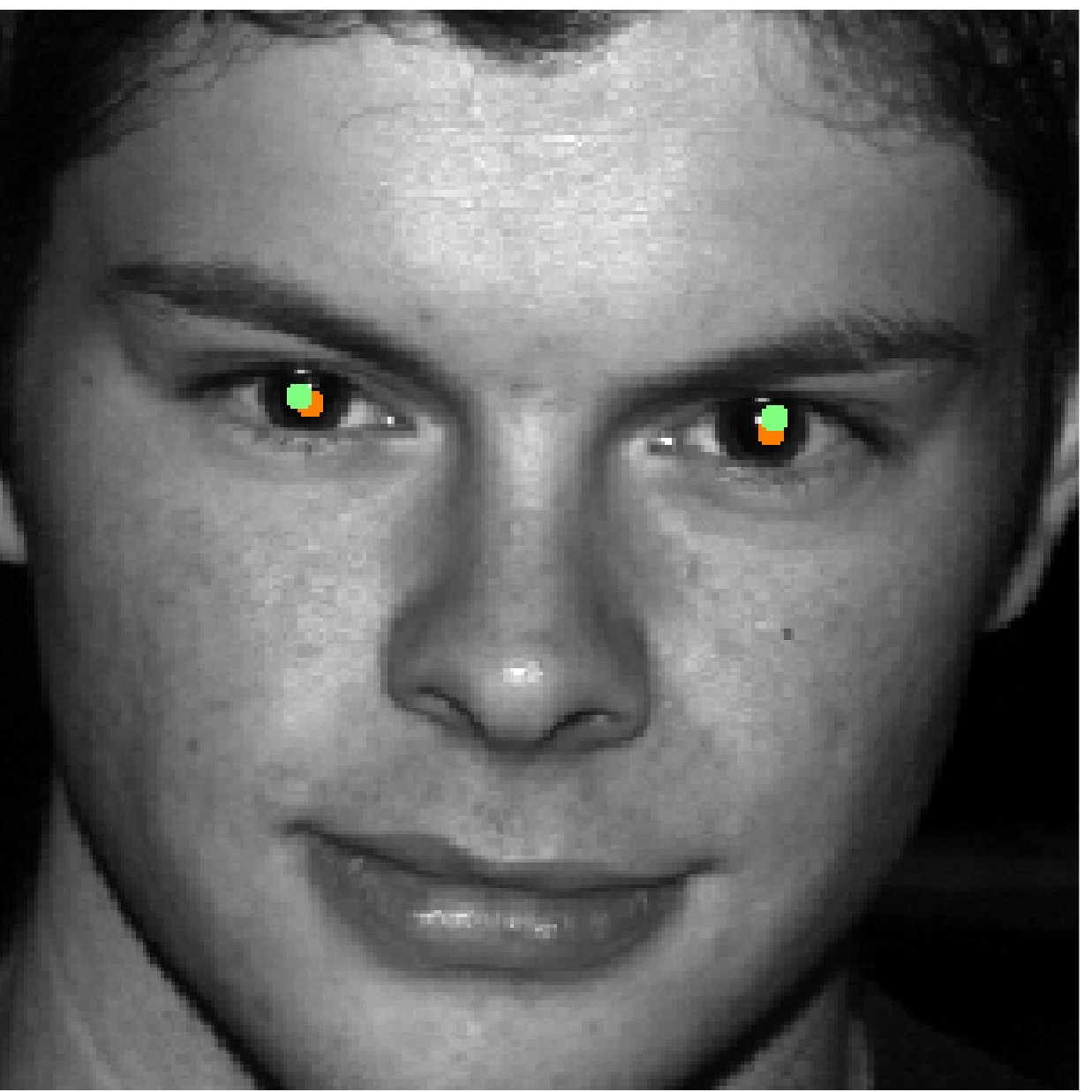}\\

        \includegraphics[width=0.14 \textwidth]{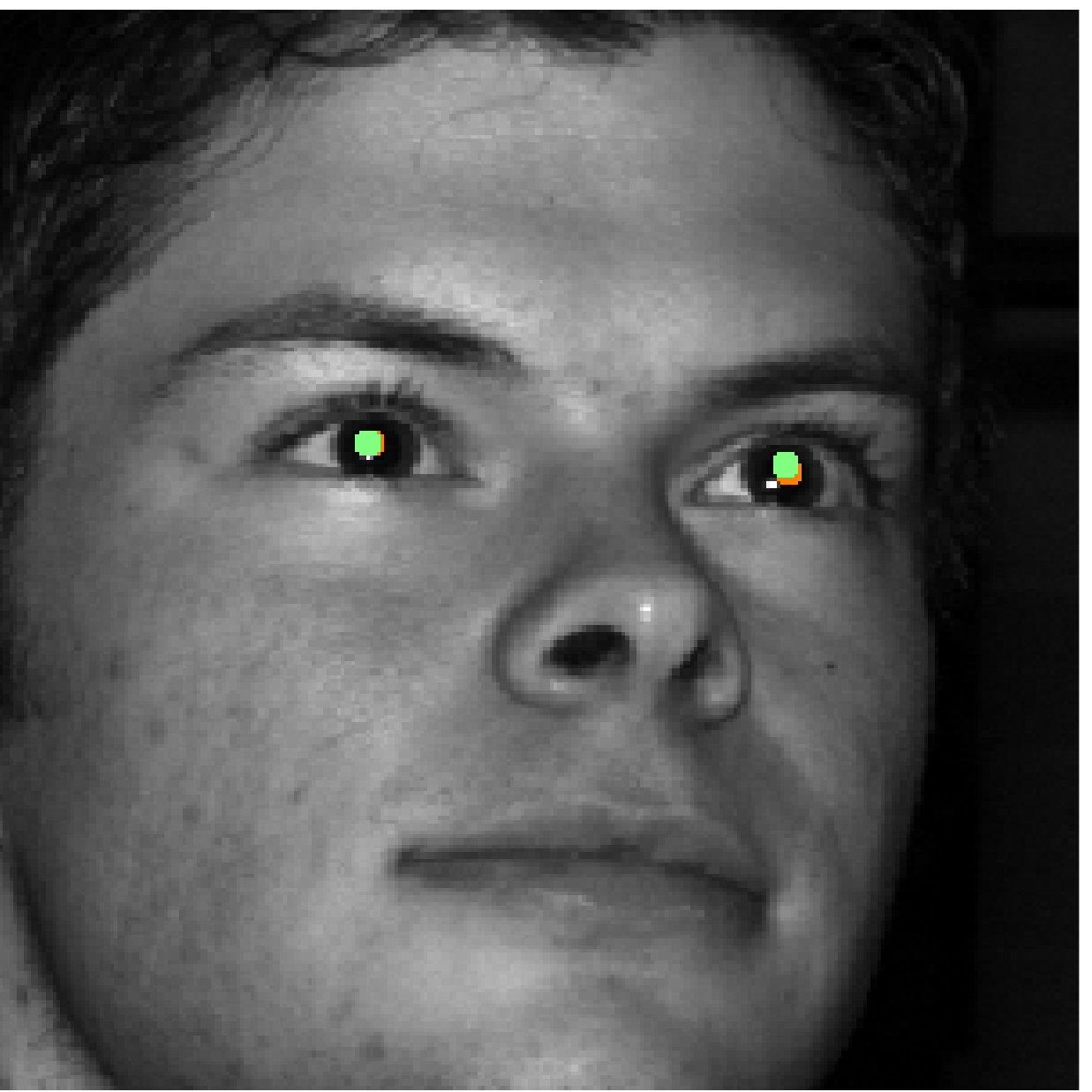}&
        \includegraphics[width=0.14 \textwidth]{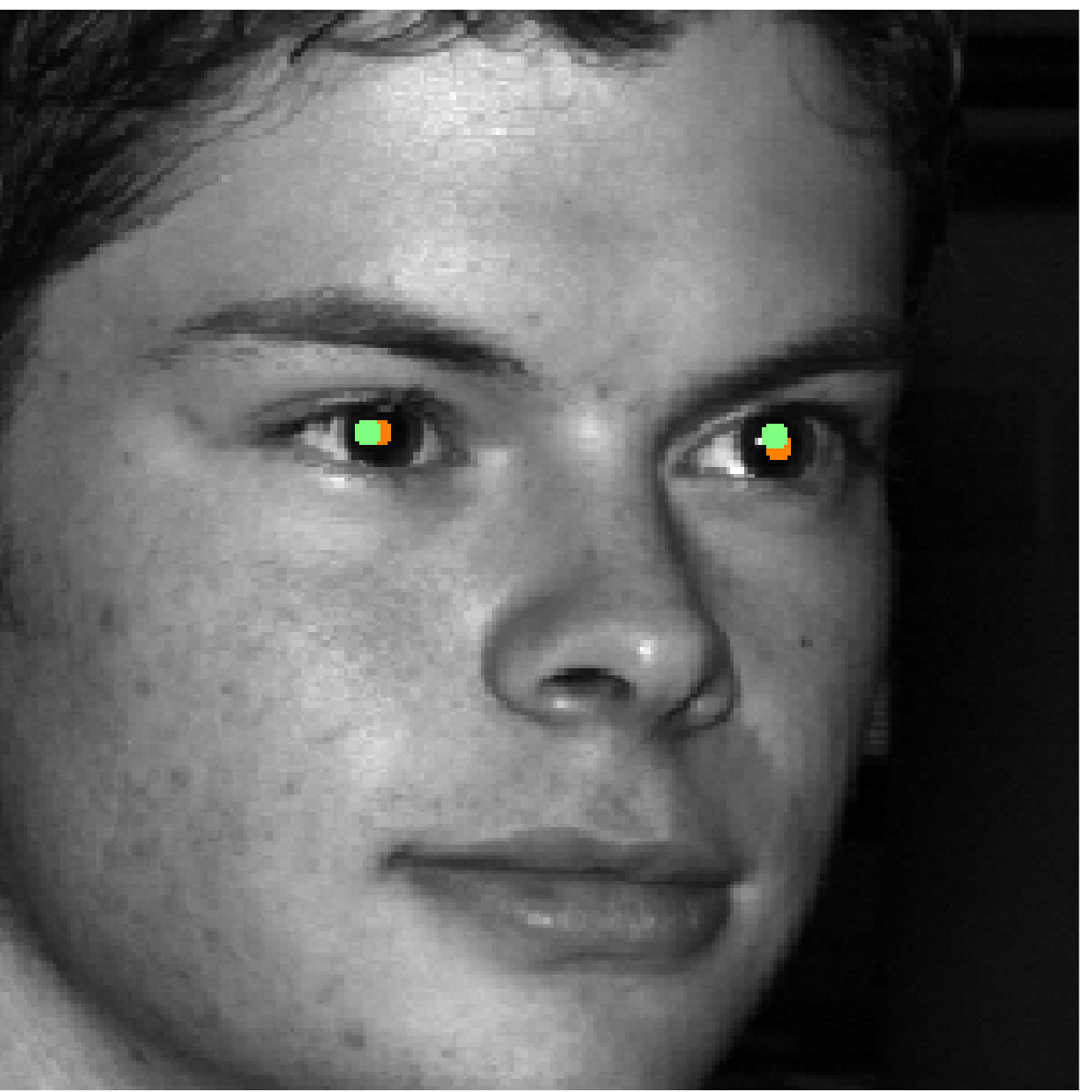}&
        \includegraphics[width=0.14 \textwidth]{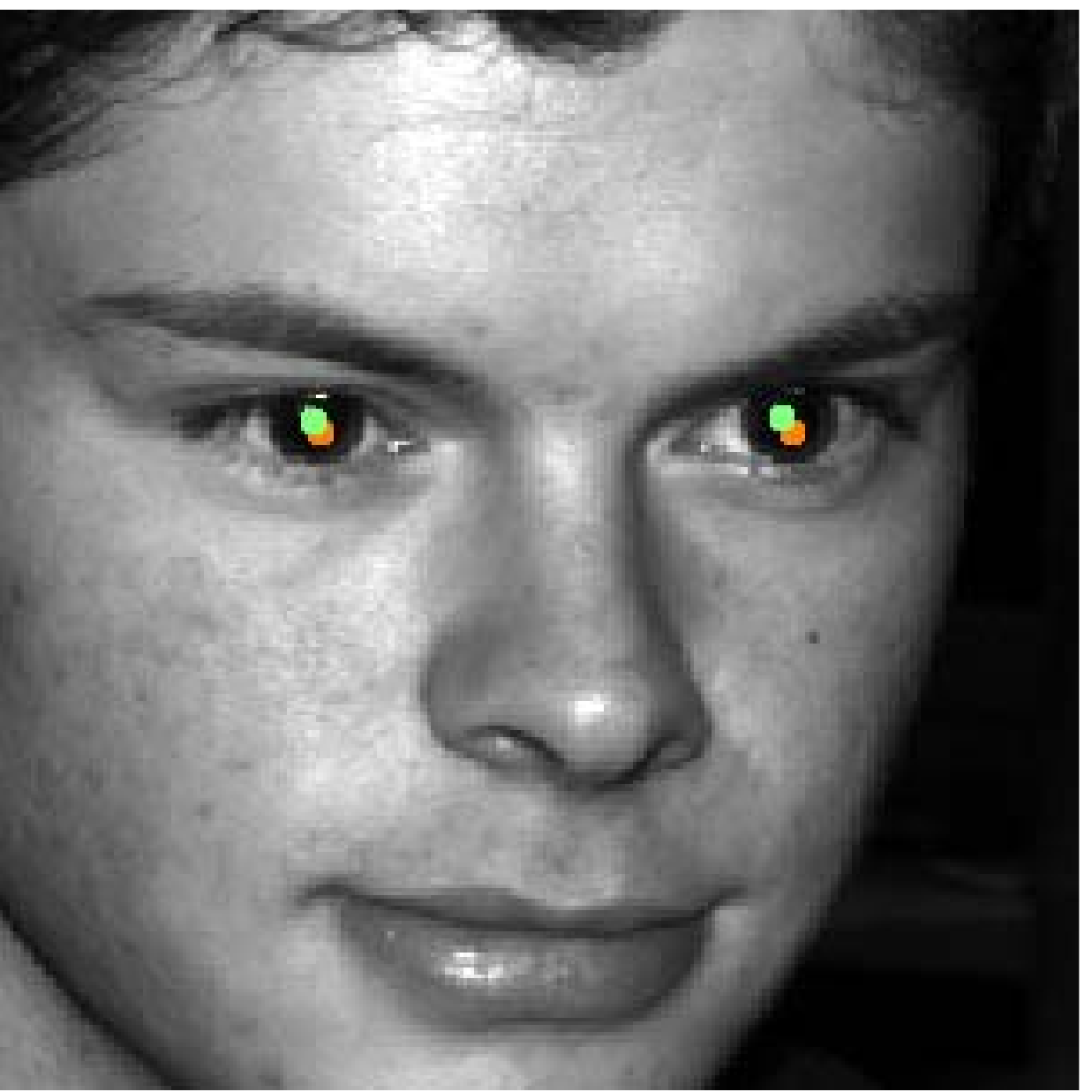} \\
    \end{tabular}
   \caption{Face cropped images from Extended YaleB database showing robustness to pose.}
   \label{Fig:ExampleYaleBPose}
\end{figure}

To evaluate the robustness of the algorithm with respect to the face pose, we consider each of the
28 persons with frontal illumination, but under varying poses (9 poses for each person). Pose
angles are in the set $\{0^0, 12^0, 24^0 \}$, thus spanning the typical range for ``frontal face''.
The results are shown in table \ref{Tab:RezPos_YaleB} and visual examples in the figure
\ref{Fig:ExampleYaleBPose}. Taking into account that the maximum number of images that have the
worst eye less accurate than $0.1$ is 2, we may truthfully say that the proposed method is robust
to face pose.

\begin{table}[tb]
\centering
 \caption{Pose variation studied on the Extended Yale B (B+) data set. The specified numerical
 values have been obtained for  accuracy  $\epsilon < 0.1$. For each category maximum two
 (corresponding to an accuracy of 92.86\%) images were missed.} \label{Tab:RezPos_YaleB}
\begin{tabular}{| c | c | c | c | }
  \hline
    \begin{tabular}{c}
     Azimuth \\ \hline Elevation\\
     \end{tabular}
                & $24^0$      &  $12^0$     &  Frontal   \\ \hline
     Up         &{ } 96.43 { }&{ } 92.86 { }&{ } 96.43 { }\\ \hline
     Neutral    &{ } 96.43 { }&{ } 96.43 { }&{ } 92.86 { }\\ \hline
     Down       &{ } 92.86 { }&{ } 92.86 { }&{ } 96.43 { }\\ \hline
\end{tabular}
\end{table}

When compared with the method proposed in \cite{Valenti:12}, our solution performs marginally
better. If we consider the results reported in the mentioned paper, then the average result for
accuracy at $\epsilon < 0.1$ is 88.07\% computed on the smaller YaleB database
\cite{Georghiades:01} while our method reaches 89.85\% on the same subset of azimuth and elevation
illumination angles on the larger Extended YaleB database \cite{Lee:05}. If we compare the full
results on the entire Extended YaleB database (including extreme illumination cases) then our
method outperforms with small margin for high accuracies as shown in table
\ref{Tab:YaleB+Comparat}. Our method performs significantly better than the ones proposed by
\citeyear{Tim:11} and respectively \citeyear{Ding:10}.
\begin{table}[tb]
    \center
    \caption{Comparative results on the Extended YaleB database \label{Tab:YaleB+Comparat}}
    \begin{tabular}{|c|c|c|c|}
        \hline
        \textbf{Method} & $\epsilon < 0.05$   & $\epsilon < 0.1$ & $\epsilon < 0.25$ \\ \hline
        \emph{Proposed} & { } \colorbox{lightgray}{\emph{39.9}}  { } & { } \colorbox{lightgray}{\emph{67.3}} { } & { } \emph{97.3} \\ \hline
          \cite{Valenti:12}& { } 37.8  { } & { } 66.6 { } & { } \colorbox{lightgray}{98.5} \\ \hline
          \cite{Tim:11}   & { } 20.1  { } & { } 34.5 { } & { } 51.5 \\ \hline
           \cite{Ding:10}   & { } 19.7  { } & { } 47.8 { } & { } 58.6 \\ \hline
    \end{tabular}
\end{table}


\subsection{Accuracy in Real-Life Scenarios}
While BioID, Cohn-Kanade and Extended YaleB databases include specific variations as they are
acquired under controlled lighting conditions with only frontal faces, they cannot be considered
too closely resembling real-life applications. In contrast, there are databases like the Labeled
Face Parts in the Wild (LFPW) \cite{Belhumeur:11} and the Labeled Faces in the Wild (LFW)
\cite{LFW:07}, which are randomly gathered from the Internet, contain large variations in the
imaging conditions. While LFPW is annotated with facial point locations, only a subset of about
1500 images is made available and contains high resolution and rather qualitative images. In
opposition, the LFW database contains more than 12000  facial images, having the resolution
$250\times 250$ pixels, with 5700 individuals that have been collected ``in the wild'' and vary in
pose, lighting conditions, resolution, quality, expression, gender, race, occlusion and make-up.

The images difficulty is certified by the performance of human evaluation error as reported in
\cite{Dantone:12}, which also provided annotations. While the ground truth is taken as the average
of human markings for each point normalized to inter-ocular distance, human evaluation error is
considered as  the averaged displacement of the one marker.

\begin{figure}[tb]
\center
    \begin{tabular}{ccc }
        \includegraphics[width=0.14 \textwidth]{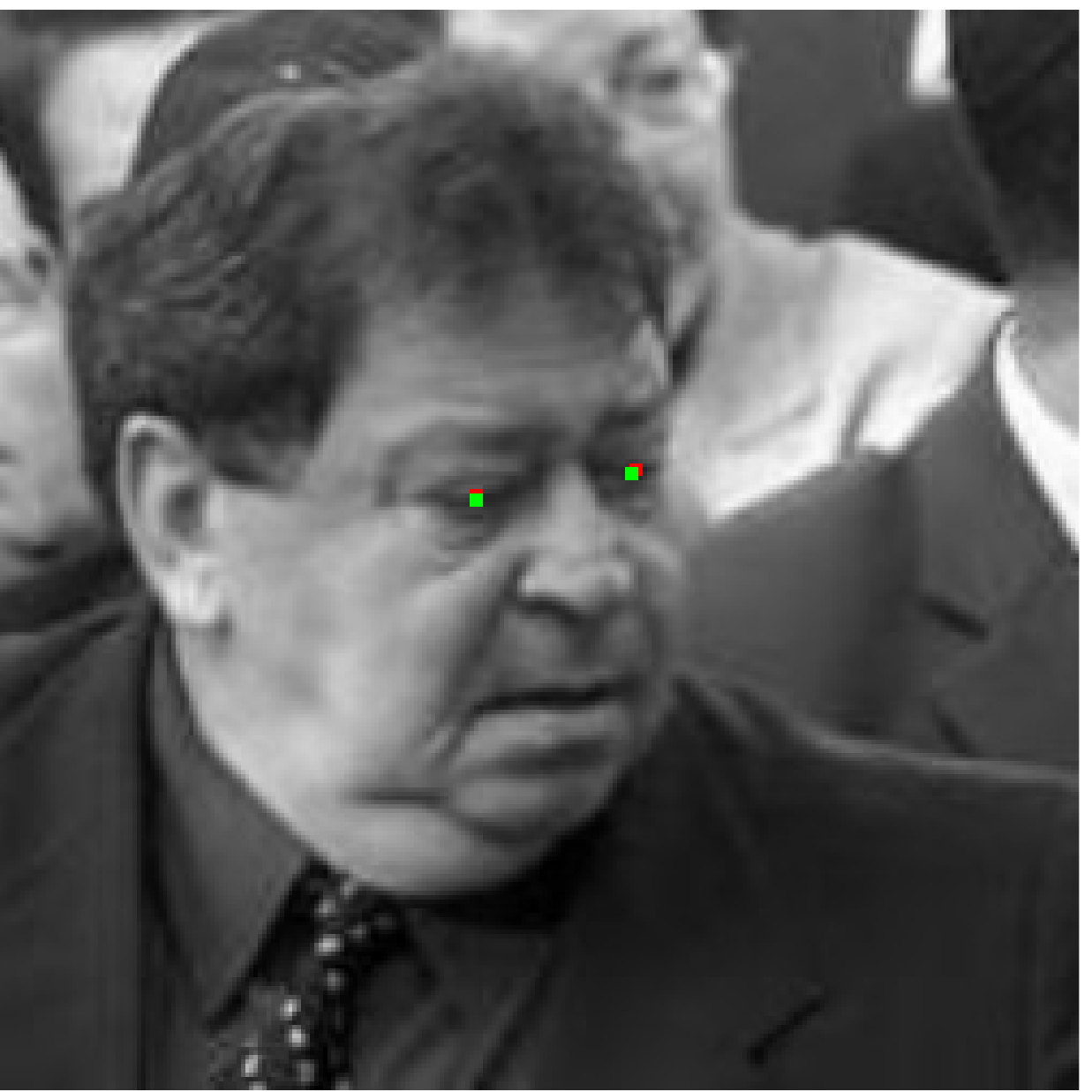}&
        \includegraphics[width=0.14 \textwidth]{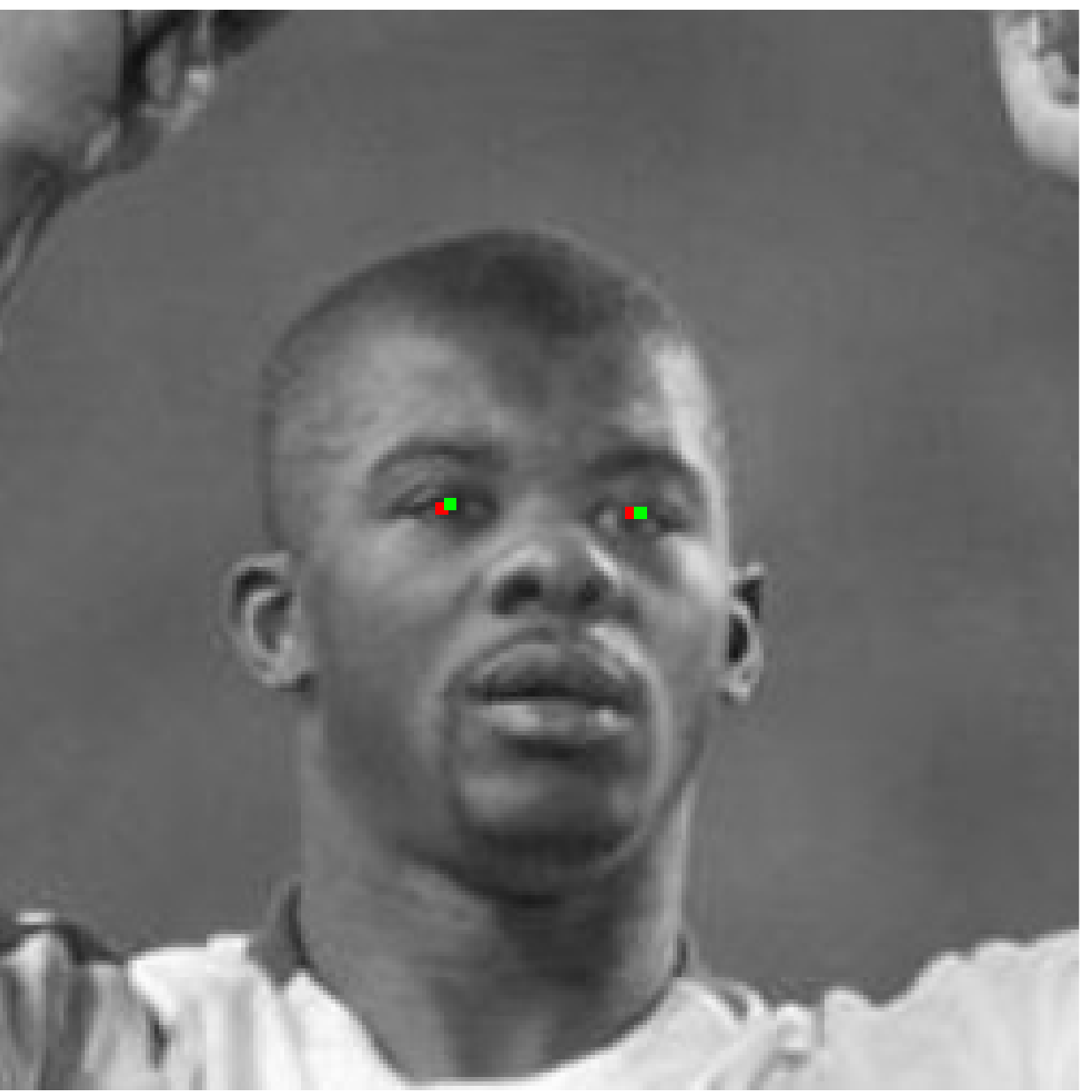}&
        \includegraphics[width=0.14 \textwidth]{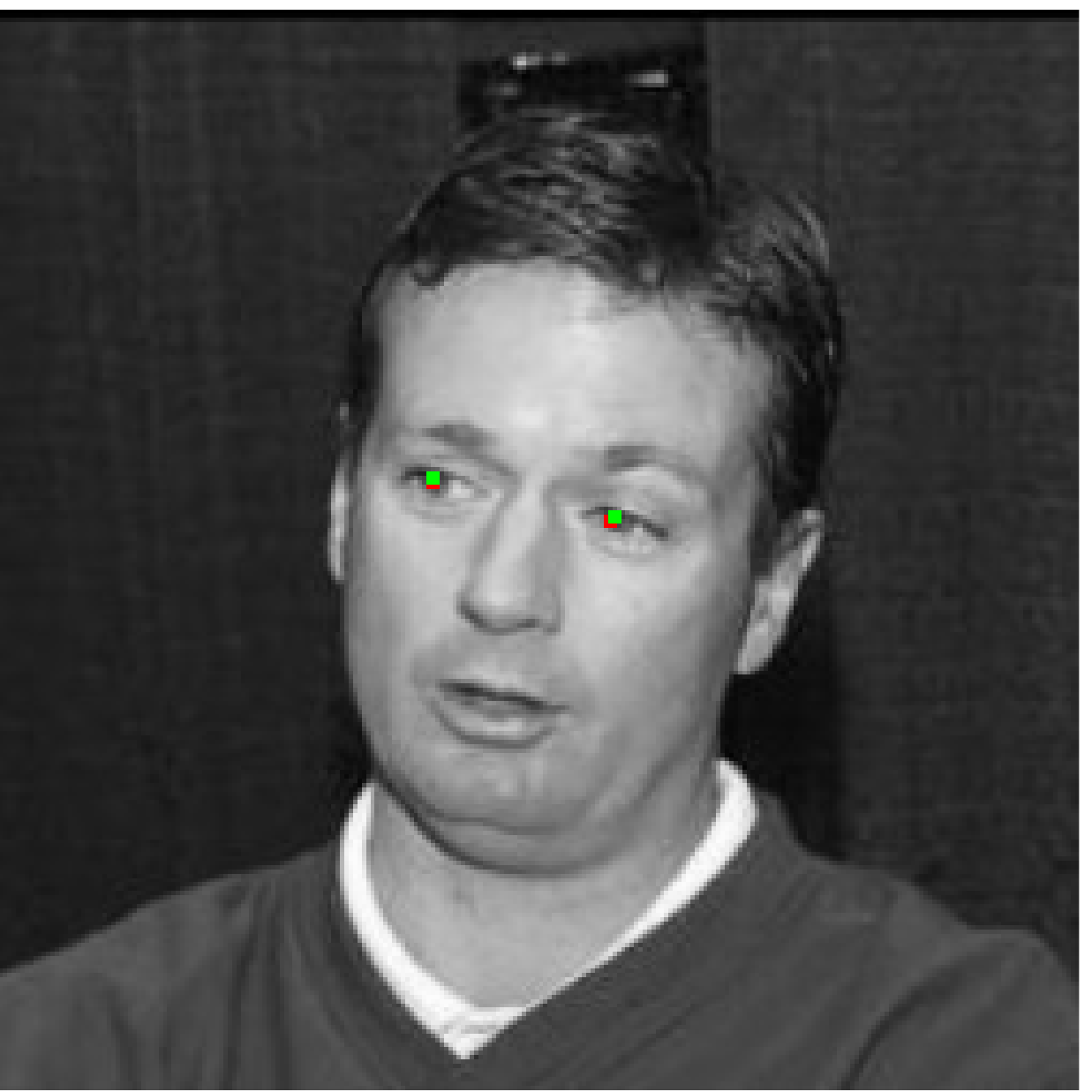}\\

        \includegraphics[width=0.14 \textwidth]{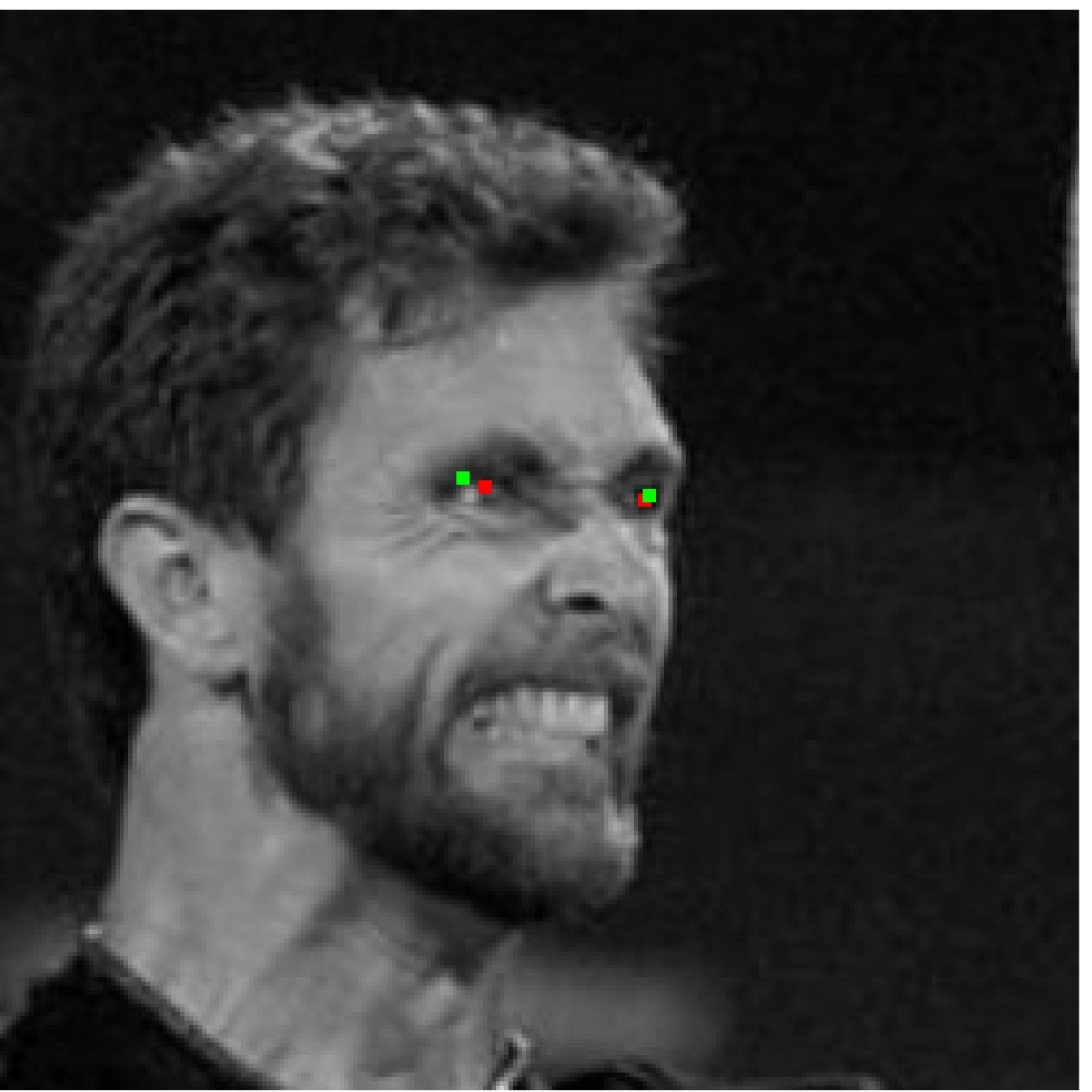}&
        \includegraphics[width=0.14 \textwidth]{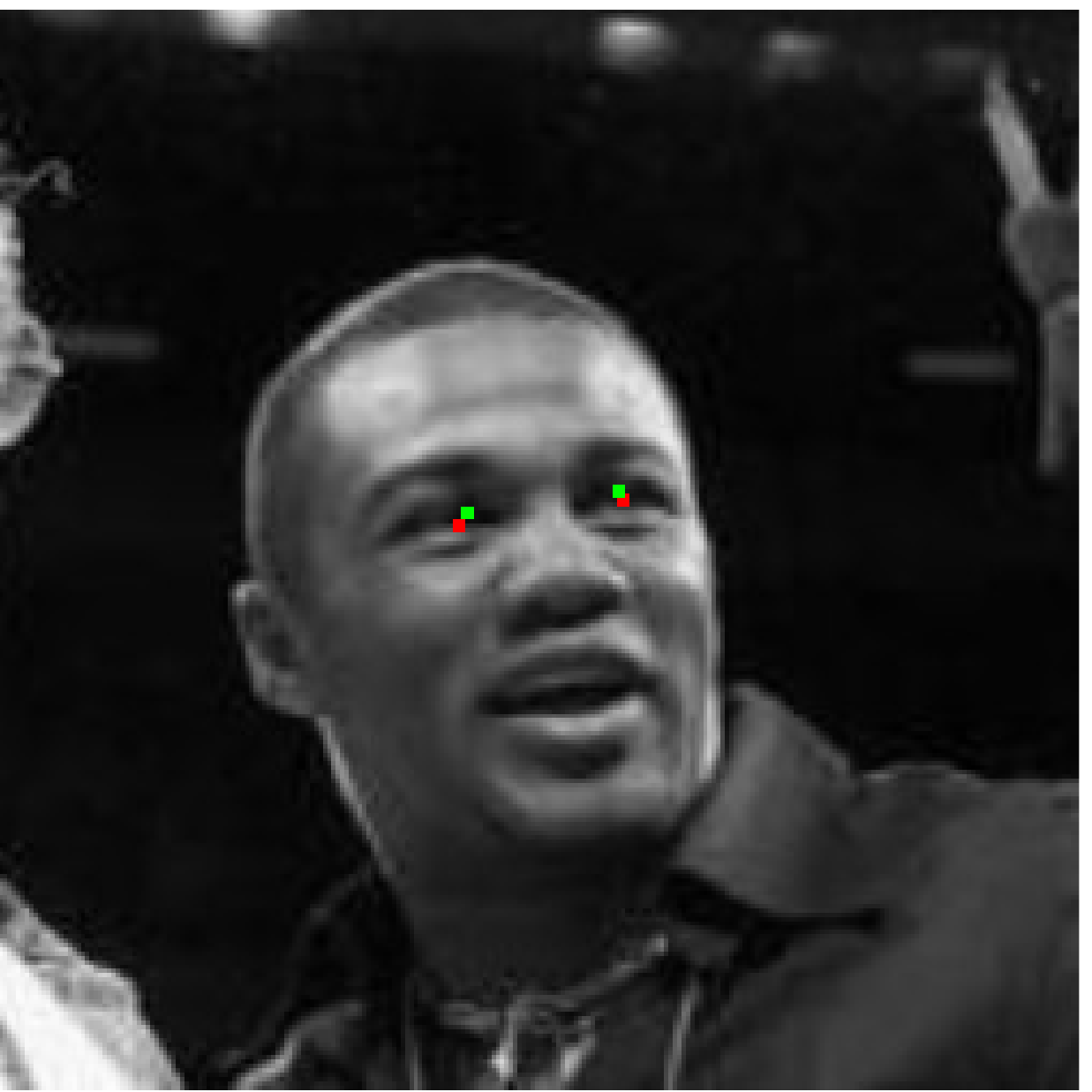}&
        \includegraphics[width=0.14 \textwidth]{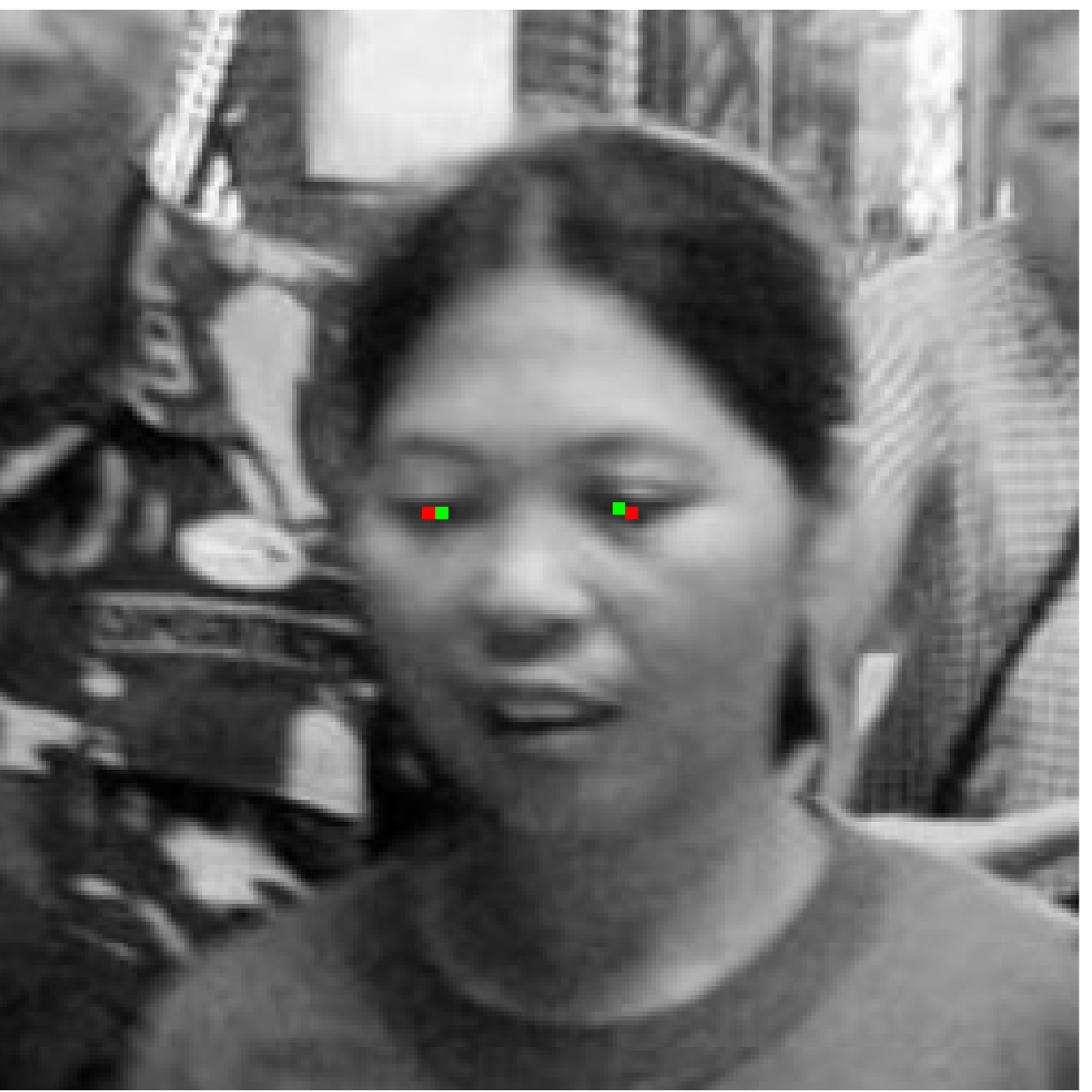}\\

        \includegraphics[width=0.14 \textwidth]{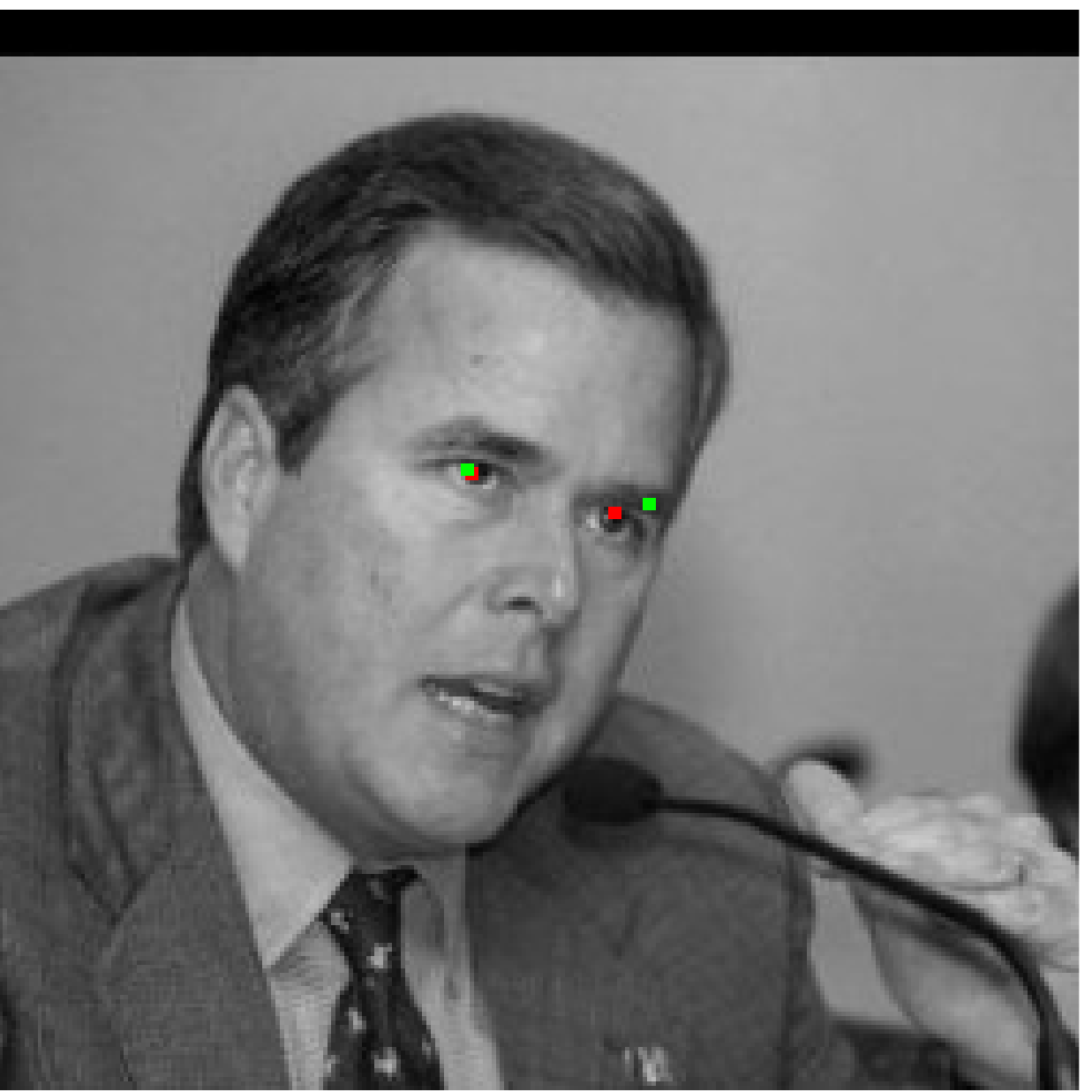}&
        \includegraphics[width=0.14 \textwidth]{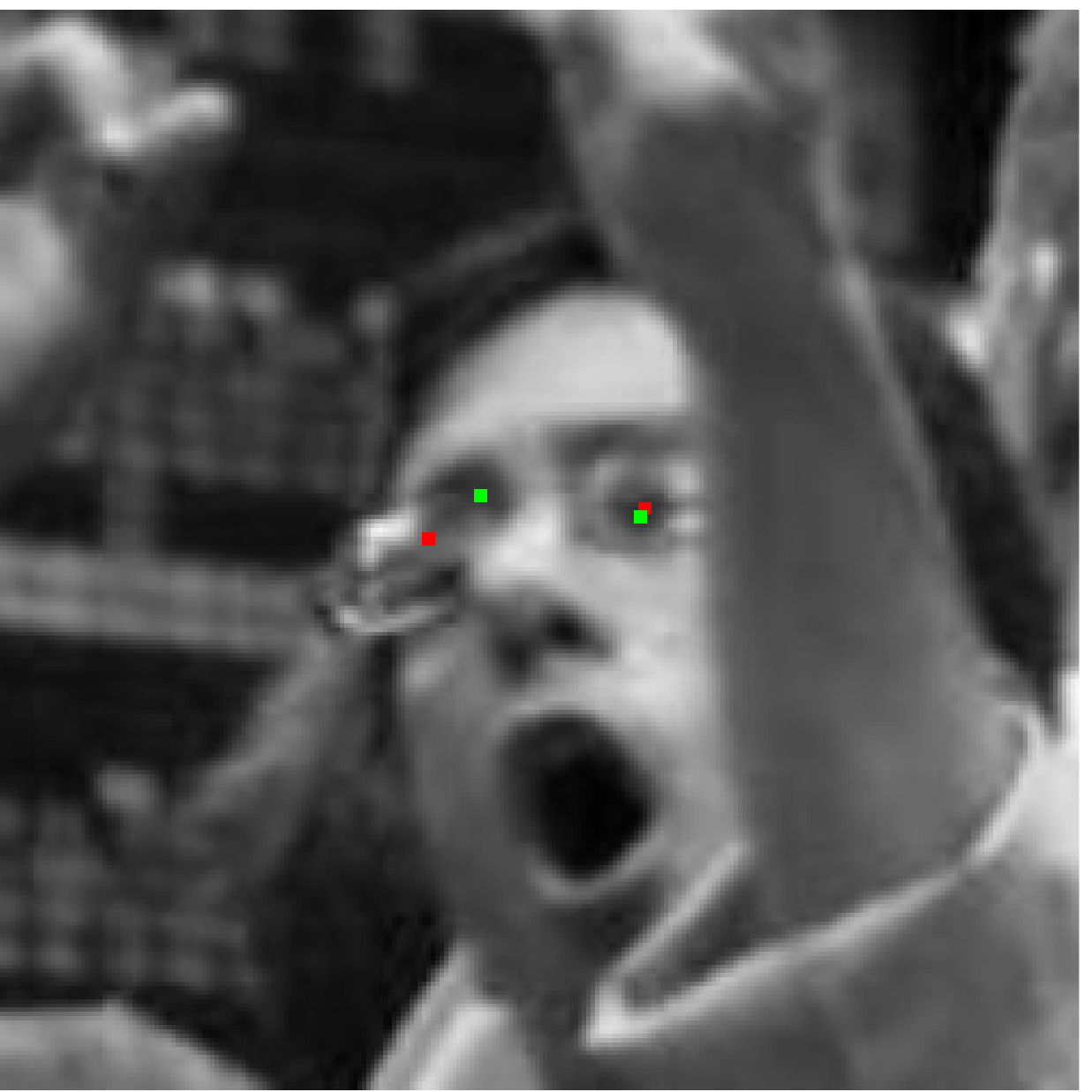}&
        \includegraphics[width=0.14 \textwidth]{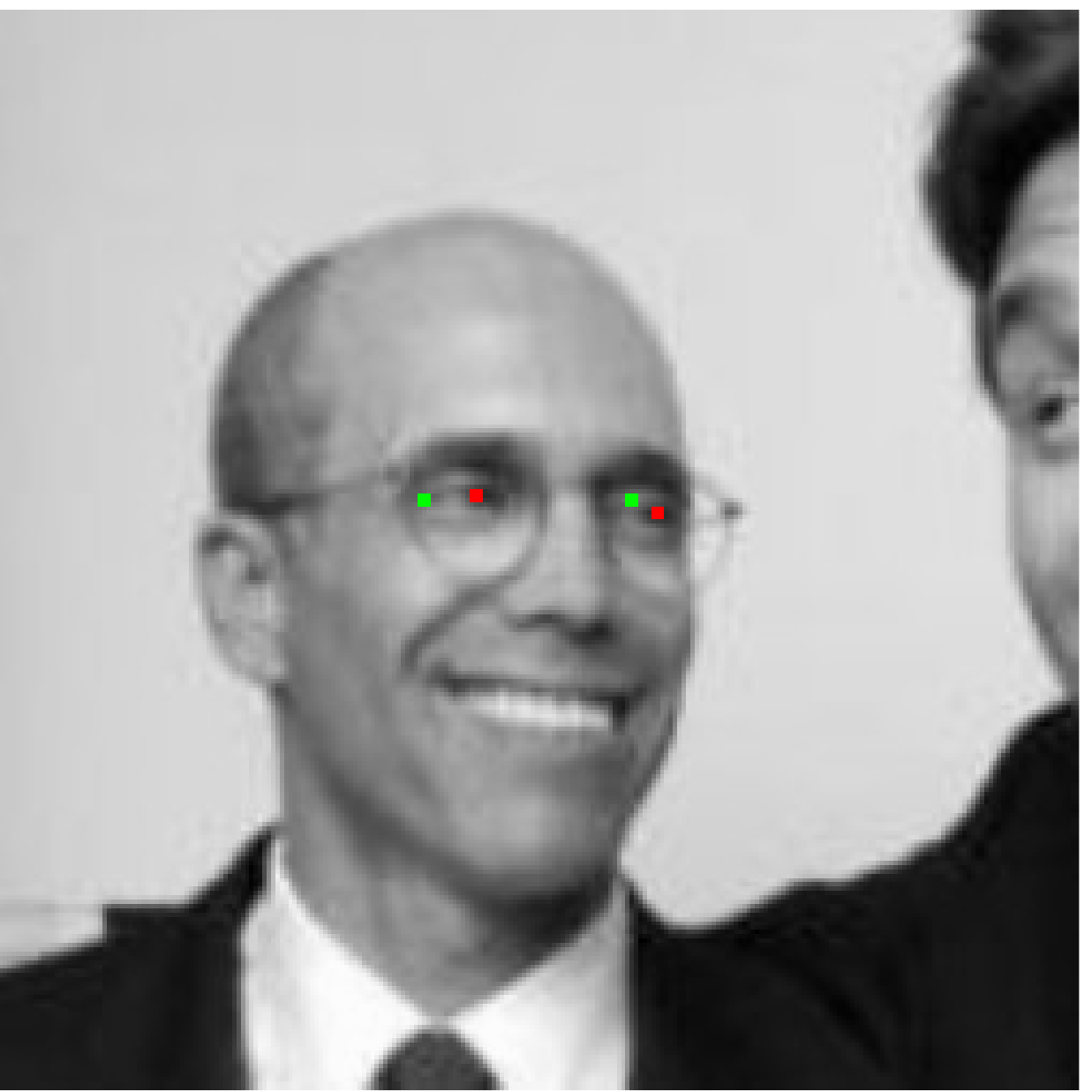} \\
    \end{tabular}
   \caption{Images from Labelled Face in the Wild (LFW) database.
   The top row show images having accuracy higher than 0.05, medium row
   accuracy between 0.05 and 0.1 and bottom row accuracy lower than 0.1.}
   \label{Fig:LFW_results}
\end{figure}

\begin{figure*}[tb]
\center
    \begin{tabular}{cc}
        \includegraphics[width=0.4 \textwidth]{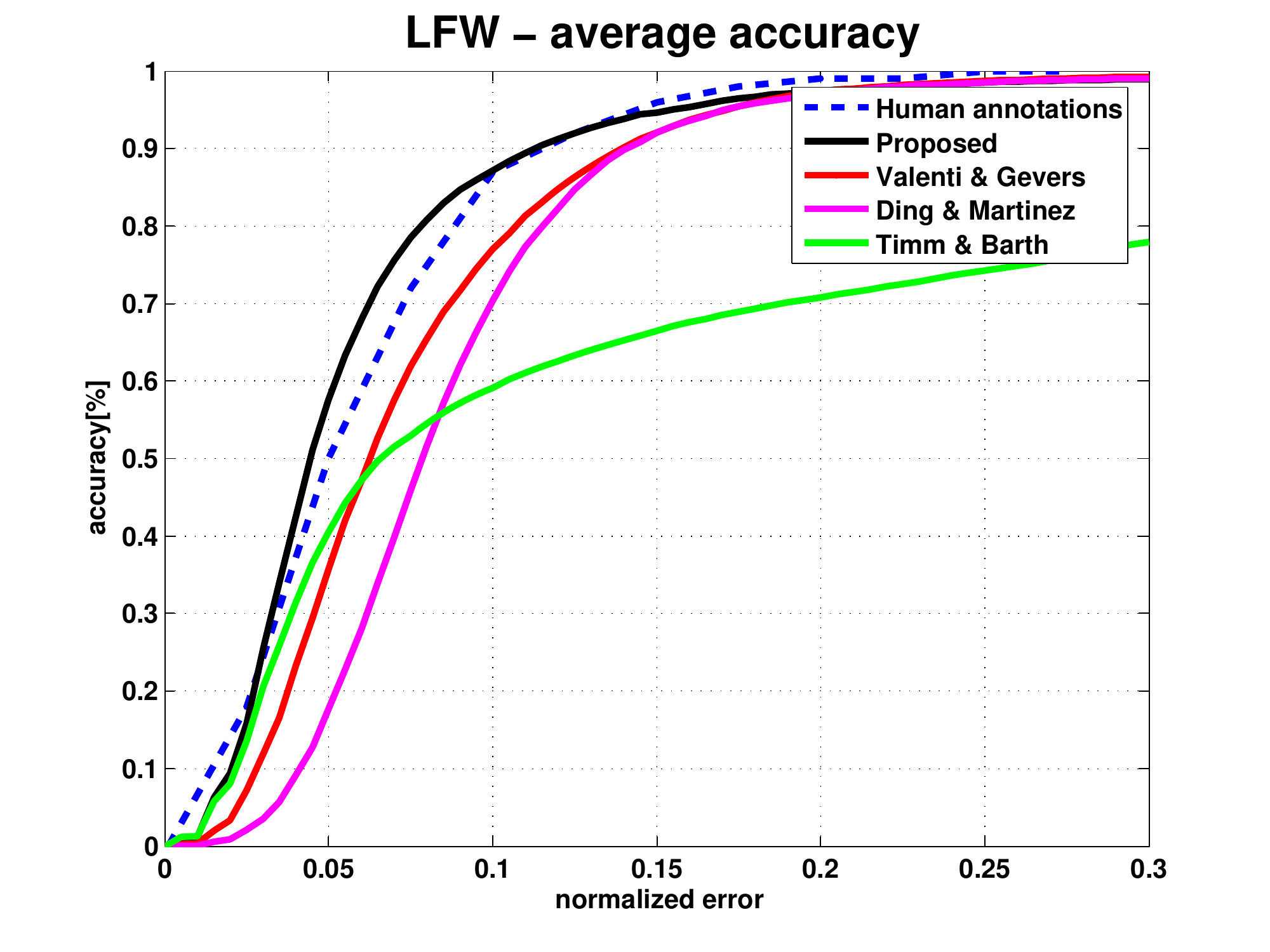} &
        \includegraphics[width=0.4 \textwidth]{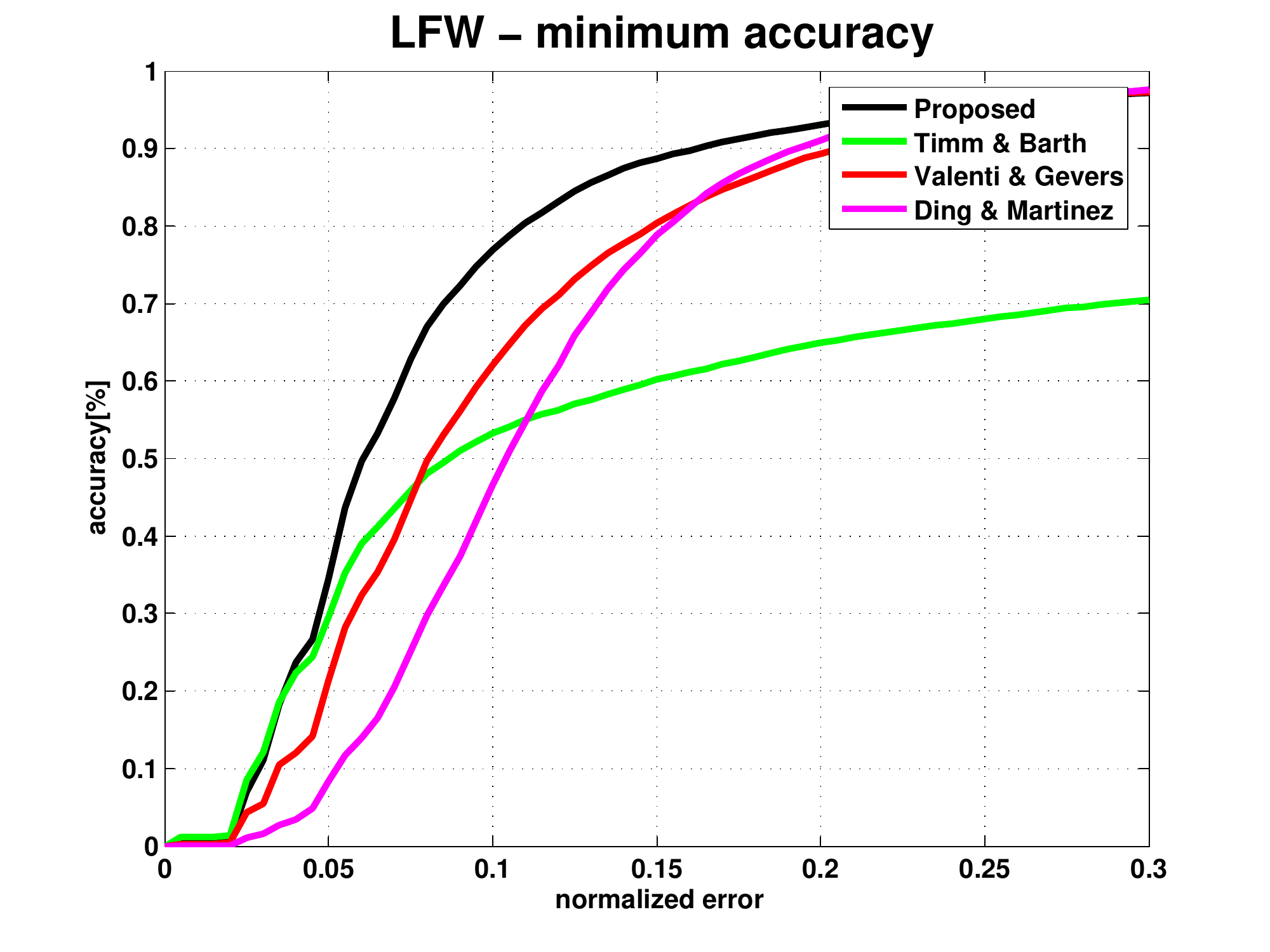} \\
        (a) & (b) \\
    \end{tabular}
   \caption{Results achieved on the LFW database: (a) Average accuracy and (b) minimum accuracy  as imposed
   by the stringent criterion - eq. (\ref{Eq:StringCrit}).
    With dashed blue line is the average error for human evaluation , black line -- proposed method,
    red line -- \citeyear{Valenti:12}, green line -- \citeyear{Tim:11} and
    with magenta line --  \citeyear{Ding:10}}.
    \label{Fig:LFW resultsNum}
\end{figure*}

Examples of the results achieved  on the LFW database may be seen in figure \ref{Fig:LFW_results}.
Numerical results, compared with the solution from \cite{Valenti:12}, \cite{Tim:11}, \cite{Ding:10}
and with human evaluation error are presented in figure \ref{Fig:LFW resultsNum}.

Regarding the achieved results, we note that even that our method was designed to work  on large
resolution faces, it provides accurate results when applied on smaller ones. As one can see in
figure \ref{Fig:LFW resultsNum} we significantly outperform the state of the art solutions
\cite{Valenti:12} and from \cite{Tim:11} by almost 50\% improvement at $\epsilon < 0.05$ accuracy,
on an over 12000 image database that presents as close to real-life as possible cases and with more
the method from \cite{Ding:10}.

\subsection{Algorithm Complexity}

The entire algorithm requires only four divisions for the projections normalization and two for
determination of the region weighting center with variable denominator per eye crop, and no high
precision operations, therefore needing only limited fixed point precision. The ZEP+MLP combination
is linear with respect to the size of scan eye rectangle $O(\delta N)$. The method was implemented
in C around OpenCV functionality, on an Intel i7 at 2.7 GHz, on single thread and it takes 6 msec
for both eyes on a face square of $ 300 \times 300 $ pixels, which is a typical face size for HD -
720p ($1280\times 720$ pixels) format.  We note that additional 7 msec are required for Face
Detection.

The code can run in real-time while including face detection and further face expression analysis.
Comparison with state of the art methods may be followed in the  table \ref{Tab:Rez_Time} when
comparing with other eye localization methods and on right hand side of table
\ref{Tab:Rez_BioIDFiduc} when discussing face fiducial points localization solutions; for some
works the authors have not reported speed performance, but taking into account algorithm
complexity, it is reasonable to presume that it is too large for real-time.

Trying to overcome the difficulty of comparison while different platforms were used for
implementation, we rely as a unifying factor on the single thread benchmarking score provided by
\cite{CPUbenchmark} for specific CPU; this score will be denoted by $CPU_{s}$. It must be noted
that such numbers should be considered with precaution since there do exist several CPUs that
correspond to the description provided by authors (and we always took the best case) and the
benchmark test may not be very relevant for the specific processing required by a solution.

To aggregate the overall time performance of a method we used the following formula:
\begin{equation}
  T_p = \frac{ fps \times \min \{M,N\} }{CPU_{power}}
  \label{Eq:T_power}
\end{equation}
where $M \times N$ is the frame size used for reported results. Note that the formula uses only one
of the two dimensions that describe an image to cope with different aspect ratios.

The results for eye localization aggregated with the measure in equation (\ref{Eq:T_power}) are
showed in the  table \ref{Tab:Rez_Time} when comparing with other eye localization methods. Our
method rank second following the one proposed by \citeyear{Jesorsky:01}, but it gives consistently
better results in terms  of accuracy.

It has to be noticed that while, initially only \citeyear{Valenti:12} reported comparable
computational time, after integrating the larger frame size with processing power, our method turns
to be 1.5 times faster. Furthermore, to be able to directly compare our computation time, we have
modified the size of input face to be $120 \times 120$ which corresponds to a $320 \times 240$
pixels image, letting everything else the same and we find out that our method requires 1.6 msec to
localize both eyes; given the additional 7 msec for face detection, we get a total time of 8.6
msec, that is equivalent with a frame rate of 116 frames/sec, proving that we clearly outperform
the method from \cite{Valenti:12}.

\begin{table*}[tb]
\centering
  \caption{ Comparison with state of the art (listed in chronological appearance) in terms of time requirements.
    Regarding time performance,we compare ourself only with papers that provide some measure of duration.
    Also we note that, in general, authors reported non-optimized results on PC platforms and various
   image sizes. The reported time for \cite{Asadifard:10} is taken from \cite{Ciesla:12}. With gray
   background we marked the best state of the art result. $T_p$ is given by equation \ref{Eq:T_power}
   and higher values are desired. }
\label{Tab:Rez_Time}
\begin{tabular}{|c|c|}
 \hline
   { } & \textbf{Time performance} \\ \hline 
   \begin{tabular}{c}
       { \phantom{Method}} \\ \hline
       \textbf{Proposed}                   \\ \hline
         \cite{Jesorsky:01}        \\ \hline
         \cite{Zhou:04}                \\ \hline
         \cite{Cristinacce:04}  \\ \hline
         \cite{Campadelli:06}    \\ \hline
         \cite{Hamouz:05}           \\ \hline
         \cite{Turkan:08}            \\ \hline
         \cite{Kroon:08}              \\ \hline
         \cite{Asteriadis:09}    \\ \hline
         \cite{Asadifard:10}      \\ \hline
        \cite{Valenti:12}+MS          \\ \hline
        \cite{Valenti:12}+SIFT        \\ \hline
        \cite{Florea:12}           \\
    \end{tabular} &

    \begin{tabular}{c|c|c|c|c}
     \emph{FrameRate}&  \emph{Image size}& \emph{Platform}& $CPU_{s}$ &  $T_p$    \\ \hline
     \textbf{76 fps} & \textbf{1280}$\times$ \textbf{720} & \textbf{i7 2.7 GHz} & \textbf{1747} & \textbf{31.23}\\ \hline
       33 fps        & $384 \times 288$  & P3 850MHz      & $\approx 200$ & \colorbox{lightgray}{47.52}\qquad \\ \hline 
      15fps          & $320\times 240$   & Core 2         & 981    & 3.67 \\ \hline 
      0.7 fps       & $384\times 286$    & P3 500MHz      & 140    & 1.42 \\ \hline 
        0.08 fps    & $384\times 286$    & P4 3.2GHz      & 720    & 0.032 \\ \hline 
        0.07 fps    & $720 \times 576$   & P4 2.8GHz      & 618    & 0.065 \\ \hline 
        12fps       & $384\times 286$    & n/a            & n/a    & n/a  \\ \hline 
        2fps        & n/a                & n/a            & n/a    & n/a  \\ \hline 
        3.84 fps    & $384\times 286$    & PM 1.6 GHz     & 514    & 2.13 \\ \hline 
        15 fps      & $320\times 240$    & Core 2         & 981    & 3.67 \\ \hline 
        90 fps      & $320\times 240$    & Core2 2.4GHz   & 981    & 22.01  \\ \hline 
        29 fps      & $320\times 240$    & Core2 2.4GHz   & 981    & 7.09 \\ \hline 
        \hline
    \end{tabular} \\
 \hline
 \end{tabular}

\end{table*}

\subsection{Discussion}

The previous subsections within this "Results and discussions" part have guided through various
experiments and measurements that present a through comparison of eye localization performance of
the here proposed ZEP eye localization method, which is shown to perform remarkably well among a
wide variety of conditions and datasets. Some of the presented numbers and experiments deserve yet
a supplemental emphasis and clarification.

A first issue of discussion is related to the experimental setup, namely the databases that are
currently used in the assessment of algorithms accuracy. BioID has gained through the years
widespread recognition, as it was one of the earliest face image databases that contain facial
landmark ground truth annotations. As such, BioID was intensively used for accuracy comparisons,
with a clear tendency over time to concentrate the efforts in getting top results on BioID alone.
As one may have noticed in table \ref{Tab:Rez_BioID}, the here proposed method is outperformed on
BioID by the algorithms proposed by  \citeyear{Tim:11} and \citeyear{Valenti:12}.

We can notice that most the methods are overtrained in standard conditions, and thus perform very
well within their over--learned domain.  As such, we claim that these approaches are not relevant
in a broader, real-life testing scenario. The approaches proposed by \citeyear{Valenti:12} and
\citeyear{Tim:11} are thus retained as a significant eye location methods and we further tested
them; we also included the solution from \cite{Ding:10} as being a high profile method build
outside the BioID database; the results showed that we outperform these methods by a gross margin.

As anyone knowledgeable in the field observes, the BioID database contains mostly frontal pose,
frontal illumination and neutral expression faces, and catches only a small glimpse of the problems
related to eye localization. As such, intensive performance comparison must be realized outside
these standard conditions, as \citeyear{Valenti:12} does in the case of varying illumination and
pose and we do in the case of noise, variable illumination, expression and pose variations. Several
tests that are reviewed again here prove the superior performance of the proposed ZEP eye
localization method in these extreme conditions.

The non-frontal illumination and the subject pose variations are key issues in real-life,
unconstrained applications. Typically these are tested within the Extended YaleB database, where
ZEP performs marginally better (+2\%) than method in \cite{Valenti:12} and significantly better
than \cite{Tim:11} and \cite{Ding:10}, as shown in Table \ref{Tab:YaleB+Comparat}. Subject
emotional expressions hugely affect eye shape and surroundings. The Cohn-Kanade database is the
state of the art testbed in emotion-related tasks; in this case, the ZEP eye localization
outperforms \cite{Tim:11} by 10\%, \cite{Valenti:12} by some 30\%, and \cite{Ding:10} by 60\%  as
shown in table \ref{Tab:RezCK}.

Within all databases, closed eyes present an independent challenge. As noticed by the authors in
\cite{Valenti:12}, their method is prone to errors in detecting the closed eye center (which is
confirmed by the experiments across all databases). The proposed ZEP method is much more robust to
closed eyes, due to the way in which the eye profile is described within the proposed encoding of
the luminance profiles.

Finally, we consider that the most relevant test is performed on the LFW database, taking into
account the size (more than 12000 images), image resolution (extremely low) and especially the fact
that images were acquired ``in the wild''. Yet, on the LFW database, which is currently one of the
most challenging tasks, we outperform the method in \cite{Tim:11} by at least 5\%  the one in
\cite{Valenti:12} by a gross margin (+13\%) and respectively the method form \cite{Ding:10} by near
30\%. Nonetheless we much closer to human accuracy (as shown in figure \ref{Fig:LFW resultsNum} (a)
).

Regarding the computational complexity, the here proposed method requires a computational time
which is inferior to the time required by the method from \cite{Jesorsky:01}; yet the accuracy of
the here proposed method is significantly higher. If we compare only the computation time, without
considering the image size, one may consider the method from \cite{Valenti:12} to be faster. Yet,
tests showed that the here proposed solution is still faster than the implementation from
\cite{Valenti:12} at equal image resolution (namely $320 \times 240$). We thus claim that the here
proposed method is the fastest solution from the select group of high accuracy methods.

\section{Conclusions}
\label{Sect:Conclusions}

In this paper, we proposed a new method to estimate the eye centers location using a combination of
Zero-based Encoded image Projections and a MLP. The eye location is determined by discriminating
between eyes and non-eyes by analyzing of the normalized image projections encoded with the
zero-crossing based method. The extensive evaluation of the proposed approach showed that it can
achieve real-time high accuracy. While the ZEP feature was used for eye description, we consider
that it is general-enough and may be used in numerous problems.

\bibliographystyle{icml2014}
\bibliography{Zep_Eye_Full_Arxiv}

\end{document}